\documentclass[sigconf]{aamas}

\usepackage{amsfonts}
\usepackage{balance} 
\usepackage{booktabs}
\usepackage[capitalize,noabbrev]{cleveref}
\usepackage[inline]{enumitem}
\usepackage{float}
\usepackage{flushend}
\usepackage{graphicx}
\usepackage{hyperref}
\usepackage{microtype}
\usepackage{placeins}
\usepackage{siunitx}
\usepackage{subcaption}
\usepackage{verbatim}
\usepackage{xcolor}
\usepackage{xspace}

\makeatletter
\DeclareRobustCommand\onedot{\futurelet\@let@token\@onedot}
\def\@onedot{\ifx\@let@token.\else.\null\fi\xspace}
\def\ie{\emph{i.e}\onedot} 

\newcommand{\longcell}[2][t]{%
  \begin{tabular}[#1]{@{}c@{}}#2\end{tabular}}

\newcommand{\vast}{\bBigg@{4}}
\newcommand{\Vast}{\bBigg@{5}}

\makeatletter
\gdef\@copyrightpermission{
  \begin{minipage}{0.2\columnwidth}
   \href{https://creativecommons.org/licenses/by/4.0/}{\includegraphics[width=0.90\textwidth]{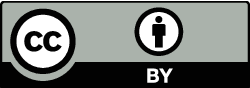}}
  \end{minipage}\hfill
  \begin{minipage}{0.8\columnwidth}
   \href{https://creativecommons.org/licenses/by/4.0/}{This work is licensed under a Creative Commons Attribution International 4.0 License.}
  \end{minipage}
  \vspace{5pt}
}
\makeatother

\setcopyright{ifaamas}
\acmConference[AAMAS '25]{Proc.\@ of the 24th International Conference
on Autonomous Agents and Multiagent Systems (AAMAS 2025)}{May 19 -- 23, 2025}
{Detroit, Michigan, USA}{Y.~Vorobeychik, S.~Das, A.~Nowé  (eds.)}
\copyrightyear{2025}
\acmYear{2025}
\acmDOI{}
\acmPrice{}
\acmISBN{}

\acmSubmissionID{1307}

\title[Implicit Repair with Reinforcement Learning in Emergent Communication]{Implicit Repair with Reinforcement Learning in Emergent Communication}

\author{F\'abio Vital}
\affiliation{
  \institution{INESC-ID \& Instituto Superior T\'ecnico}
  \city{Lisboa}
  \country{Portugal}}

\author{Alberto Sardinha}
\affiliation{
  \institution{INESC-ID \& PUC-Rio}
  \city{Rio de Janeiro}
  \country{Brazil}}

\author{Francisco S. Melo}
\affiliation{
  \institution{INESC-ID \& Instituto Superior T\'ecnico}
  \city{Lisboa}
  \country{Portugal}}

\usepackage{amsmath,amsfonts,bm}

\def\eqref#1{equation~\ref{#1}}
\def\Eqref#1{Equation~\ref{#1}}

\def\1{\bm{1}}

\def\rp{{\textnormal{p}}}

\def\vzero{{\bm{0}}}

\def\vtheta{{\bm{\theta}}}
\def\vphi{{\bm{\phi}}}

\def\vvarepsilon{{\bm{\varepsilon}}}

\def\vc{{\bm{c}}}
\def\vd{{\bm{d}}}

\def\vl{{\bm{l}}}
\def\vm{{\bm{m}}}

\def\vs{{\bm{s}}}

\def\vx{{\bm{x}}}

\def\vz{{\bm{z}}}

\def\mI{{\bm{I}}}

\def\mM{{\bm{M}}}

\DeclareMathAlphabet{\mathsfit}{\encodingdefault}{\sfdefault}{m}{sl}
\SetMathAlphabet{\mathsfit}{bold}{\encodingdefault}{\sfdefault}{bx}{n}

\def\sC{{\mathbb{C}}}

\def\sJ{{\mathbb{J}}}

\def\sN{{\mathbb{N}}}

\def\sR{{\mathbb{R}}}

\def\sW{{\mathbb{W}}}
\def\sX{{\mathbb{X}}}

\newcommand{\E}{\mathbb{E}}

\DeclareMathOperator{\sg}{sg}

\DeclareMathOperator{\choice}{choice}
  
\begin{abstract}
\emph{Conversational repair} is a mechanism used to detect and resolve miscommunication and misinformation problems when two or more agents interact. One particular and underexplored form of repair in emergent communication is the implicit repair mechanism, where the interlocutor purposely conveys the desired information in such a way as to prevent misinformation from any other interlocutor. This work explores how redundancy can modify the emergent communication protocol to continue conveying the necessary information to complete the underlying task, even with additional external environmental pressures such as noise. We focus on extending the signaling game, called the Lewis Game, by adding noise in the communication channel and inputs received by the agents. Our analysis shows that agents add redundancy to the transmitted messages as an outcome to prevent the negative impact of noise on the task success. Additionally, we observe that the emerging communication protocol's generalization capabilities remain equivalent to architectures employed in simpler games that are entirely deterministic. Additionally, our method is the only one suitable for producing robust communication protocols that can handle cases with and without noise while maintaining increased generalization performance levels. Our code and appendix are available at \url{https://fgmv.me/projects/noisy-emcom}. First author correspondence: \href{mailto:fabiovital@tecnico.ulisboa.pt}{fabiovital@tecnico.ulisboa.pt}.
\end{abstract}


\newcommand{\BibTeX}{\rm B\kern-.05em{\sc i\kern-.025em b}\kern-.08em\TeX}

\begin{document}


\pagestyle{fancy}
\fancyhead{}


\maketitle 


\section{Introduction}

\emph{Emergent Communication} (EC) is a field that recently gained attention in  Machine Learning (ML) research. The progress of language evolution research~\citep{wagner2003progress,choicompositional,Ren2020Compositional,chaabouni-etal-2020-compositionality,chaabouni2019anti,ueda-washio-2021-relationship} and the conceptualization of artificial languages for robot and human-robot communication~\citep{bogin2018emergence,mordatch2018emergence,havrylov2017emergence,jorge2016learning} are some of the fundamental motivations behind the recent rise in interest. Mainly, EC focuses on developing experiments where a group of agents must learn how to communicate without prior knowledge to achieve a common goal, where coordination and cooperation are essential for the group's success~\citep{winograd1972understanding,Lewis1979,nowak1999evolution}. This approach differs from the current state of the art in natural language processing (NLP), where large language models (LLMs) dominate the field. LLMs are supervised statistical models that optimize the prediction of the next token given a context (textual input)~\citep{openai2023gpt4,jiang2023mistral,dubey2024llama,dao2024transformers}. It is still an open question whether working only in the language space (as LLMs do) is enough to create agents with an intrinsic and deeper meaning about the world that are capable of adapting to novel circumstances effectively~\citep{bisk2020experience}. As such, we argue that exploring different approaches, like EC, is crucial to continue advancing the field of NLP.

The main focus of this work comprises the study of a specific topic in language evolution called \emph{conversational repair}~\citep{schegloff1977preference}. In linguistics, conversational repair is already a known topic that plays an important role in establishing complex and efficient communication protocols~\citep{albert2018repair}. In short, conversational repair aggregates any communication mechanism employed by any interlocutor to initiate a process to detect and clarify some information being transferred by any other interlocutor. To give more context on how our work relates to previous literature, we divide repair mechanisms into two broad categories: implicit and explicit, a coarser partitioning of the one introduced by~\citet{lemon2022conversational}. Explicit repair mechanisms happen when an interlocutor distinctly starts a follow-up interaction (communication) in order to clarify some conveyed past information, e.g., ``Is it the blue one?'' (confirmation); ``Is it the first or the second one?'' (clarification). On the other hand, implicit mechanisms happen in a subtle way where the interlocutor, conveying the original information, intentionally expresses it in such a way as to minimize misinformation and preemptively avoid posterior conversational repair phases altogether. The implicit conversational repair mechanism also has connections to the concept of redundancy in linguistics and communication analysis~\citep{cherry1966human}. Redundancy appears in every human language, at the semantic and syntax level, and appears in the form of repetition or extra content to send. Additionally, the sending interlocutor may apply different levels of redundancy that she/he finds necessary to convey the information given the target audience and medium used for communication.

Most previous works in EC employ variations of a signaling game called the Lewis Game (LG)~\citep{Lewis1979}, with the primary purpose of analyzing how a communication protocol emerges as the result of achieving cooperation to solve the game~\citep{choicompositional,jorge2016learning,guo2019emergence}. In the LG, the Speaker describes an object to the Listener, who then has to discriminate it against a set of distractor objects. We call the union of the assigned object and distractors the \emph{candidates}. Regarding this study, we extend a variation introduced by~\citet{chaabouni2022emergent}, where real images are used as the objects to discriminate instead of categorical inputs, and the number of candidates given to the Listener increases in several orders of magnitude, conveying a more complex game than in previous works~\citep{ueda-washio-2021-relationship,mordatch2018emergence,lazaridou2018emergence,rita2022on,chaabouni-etal-2020-compositionality,chaabouni2019anti}. In the original work, the authors propose a supervised training routine where the Listener receives the correct answer after each game. However, this implementation diverges from human communication, where there is usually no direct supervision on how effective a particular dialogue can be~\citep{hayashi2013conversational}. We propose modeling both agents as RL agents. As such, the only (semi) supervised information given to the agents is the outcome of the game. Similar to previous works~\citep{havrylov2017emergence,lazaridou2018emergence,li2019ease,rita2022on}, we model both agents using Reinforce~\citep{williams1992simple}.

Furthermore, as a means to study implicit repair mechanisms, we define a new game variation with faulty communication channels that can introduce noise into the messages. This new game setup has the necessary conditions to study if agents can detect and overcome miscommunication/misinformation to solve the game cooperatively. Our analysis shows that the emerging communication protocols have redundancy built in to prevent the adverse effects of noise, where even partial messages have enough information for the Listener to select the correct candidate. Additionally, we show that the training in the noisy game produces communication protocols that are highly robust to noise, being effective in different noise levels, even without noise, at test time.

Previous literature has already addressed the problem of explicit conversational repair. \citet{lemon2022conversational} propose new research directions on how to embed conversational repair into EC tasks, where the repair mechanism acts as a catalyst to fix misalignments, for example, in the language learned by each agent for a specific cooperative task. Moreover, another recent work develops an extended version of the Lewis Game to enable a feedback mechanism from the Listener to the Speaker, mimicking the initialization of an external repair mechanism phase~\citep{nikolaus2023emergent}. However, some limitations compromise the co-relation to human languages. First, the feedback sent by the Listener contains minimal information (single binary token), and such feedback is sent after every token in the message, breaking the turn-taking nature of the dialogue. Compared to our work, we designed a more challenging game where we prevent cyclic feedback (from the Listener to the Speaker), meaning the Speaker does not receive direct feedback about how noise affects the messages being transferred. In our case, the Speaker only knows the result of the game. Consequently, the Speaker needs to understand through trial and error how to convey information to facilitate the Listener's job, inducing an implicit repair mechanism, as explained previously. 

To summarize, our contributions are 3-fold. First, although previous works introduce game designs featuring noisy channels, we contribute with a rigorous mathematical derivation on how to aggregate noise into the LG. We found such inference lacking in the literature. Furthermore, we define a new noisy game variant where the input objects to discriminate are injected with noise. We use this new variant as an out-of-distribution game to evaluate how the trained protocols react to new forms of noise. Secondly, we demonstrate the effectiveness of employing RL agents on complex LG variants featuring the discrimination of natural images and noisy communication channels. Additionally, we showcase that the RL variant achieves better results than the original architecture with a supervised Listener. We emphasize that our objective is not to benchmark different RL algorithms but to show that implementing the Listener as an RL agent can bring advantages against the Supervised counterpart, where even a straightforward implementation of Reinforce is enough to observe substantial gains already. Third, we analyze and show how more complex game designs, such as introducing noise to the LG, guide the agents to resort to redundancy measures to complete the game efficiently, mimicking implicit conversational repair mechanisms. We show how the protocols emerging to communicate through noisy channels have better generalization capabilities and robustness to different noise levels at test time. Additionally, we illustrate that these improvements are a side-effect of resorting to redundancy in the messages sent.



\section{Methodology} \label{sec:meth}

We start this section by defining a noisy variation of the LG ,called the Noisy Lewis Game (NLG). The main change in the NLG incorporates a faulty communication channel where noise interferes with the transmitted messages by masking a subset of the tokens. This game variation is more complex than the LG, where the (RL) game environment becomes stochastic. We further note that the original LG is a simplification of the NLG where we fix the noise level at zero. Afterward, we detail how the Speaker converts the received input into a message, a sequence of discrete tokens, and how the Listener processes the message and candidates to make decisions. We impose a RIAL setting~\citep{foerster2016learning}, where agents are independent and perceive the other as part of the environment. Hence, we describe the learning strategy for both agents independently, explaining the loss composition and the importance of each loss term to guide training where functional communications protocols can emerge.

\subsection{Noisy Lewis Game (NLG)} \label{sec:meth-nlg}
The Noisy Lewis Game (NLG) is a discrimination game in which one of the agents, the \emph{Speaker}, must describe an object by sending a message to the other agent, the \emph{Listener}. When the game starts, the Speaker receives a target image \(\vx\) retrieved from a fixed dataset \(\vx\in\sX\) and describes it by generating a message, \(m: \sX\times\sR^K\rightarrow \sW^N\), where \(\sW\) is a finite vocabulary, and \(\vtheta\in\sR^K\) parametrizes \(m\). The message comprises \(N\) discrete tokens, \(m\left(\vx;\vtheta\right)=\left(m_t\left(\vx;\vtheta\right)\right)_{t=1}^{N}\), where \(m_t(\vx;\vtheta)\in\sW\). Due to the noisy nature of the communication channel, the Listener can receive a message with unexpected modifications. We model this perturbation with the function \(n:\sW^N\rightarrow\sW'^N\), where the function processes each token independently and converts it into a default unknown token with a given probability. As such, \(\sW'\) is the union of the original vocabulary plus the unknown token, \(\sW'=\sW\cup\{\mathtt{unk}\}\). We describe introduce \(n\) as :
\begin{equation}
\begin{split}
    n&\left(m\left(\vx;\vtheta\right)\right) = \left( n_t\left(m_t\left(\vx;\vtheta\right)\right) \right)_{t=1}^{N}\\
    & \text{s.t.}\;\;n_t\left(m_t\left(\vx;\vtheta\right)\right) = 
    \begin{cases}
        m_t(\vx;\vtheta), & \mbox{if } \rp > \lambda \\
        \mathtt{unk}, & \mbox{otherwise},
    \end{cases}
\end{split}
\label{eq:lg-noise}
\end{equation}
where \(\rp\) is sampled from a uniform distribution, \(\rp\sim \mathcal{U}\left(0,1\right)\), and \(\lambda\in\left[0,1\right)\) is a fixed threshold, indicating the noise level present in the communication channel. By definition, the Speaker is agnostic to this process and will never know if the message was modified. For simplicity, we define \(\vm\) to describe a (noisy) message given some input, \(\vm=m\left(\vx,\vtheta\right)\) or \(\vm=n\left(m\left(\vx,\vtheta\right)\right)\). We also refer each message token as \(m_t\) instead of \(m_t(\vx;\vtheta)\), omitting the domain.

Subsequently, the Listener receives the message along with a set of candidate images, \(\sC\in \left[\sX\right]^C\), where \(\left[\sX\right]^C\) defines the set of all subsets with \(C\) elements from \(\sX\). With both inputs, the Listener tries to identify the image the Speaker received, \(\vx\). We define \(\choice:\sW'^N\times\left[\sX\right]^C\times\sR^{k'}\rightarrow\sJ\) to specify the Listener's discrimination process, where \(\sJ\subset\sN\) is a particular enumeration of \(\sC\), such that \(\sC=\bigcup_{j\in\sJ}\sC_j\), \(|\sJ|=C\), and \(\vphi\in\sR^{k'}\) parametrizes \(\choice\). Therefore, the index outputted by the Listener, \(j=\choice\left(\vm,\sC;\vphi\right)\), is the \(j\)-th element in \(\sC\), denoting the final guess \(\hat{\vx}=\sC_j\).

\begin{figure}[!t]
  \centering
  \includegraphics[width=0.81\linewidth,keepaspectratio]{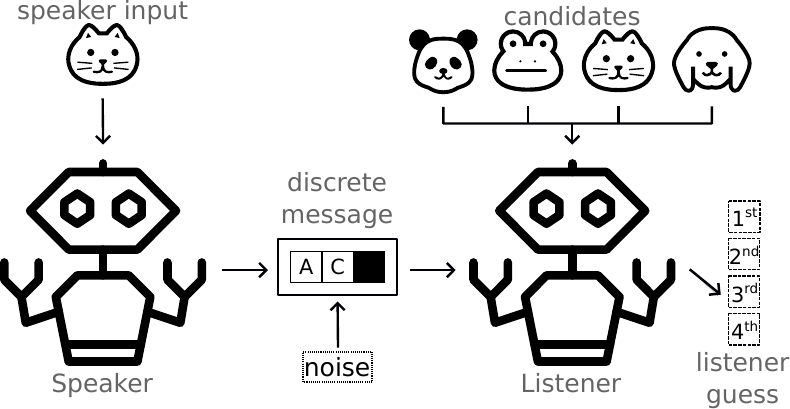}
  \caption{Visual Representation of the Noisy Lewis Game (NLG). In this illustration, the message, \(\vm\), contains three tokens (\(N=3\)), where the last one is masked.}
  \label{fig:emcom-nlg}
\end{figure}

Both agents receive a positive reward if the Listener correctly identifies the target image \(\vx\) and a negative reward otherwise:
\begin{equation}
    R(\vx,\hat{\vx}) = 
    \begin{cases}
        1, & \mbox{if } \hat{\vx}=\vx \\
        -1, & \mbox{if } \text{otherwise}.
    \end{cases}
    \label{eq:lg-rwd}
\end{equation}
Lastly, note that the original LG is a specification of the NLG where we set \(\lambda=0\). \Cref{fig:emcom-nlg} depicts a visual representation of NLG. For completeness, the LG is depicted in \Cref{app:lg}.

\subsection{Agent Architectures} \label{sec:meth-arch}
\begin{figure*}[!t]
    \begin{center}
    \begin{subfigure}{.49\textwidth}
      \centering
      \centerline{\includegraphics[width=0.4\linewidth]{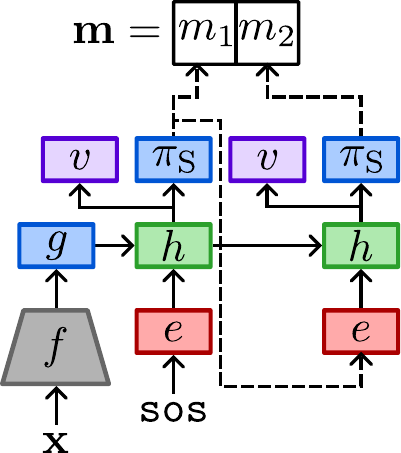}}
      \caption{}
      \label{fig:speaker-arch}
    \end{subfigure}
    \hfill
    \begin{subfigure}{.49\textwidth}
      \centering
      \centerline{\includegraphics[width=0.52\linewidth]{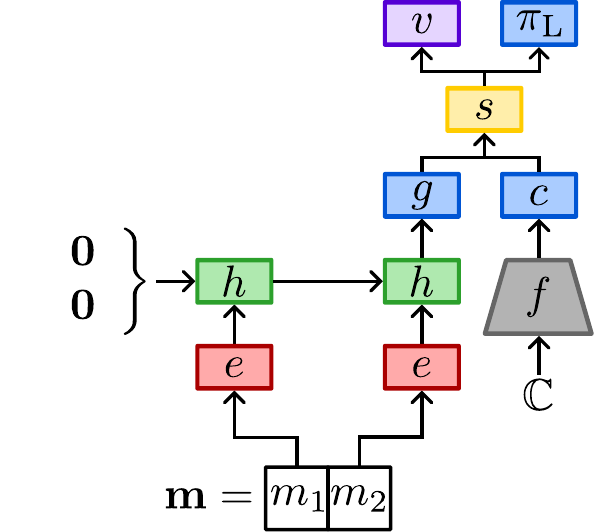}}
      \caption{}
      \label{fig:list-arch}
    \end{subfigure}
    \end{center}
    \caption{Graphical representation of Speaker, \Cref{sub@fig:speaker-arch}, and Listener, \Cref{sub@fig:list-arch}, architectures for the NLG. In this illustration, the message, \(\vm\), contains only two tokens, \(N=2\).}
    \label{fig:speaker-list-archs}
    \vspace{-2ex}
\end{figure*}

We now describe the architectures implemented for both agents, the Speaker and the Listener (\Cref{fig:speaker-list-archs}). We design both architectures to be able to model policy gradient RL algorithms~\citep{sutton2018reinforcement}.

As an overview, the Speaker's objective is to encode a discrete message, \(\vm\), describing an input image, \(\vx\), see \Cref{sec:meth-nlg}. First, we encode the image using a pre-trained image encoder~\citep{grill2020bootstrap}, \(f\), to reduce its dimensionality and extract valuable features, \(\vx'=f\left(\vx\right)\). Subsequently, a trainable encoder \(g\) processes the new sequence of features, outputting the initial hidden and cell values, \((\vz_{0,\vtheta},\vc_{0,\vtheta})=g\left(\vx';\vtheta\right)\), used by the recurrent module \(h\), in this case, an LSTM~\citep{hochreiter1997long}.

Subsequently, the Speaker will select each token \(m_t\) to add to the message iteratively, using \(h\). On this account, we define a complementary embedding module, \(e\), to convert the previous discrete token \(m_{t-1}\) into a dense vector \(\vd_{t,\vtheta}=e\left(m_{t-1};\vtheta\right)\). Then, the recurrent module, \(h\), consumes the new dense vector and previous internal states to produce the new ones, \((\vz_{t,\vtheta},\vc_{t,\vtheta})=h\left(\vd_{t,\vtheta},\vz_{t-1,\vtheta},\vc_{t-1,\vtheta};\vtheta\right)\). We then pass \(\vz_{t,\vtheta}\) through two concurrent heads:
\begin{enumerate*}[label=(\roman*)]
    \item The actor head yields the probability of choosing each token as the next one, \(m_t\sim\pi_S\left(\cdot|\vz_{t,\vtheta};\vtheta\right)\);
    \item The critic head estimates the expected reward \(V\left(\vx\right) := v\left(\vz_{t,\theta};\theta\right)\).
\end{enumerate*}
After the new token is sampled, we feed it back to \(e(\cdot\,;\theta)\), and the process repeats itself until we generate \(N\) tokens. The first token \(m_0\) is a predefined \emph{start-of-string} token and is not included in the message. Following~\cite{chaabouni2022emergent}, we maintain the original vocabulary and message sizes, where \(|\sW|=20\), and \(N=10\), making the set of all possible message much larger than the size of the dataset used (\(|\sX|\approx10^6\) for ImageNet~\cite{ILSVRC15}). We depict the Speaker's architecture in \Cref{fig:speaker-arch}.

The Listener architecture has two different modules to process the message, sent from the Speaker, \(\vm\) and the images obtained as candidates \(\sC\). Additionally, a third module combines the output of both input components and provides it to the actor and critic heads. We now describe each component.

To process the candidate images \(\sC\), the Listener uses the same pre-trained encoder \(f\) combined with the network \(c\) to embed the candidate images, \(\vl_{j}=c(\vx_j';\vphi)\), where \(\vx_j'=f(\vx_j)\) and \(\vx_j\in\sC\).

Concerning the message received, the Listener uses the recurrent model \(h\) (an LSTM) to handle each token, \(m_t\), iteratively. Similarly to the Speaker, there is an embedding layer, \(e(\cdot\,;\vphi)\), to convert the discrete token into a dense vector before giving it to \(h\), where we have \((\vz_{t,\vphi}, \vc_{t,\vphi}) = h(e(m_t;\vphi), \vz_{t-1,\vphi}, \vc_{t-1};\vphi)\). The initial internal states of \(h\) are initialized as \(\vz_{0,\vphi}=\vzero\) and \(\vc_{0,\vphi}=\vzero\). After processing all message tokens, the final hidden state, \(\vz_{N,\vphi}\), goes through a final network \(g\) to output the message's hidden value \(\vl_\text{m}=g(\vz_{N,\vphi};\vphi)\). Finally, the generated hidden values for the message and all candidates flow through to the head module.

The first operation in the head module executes an attention mechanism to combine the message features with each candidate's counterpart. The output includes a value per candidate which we concatenate into a vector \(\vs=\left[\begin{matrix}\vl_\text{m}\cdot \vl_{1} & \ldots & \vl_\text{m}\cdot \vl_{{|\sC|}}\end{matrix}\right]^T\), called the candidates' score. We define the actor head as \(\pi_L(\cdot|\vs;\vphi)\) to output the Listener's policy \(\hat{\vx}\sim\pi_L(\cdot|\vm,\sC)\), which is a valid approximation since \(\vs\) holds information from the message and candidates. Parallelly, the critic head \(v(\cdot\,;\vphi)\) receives the same scores \(\vs\) and estimates the expected cumulative reward, as detailed in \Cref{sec:meth-learn}.


\subsection{Learning Strategy} \label{sec:meth-learn}
As described at the start of \cref{sec:meth-nlg}, the agents can only transmit information via the communication channel, which has only one direction: from the Speaker to the Listener. Additionally, agents learn how to communicate following the RIAL protocol, where agents are independent and treat others as part of the environment. As such, we have a decentralized training scheme where the agents improve their own parameters solely by maximizing the game's reward, see~\eqref{eq:lg-rwd}.

To perform well and consistently when playing the NLG, the Speaker must learn how to utilize the vocabulary to distinctively encode each image into a message to obtain the highest expected reward possible. We use Reinforce~\citep{williams1992simple}, a policy gradient algorithm, to train the Speaker. Given a target image \(\vx\) and the corresponding Listener's action \(\hat{\vx}\), we have a loss, \(L_\text{S,A}\), to fit the actor's head and another one, \(L_\text{S,C}\), for the critic's head. We define,
\begin{equation*}
L_\text{S,A}(\vtheta) = -\sum_{t=1}^N\sg\left(R(\vx,\hat{\vx})-v\left(\vz_{t,\vtheta};\vtheta\right)\right)\cdot\log\pi_S\left(m_t|\vz_{t,\vtheta};\vtheta\right),
\end{equation*}
where \(\sg\left(\cdot\right)\) is the \textit{stop-gradient} function, in order to optimize the policy. Note that the Speaker is in a sparse reward setting~\citep{10.5555/3042573.3042686}, where the sum of returns is the same as the game reward \(R\left(\vx,\hat{\vx}\right)\). Further, we subtract a baseline (critic head's value \(v\left(\vz_{t,\vtheta}\right)\)) from the returns to reduce variance. Regarding the critic loss, we devise
\begin{equation*}
L_\text{S,C}(\vtheta)=\sum_{t=1}^{N}\left(R(\vx, \hat{\vx}) - v(\vz_{t,\vtheta};\vtheta)\right)^2,
\end{equation*}
to approximate the state-value function \(V(\vx)=\E_{\pi_S}\left[R(\vx,\hat{\vx})\right]\).

We also use an additional entropy regularization term, \(L_{\text{S},\mathcal{H}}\), to make sure the language learned by the Speaker will not entirely stagnate by encouraging new combinations of tokens that increase entropy, further incentivizing exploration. Moreover, we define a target policy for the Speaker to minimize an additional KL divergent term, \(L_\text{S,KL}\), between the online and target policies, \(\vtheta\) and \(\bar{\vtheta}\), respectively. We update \(\bar{\vtheta}\) using an exponential moving average (EMA) over \(\vtheta\), \ie \(\bar{\vtheta}\leftarrow (1-\eta)\vtheta + \eta\bar{\vtheta}\) where \(\eta\) is the EMA weight parameter. With \(L_\text{S,KL}\), we prevent steep changes in the parameter space, which helps stabilize training~\citep{rawlik2012stochastic,chane2021goal}. We refer to~\citet{chaabouni2022emergent} for a complete analysis on the impact of \(L_\text{S,KL}\). Finally, we weigh each loss term and average the resulting sum given a batch of input images, \(\sX'\subset\sX\), to obtain the overall Speaker loss:
\begin{equation*}
\begin{split}
L_\text{S}(\vtheta) = \frac{1}{|\sX'|} \sum_{\vx\in\sX'} &\alpha_\text{S,A} L_\text{S,A}(\vtheta) + \alpha_\text{S,C} L_\text{S,C}(\vtheta) \\
& + \alpha_{\text{S},\mathcal{H}} L_{\text{S},\mathcal{H}}(\vtheta) + \alpha_\text{S,KL} L_\text{S,KL}(\vtheta),
\end{split}
\end{equation*}
%
where \(\alpha_\text{S,A}\), \(\alpha_\text{S,C}\), \(\alpha_{\text{S},\mathcal{H}}\), \(\alpha_\text{S,KL}\) are constants.

We also use Reinforce~\citep{williams1992simple} to train the Listener. We define the loss \(L_\text{L,A}\) to train the Listener's policy:
\begin{equation*}
L_\text{L,A}(\vphi)=-\sg\left(R(\vx,\hat{\vx})-v(\vs;\vphi)\right)\cdot\log\pi_L(\hat{\vx}|\vs;\vphi),
\end{equation*}
where cumulative returns is again the game reward \(R(\vx,\hat{\vx})\) since the Listener is in a single-step episode format where the game ends after choosing a candidate, \(\hat{\vx}\in\sC\). Identically to the Speaker, we subtract the Listener critic's value \(v(s;\vphi)\) from the game reward. The critic sub-network optimizes
\begin{equation*}
L_\text{L,C}(\vphi)=\left(R(\vx,\hat{\vx})-v(s;\vphi)\right)^2.
\end{equation*}
Similarly to the Speaker loss, we add an entropy loss term \(L_{\text{L},\mathcal{H}}(\vphi)\) to encourage exploration. The final Listener loss for a batch of images \(\sX'\) is:
\begin{equation*}
L_\text{L}(\vphi) = \frac{1}{|\sX'|} \sum_{x\in\sX'} \alpha_\text{L,A} L_\text{L,A}(\vphi) + \alpha_\text{L,C} L_\text{L,C}(\vphi) + \alpha_{\text{L},\mathcal{H}} L_{L,\mathcal{H}}(\vphi),
\end{equation*}
%
where \(\alpha_\text{L,A}\), \(\alpha_\text{L,C}\), and \(\alpha_{\text{L},\mathcal{H}}\) are constants.

A detailed analysis of the learning strategy, for both agents, can be found in \Cref{app:learn-strat}. Additionally, due to the complexity and non-stationarity of NLG, we define a scheduler for the noise level in the communication channel, during training. Namely, we linearly increase the noise level from \(0\) to \(\lambda\) at the beginning of training. This phase is optional and only helps with data efficiency (we refer to \Cref{app:noise-schedule} for more details).



\section{Evaluation} \label{sec:eval}

We provide an extensive evaluation of NLG and variants. For completeness, we also consider the original architecture proposed by \citet{chaabouni2022emergent} and our novel agent architecure to play the original LG (without message noise) as baselines. In this game variant, our model surpasses the original architecture at a slight cost of data efficiency. This trade-off is expected and fully explained in \cref{sec:eval-comm}. At a glance, this happens because the baseline version can retrieve more information than our implementation, during training. Having a progressive sequence of LG variants enables us to assess how each modification influences the emergent communication protocol learned by the agents.

We continue this section by introducing all LG variants, giving a broader view of each game, agent architectures, and learning strategy. Next, we evaluate the generality of the emerging language for each game variant when providing new and unseen images. We compare LG and NLG variants when testing with and without noise in the communication channel. Additionally, we investigate how the candidate set, \(\sC\), impacts the generalization capabilities of the message protocols. Moreover, we investigate the internal message structure to understand how robust communication protocols emerge when agents play in the NLG. Finally, we perform an out-of-distribution evaluation to discern how each variant reacts to novel forms of noise. We always report results using the average (plus SD) over 10 different seeds.

Due to space constraints, \Cref{app:add-results} contains the results obtained in all experiments, for both ImageNet~\citep{ILSVRC15} and CelebA~\citep{liu2015faceattributes} datastes, used to devise the analyses detailed in this section. Additionally, we report a supplementary evaluation to assess the capacity of each game variant to adapt to new tasks in a transfer learning manner, called \emph{ease and transfer learning} (ETL)~\citep{chaabouni2022emergent} (see \Cref{app:etl}). This supplementary evaluation gives yet another frame of reference to evaluate the generality and robustness of the learned languages. Finally, we also refer to the appendix for further details regarding related work (\Cref{app:rw}), every game variant (\Cref{app:lg}), model architectures (\Cref{app:arch}), and datasets used (\Cref{app:data}).


\subsection{Lewis Game Variants} \label{sec:eval-variants}
We briefly report essential aspects of each game variant, while referring to supplementary information when necessary. We consider three variants of the LG, all of which share the same Speaker architecture. The Listener architecture differs in all games. We refer to \Cref{app:arch} for a detailed description of the implementation of these architectures. Additionally, all variants except for LG (S) are a contribution of this work. The LG variants considered are:

\begin{itemize}
    \item LG (S): Original LG variant introduced in~\citet{chaabouni2022emergent}. The Listener trains under supervised data by using InfoNCE loss~\citep{oord2018representation,dessi2021interpretable} to find similarities between the message and the correct candidate, see \Cref{app:list-arch}.
    \item LG (RL): LG with a deterministic communication channel (no noise) where both agents implement RL architectures. The only semi-supervised information given to both agents is whether the game ended successfully or not. We refer to \Cref{sec:meth-arch,sec:meth-learn} for a comprehensive description of the agents' architectures and learning strategies, respectively.
    \item NLG: LG variant introduced in \Cref{sec:meth-nlg}, where we apply an external environmental pressure by adding noise to the message during transmission. Both agents function as RL agents, as in LG (RL). Agents' architectures and learning strategies appear in \Cref{sec:meth-arch,sec:meth-learn}, respectively. For an overall understanding of NLG, we define 3 different versions: NLG (0.25), NLG (0.5), and NLG (0.75); where NLG (\(x\)) means that, during training, we fix the noise threshold at \(\lambda=x\).
\end{itemize}

\subsection{Robust Communication Protocols} \label{sec:eval-comm}

This section analyzes the performance of all LG variants described above. Since there is the possibility to apply different hyper-parameters depending on the current phase (training or testing phase), we define two extra variables, \(\lambda_\text{test}\) and \(\sC_\text{test}\), to define the noise threshold and candidate set applied during the test phase, respectively.

\begin{figure}[!t]
  \centering
  \includegraphics[width=0.71\linewidth]{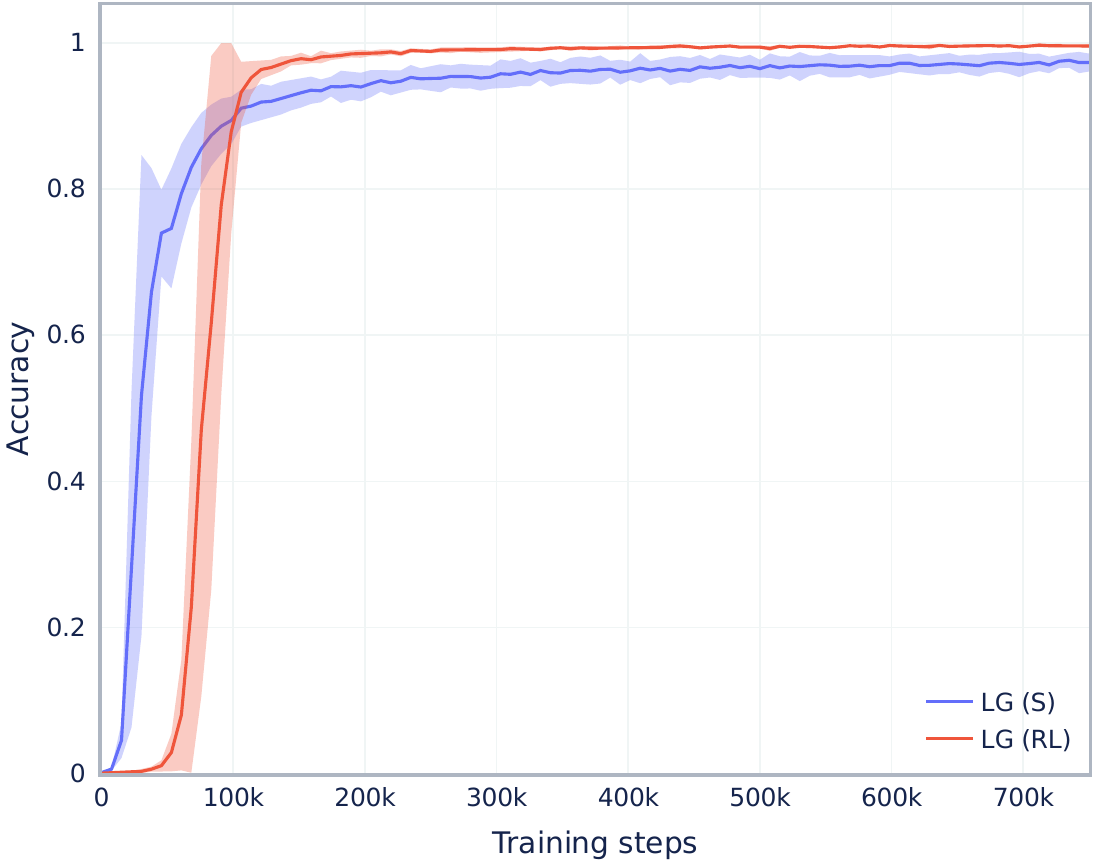}
  \caption{Training accuracy for LG (S) and LG (RL). Trained on ImageNet dataset and  \(|\sC|=1024\).}
  \label{fig:compare-lg}
\end{figure}

Starting by comparing LG (S) with LG (RL), we can see an apparent performance boost for the LG when the Listener is an RL agent. \Cref{fig:compare-lg} shows that, during training, the RL version performs better than the supervised version. Equivalent results occur in the testing phase. From \Cref{fig:compare-lg1,fig:compare-lg2}, and focusing on the results obtained with a deterministic communication channel, \(\lambda_\text{test}=0\), the RL version surpasses the accuracy achieved by the supervised counterpart. This performance gap becomes more predominant as we increase game complexity, as seen in \Cref{sec:eval-cand}. From \Cref{fig:compare-lg}, we also observe a trade-off between performance and sample efficiency, where the RL version is less sample efficient. We can trace these differences back to the loss function employed by each version. For instance, the supervised version employs the InfoNCE loss (\Cref{app:arch-list-ss}), which we can see as a Reinforce variant with only a policy to optimize and, particularly, with access to an oracle giving information about which action (candidate) is the right one for each received message. As such, the Listener (S) can efficiently learn how to map messages to the correct candidates. On the other hand, the RL version has no access to such oracle and needs to interact with the environment to build this knowledge. The decrease in sample efficiency from supervised to RL is, therefore, a natural phenomenon. Nonetheless, the RL version introduces a critic loss term whose synergy with the policy loss term helps to improve the final performance when compared to the supervised version.

One disadvantage of employing, at inference time, the communication protocols learned by playing default LG variants (LG (S) and LG (RL)) is that they are not robust to deal with message perturbations. Since agents train only with perfect communication, they never experience noisy communication. When testing the performance of LG (S) and LG (RL) with noisy communication channels \(\lambda_\text{test}>0\),  we observe a noticeable dominance of RL against S. Nonetheless, there is a massive drop in performance for both variants compared to the noiseless case \(\lambda_\text{test}=0\), see \Cref{fig:compare-lg1,fig:compare-lg2}.

Conversely, NLG puts agents in a more complex environment where only random fractions of the message are visible during training time. Despite such modifications, the pair of agents can still adapt to the environment and learn robust communication protocols that handle both types of messages (with and without noise). We notice equivalent accuracy performance for NLG and LG (RL) when testing with deterministic communication channels, see \Cref{fig:compare-lg1,fig:compare-lg2}. Notably, every NLG version only suffers a negligible performance loss when testing with \(\lambda_\text{test}=0.25\). This loss starts to be more noticeable at higher noise levels, where the accuracy drops to around 80\%, and further to the interval between 30\%-40\% when \(\lambda_\text{test}\) is \(0.5\) and \(0.75\), respectively. Still, NLG is considerably more effective than LG (S) and LG (RL) when communicating in noisy environments, as seen in the considerable performance gap visible in each tested noise level \(\lambda_\text{test}\). This increased performance suggests that, in NLG, agents can encode redundant information where communication is still functional when random parts of the message are hidden. Additionally, the performance obtained for each test threshold \(\lambda_\text{test}\) is similar for every NLG version. As such, each version displays similar capacities to handle noise in the communication channel, independent of the noise threshold \(\lambda\) applied during training. Please refer to \Cref{app:add-results} for additional results on the ImageNet and CelebA datasets.

\subsubsection{Comparing Different Noise Levels}
\begin{figure*}[!t]
  \begin{minipage}[!t]{1\textwidth}
  \centering
  \includegraphics[width=0.55\linewidth]{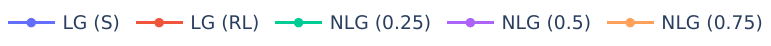}
  \caption*{}
  \end{minipage}
  \begin{minipage}[!t]{0.49\textwidth}
  \vspace{-7ex}
  \centering
  \includegraphics[width=0.81\linewidth]{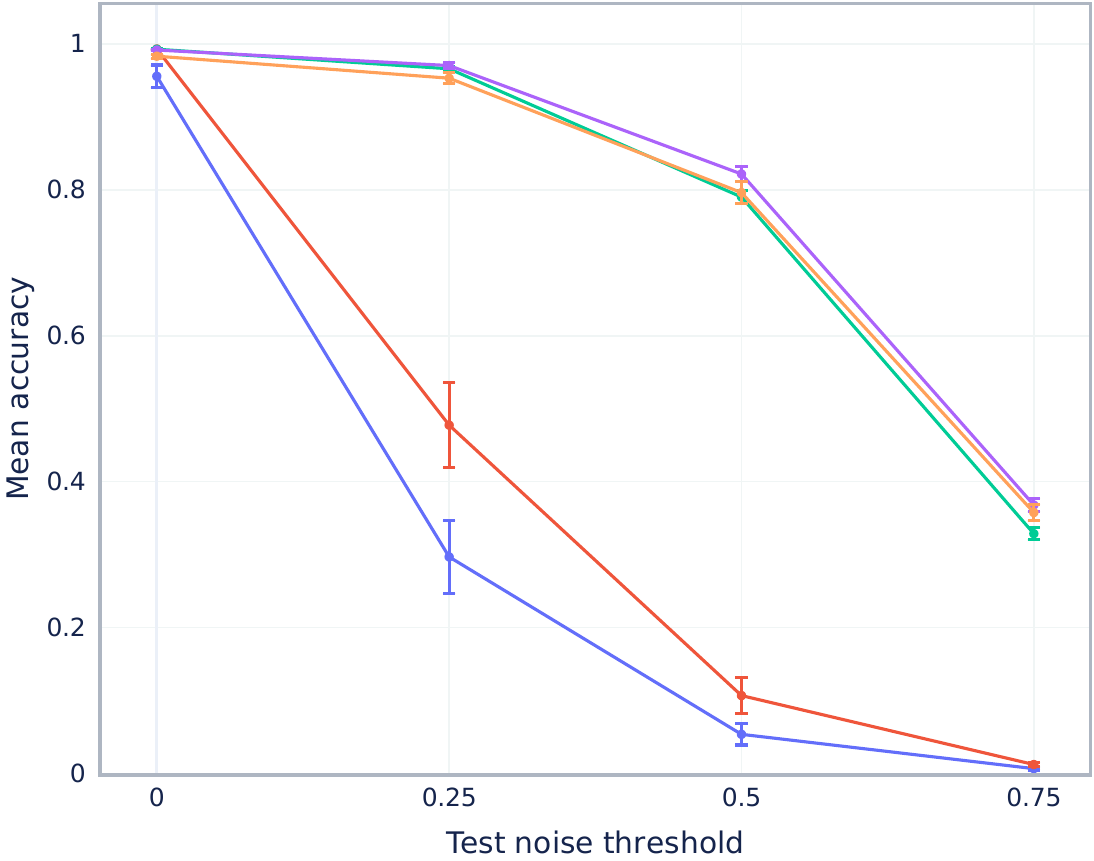}
  \caption{Test accuracy for all variants with \(|\sC_\text{test}|=1024\). Trained on ImageNet and \(|\sC|=1024\). The \(x\)-axis denotes \(\lambda_\text{test}\).}
  \label{fig:compare-lg1}
  \end{minipage}
  \hfill
  \begin{minipage}[!t]{0.49\textwidth}
  \vspace{-7ex}
  \centering
  \includegraphics[width=0.81\linewidth]{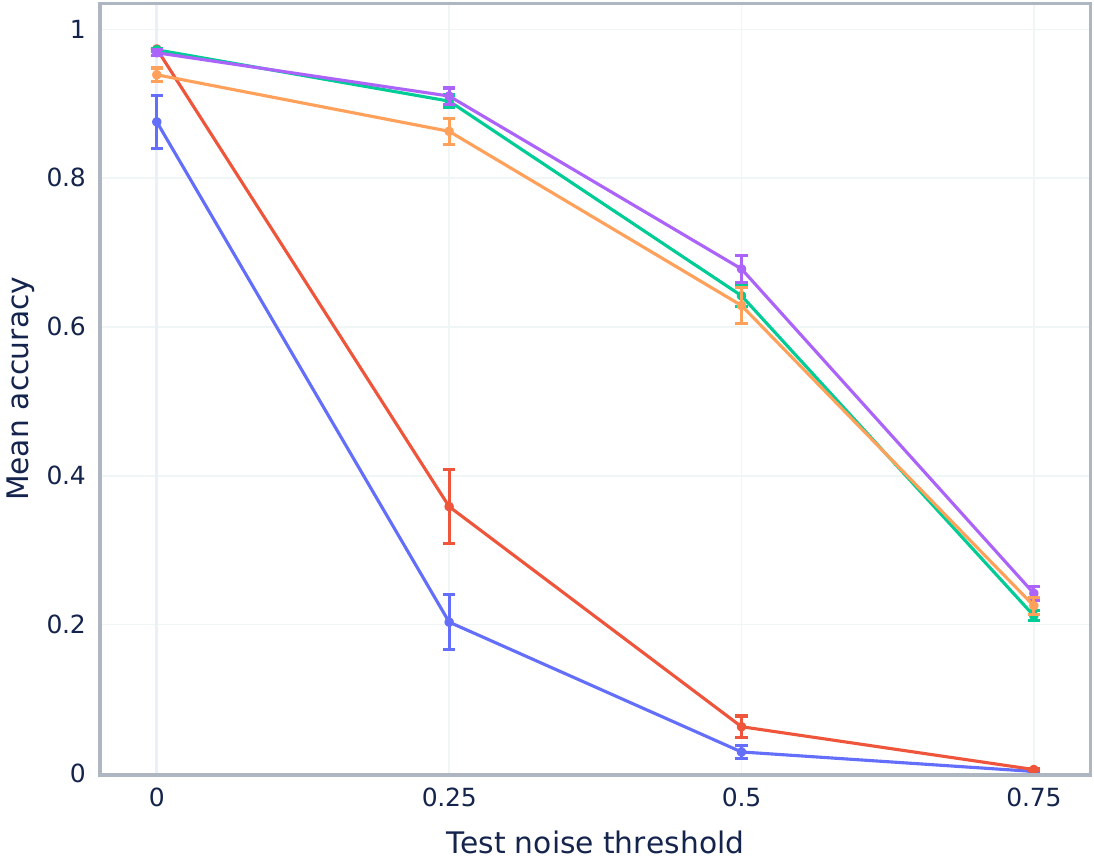}
  \caption{Test accuracy for all variants with \(|\sC_\text{test}|=4096\). Trained on ImageNet and \(|\sC|=1024\). The \(x\)-axis denotes \(\lambda_\text{test}\).}
  \label{fig:compare-lg2}
  \end{minipage}
\end{figure*}

Comparing the different variants of NLG, we observe that the mean accuracy obtained for NLG (0.75) is slightly lower than NLG (0.25) and NLG (0.5) when we set \(\lambda_\text{test}\) to \(0\) or \(0.25\), see \Cref{fig:compare-lg1,fig:compare-lg2}. When \(\lambda_\text{test}=0.5\), NLG (0.5) performs slightly better than its counterparts. Finally, NLG (0.5) and NLG (0.75) seem to perform slightly better than NLG (0.25) is the extreme noise case (\(\lambda_\text{test}\) is \(0.75\)).

Henceforth, having a threshold of \(\lambda=0.5\) during training appears to give a good balance for the pair of agents to develop a communication protocol that can effectively act in a broad range of noise levels, even when there is no noise during communication.

\subsubsection{Scaling the Number of Candidates} \label{sec:eval-cand}

We train every game variant with different candidate sizes \(|\sC|\). We scale \(|\sC|\) from 16 to 1024, using a ratio of 4. At test time, we evaluate all experiments using larger sizes, for the candidates set, to inspect generalization capabilities. In our case, we use \(|\sC_\text{test}|=1024\) and \(|\sC_\text{test}|=4096\). Looking at \cref{table:lg_imagenet_0_test_cand,table:lg_imagenet_noise_test_cand}, we can see an evident generalization boost when the number of candidates increases for every game. We posit that increasing the game's difficulty (increasing the number of candidates) helps the agents to generalize. As the candidates' set gets additional images, the input diversity increases, which affects how agents encode and interpret more information to distinguish the correct image from all others. Even when adding noise, the agents can quickly adapt to such changes conveying information in such a way as to repress the noise mechanism. We argue that agents have two correlated ways to achieve such adaptation:
\begin{enumerate*}
  \item send redundant information regarding specific features of the image; and
  \item create a spatial mapping from the image to the message space.
\end{enumerate*}

We note that as \(|\sC|\) increases, the test performance also increases, but at a smaller scale. As an example, consider LG (RL) variant when \(\lambda_\text{test}=0\) and \(|\sC_\text{test}|=4096\). In this case, the test performance gap between \(|\sC|=16\) and \(|\sC|=64\) is \(0.4\), and only \(0.03\) between \(|\sC|=256\) and \(|\sC|=1024\), see \Cref{table:lg_imagenet_0_test_cand}. Regarding NLG, the accuracy starts lower for smaller candidate sizes, e.g., \(0.27\) when \(|\sC|=16\), against \(0.67\) for the LG (RL) counterpart. Nonetheless, as the candidate set size increases, the noise effect becomes less predominant and the NLG's performance reaches the same level as in LG (RL), both achieving an accuracy of \(0.97\) when \(|\sC|=1024\).


Equivalently to the results introduced in the beginning of \Cref{sec:eval-comm}, we observe low accuracy for LG (RL) when testing with noisy communication channels (\Cref{table:lg_imagenet_noise_test_cand}), regardless of the candidate size. In respect to NLG, there is an apparent increase in performance as \(|\sC|\) increases during training. Additionally, the performance gap between consecutive candidate set sizes is approximately the same (between \(0.14\) and \(0.21\), when \(|\sC_\text{test}|=4096\)).


\begin{table*}[t]
\linespread{0.6}\selectfont\centering

\begin{minipage}[t]{0.48\linewidth}
\centering
\caption{Test accuracy including SD, when \(\lambda_{\text{test}}=0\), for LG (RL) and NLG, on ImageNet dataset.}
\label{table:lg_imagenet_0_test_cand}
\begin{tabular}{lrrrr}
\toprule
Game & \(\lambda\) & \(|\sC|\) & \multicolumn{2}{c}{\(|\sC|\) (test)} \\[1ex]\cmidrule(r){4-5}
 &  &  & \multicolumn{1}{c}{\(1024\)} & \multicolumn{1}{c}{\(4096\)} \\
\midrule
LG {\scriptsize(RL)} & 0 & 16 & \longcell{0.67\\{\tiny(0.04)}} & \longcell{0.39\\{\tiny(0.04)}} \\[2.2ex]
LG {\scriptsize(RL)} & 0 & 64 & \longcell{0.93\\{\tiny(0.01)}} & \longcell{0.79\\{\tiny(0.03)}} \\[2.2ex]
LG {\scriptsize(RL)} & 0 & 256 & \longcell{0.98\\{\tiny(0.00)}} & \longcell{0.94\\{\tiny(0.01)}} \\[2.2ex]
LG {\scriptsize(RL)} & 0 & 1024 & \longcell{0.99\\{\tiny(0.00)}} & \longcell{0.97\\{\tiny(0.00)}} \\[2.2ex]
NLG & 0.5 & 16 & \longcell{0.55\\{\tiny(0.03)}} & \longcell{0.27\\{\tiny(0.02)}} \\[2.2ex]
NLG & 0.5 & 64 & \longcell{0.87\\{\tiny(0.01)}} & \longcell{0.67\\{\tiny(0.03)}} \\[2.2ex]
NLG & 0.5 & 256 & \longcell{0.98\\{\tiny(0.00)}} & \longcell{0.91\\{\tiny(0.01)}} \\[2.2ex]
NLG & 0.5 & 1024 & \longcell{0.99\\{\tiny(0.00)}} & \longcell{0.97\\{\tiny(0.00)}} \\[2.2ex]
\bottomrule
\end{tabular}
\end{minipage}
\hfill
\begin{minipage}[t]{0.48\linewidth}
    
\centering
\caption{Test accuracy including SD, when \(\lambda_{\text{test}}=0.5\), for LG (RL) and NLG, on ImageNet dataset.}
\label{table:lg_imagenet_noise_test_cand}
\begin{tabular}{lrrrr}
\toprule
Game & \(\lambda\) & \(|\sC|\) & \multicolumn{2}{c}{\(|\sC|\) (test)} \\[1ex]\cmidrule(r){4-5}
 &  &  & \multicolumn{1}{c}{\(1024\)} & \multicolumn{1}{c}{\(4096\)} \\
\midrule
LG {\scriptsize(RL)} & 0 & 16 & \longcell{0.03\\{\tiny(0.01)}} & \longcell{0.01\\{\tiny(0.01)}} \\[2.2ex]
LG {\scriptsize(RL)} & 0 & 64 & \longcell{0.09\\{\tiny(0.07)}} & \longcell{0.05\\{\tiny(0.03)}} \\[2.2ex]
LG {\scriptsize(RL)} & 0 & 256 & \longcell{0.11\\{\tiny(0.06)}} & \longcell{0.06\\{\tiny(0.04)}} \\[2.2ex]
LG {\scriptsize(RL)} & 0 & 1024 & \longcell{0.11\\{\tiny(0.02)}} & \longcell{0.06\\{\tiny(0.01)}} \\[2.2ex]
NLG & 0.5 & 16 & \longcell{0.32\\{\tiny(0.02)}} & \longcell{0.14\\{\tiny(0.01)}} \\[2.2ex]
NLG & 0.5 & 64 & \longcell{0.57\\{\tiny(0.02)}} & \longcell{0.33\\{\tiny(0.02)}} \\[2.2ex]
NLG & 0.5 & 256 & \longcell{0.73\\{\tiny(0.01)}} & \longcell{0.54\\{\tiny(0.02)}} \\[2.2ex]
NLG & 0.5 & 1024 & \longcell{0.82\\{\tiny(0.01)}} & \longcell{0.68\\{\tiny(0.02)}} \\[2.2ex]
\bottomrule
\end{tabular}
\end{minipage}
\vspace{-2ex}
\end{table*}

\begin{figure*}[!t]
  \begin{minipage}[!t]{1\textwidth}
  \centering
  \includegraphics[width=0.55\linewidth]{figures/legend}
  \caption*{}
  \end{minipage}
  \begin{minipage}[!t]{0.49\textwidth}
  \vspace{-7ex}
  \centering
  \includegraphics[width=0.81\linewidth]{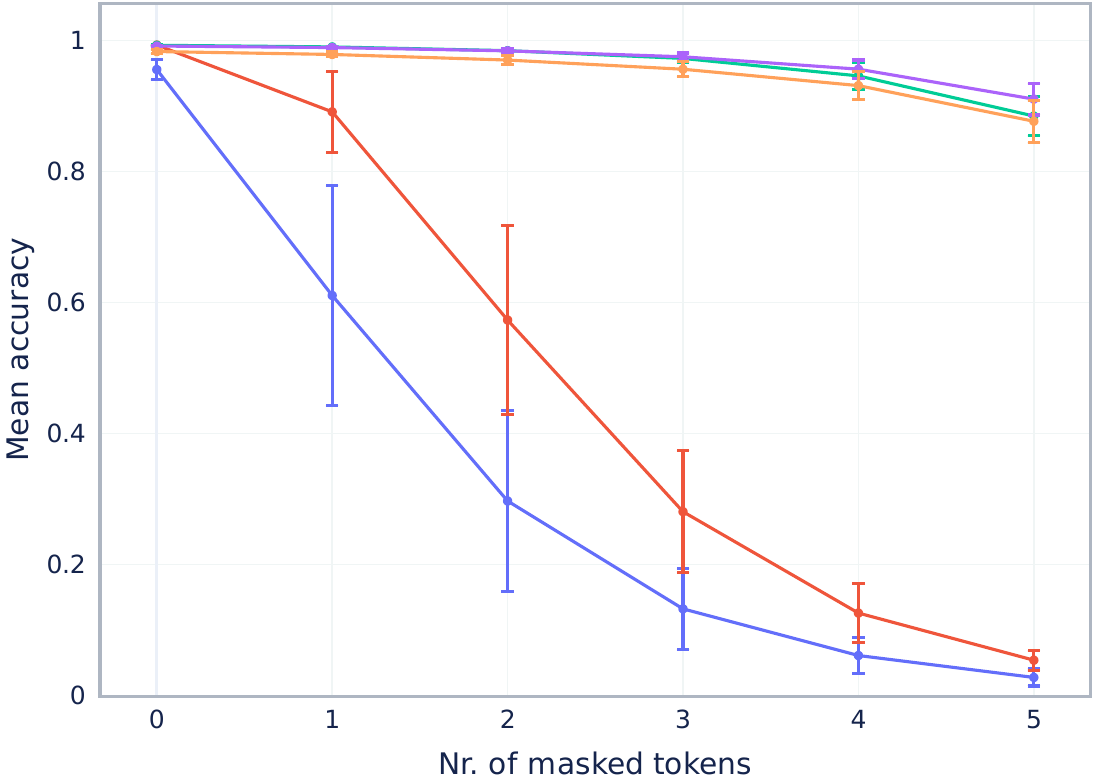}
  \caption{Mean test accuracy for all LG variants, trained with \(|\sC|=1024\) and on ImageNet dataset. At test time, \(|\sC_\text{test}|=1024\). We report 10 different combinations of masked tokens.}
  \label{fig:compare-lg-mess-1024}
  \end{minipage}
  \hfill
  \begin{minipage}[!t]{0.49\textwidth}
    \vspace{-7ex}
  \centering
  \includegraphics[width=0.81\linewidth]{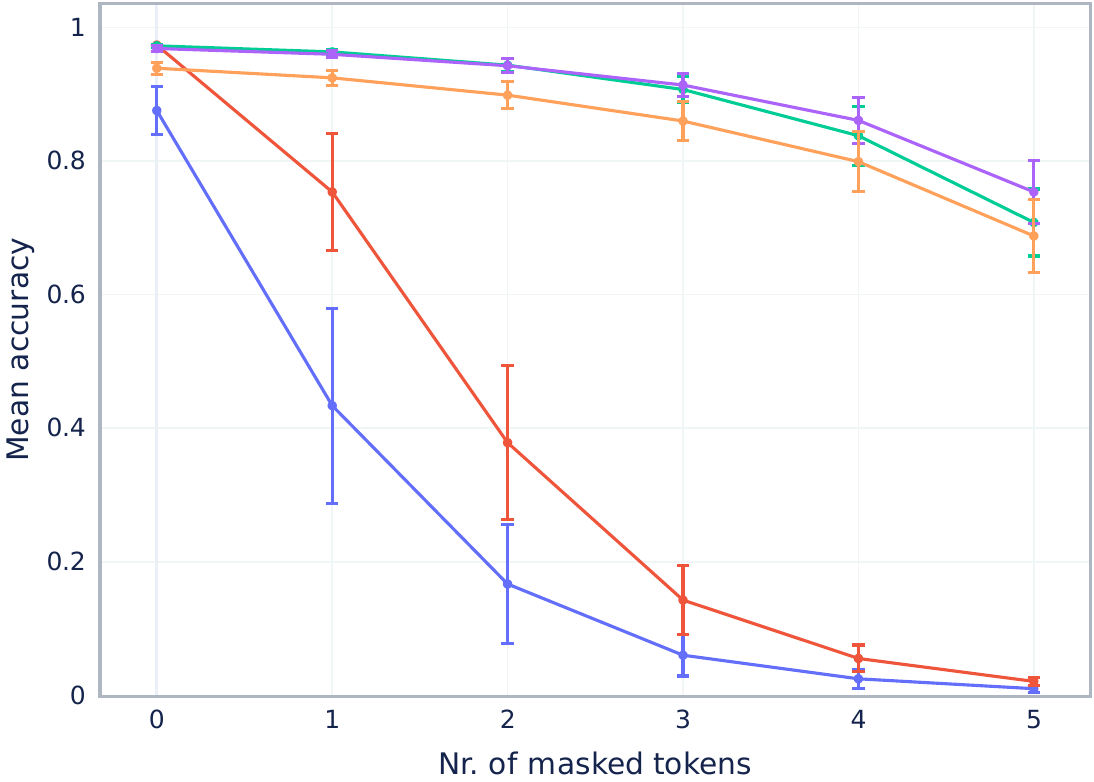}
  \caption{Mean test accuracy for all LG variants, trained with \(|\sC|=1024\) and on ImageNet dataset. At test time, \(|\sC_\text{test}|=4096\). We report 10 different combinations of masked tokens.}
  \label{fig:compare-lg-mess-4096}
  \end{minipage}
\end{figure*}

\begin{figure*}[!t]
  \begin{minipage}[!t]{1\textwidth}
  \centering
  \includegraphics[width=0.55\linewidth]{figures/legend}
  \caption*{}
  \end{minipage}
  \begin{minipage}{0.49\textwidth}
  \vspace{-7ex}
  \centering
  \includegraphics[width=0.81\linewidth]{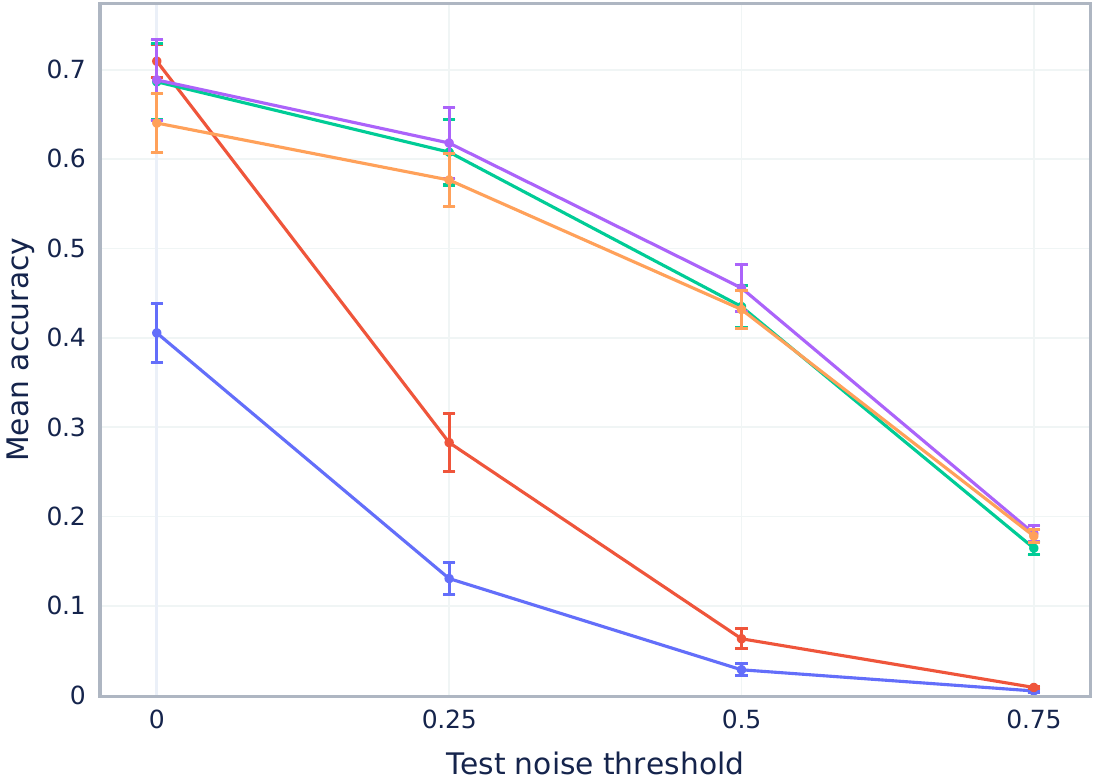}
  \caption{Mean test accuracy for all variants with \(|\sC_\text{test}|=1024\). Trained on ImageNet and \(|\sC|=1024\). During test noise is added to the inputs (\Cref{sec:eval-ext-noise}). The \(x\)-axis denotes \(\lambda_\text{test}\).}
  \label{fig:compare-lg-noisy-input-1024}
  \end{minipage}
  \hfill
  \begin{minipage}{0.49\textwidth}
  \vspace{-7ex}
  \centering
  \includegraphics[width=0.81\linewidth]{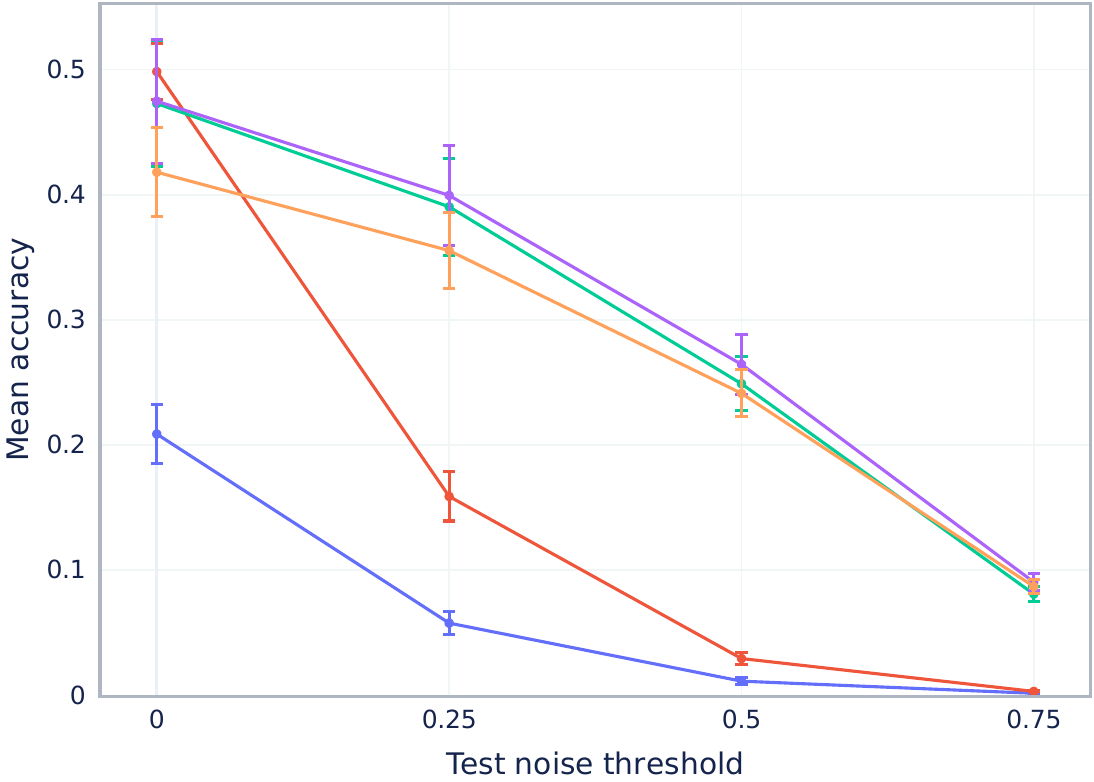}
  \caption{Mean test accuracy for all variants with \(|\sC_\text{test}|=4096\). Trained on ImageNet and \(|\sC|=1024\). During test noise is added to the inputs (\Cref{sec:eval-ext-noise}). The \(x\)-axis denotes \(\lambda_\text{test}\).}
  \label{fig:compare-lg-noisy-input-4096}
  \end{minipage}
\end{figure*}

\subsection{Message Structure Analysis} \label{sec:eval-mess-struct}

With the aim of addressing how NLG variations develop robust communication protocols, we propose to analyze the message structure of the language protocols. The NLG includes additional adverse pressures where the communication channel seems unreliable. In order to overcome the noise introduced by the faulty channel, the pair of agents must find alternative ways to coordinate how to send information. Moreover, since the only feedback received by the Speaker is the game's outcome, the new coordination mechanism becomes essential to complete the game with a high success rate. The most intuitive behavior for the Speaker focuses on developing implicit repair mechanisms where messages incorporate redundant information. As such, even if a subset of the message tokens become masked, the Listener can still parse the remaining message and select the correct candidate, making the communication robust to the noise being introduced by the faulty channel.

To test our assumption, we indirectly analyze the internal structure of the message being transmitted in all games. Since we want to examine the existence of redundancy in the communication protocol, we propose an evaluation where we iteratively increase the number of masked tokens and retrieve the performance obtained. The range of masked tokens spreads from \(0\) to \(N/2=5\) (half of the tokens). Additionally, when selecting the number of tokens to conceal, we evaluate the performance over ten different combinations of masked tokens (except when no tokens are masked), ensuring the results represent the average case. As such, this analysis allows us to examine, in greater detail, how the accuracy changes as we increase the number of masked tokens iteratively. As shown in \Cref{fig:compare-lg-mess-1024,fig:compare-lg-mess-4096}, the NLG variations only decrease slightly in accuracy as the number of masked tokens increases, conveying that our assumption is accurate and messages contain redundant information. Consequently, the Listener still contains enough information to select the right candidate, even with partial messages. On the other hand, for the deterministic variations, LG (S) and LG (RL), we observe a faster decrease in performance as more tokens are masked, conveying that every token is essential for the Listener to infer the right candidate. Furthermore, we highlight that the average accuracy obtained by the NLG (0.5) variant, with five tokens masked, is similar to the performance obtained by LG (RL) when masking a single token, which is around 75\% (\Cref{fig:compare-lg-mess-4096}). This particular result illustrates how much noise both variants can manage for a particular and similar performance level. As such, NLG (0.5) can deal with \texttt{5x} more noise than LG (RL) when discriminating the most amount of images, \(|\sC_\text{test}|=4096\). Please refer to \Cref{app:add-results} for additional results on the ImageNet and CelebA datasets.

In addition to the results presented above, we observe an interesting occurrence when communication protocols emerge with deterministic channels, as in the case of LG (S) and LG (RL). Our analysis shows that when masking a single token, the first token of the message seems to carry more information than the others, or, at least, is crucial to derive meaning from the rest of the message. We show these results in the \Cref{app:add-results}, where the drop in performance can, in some cases, drop down around 10\% when the first token is masked in opposition to mask any other token. The drop in performance is more noticeable as the number of candidates given to the Listener increases, where it seems the first token plays a crucial role to reduce the number of final choices to consider.

\subsection{External Noise Interference} \label{sec:eval-ext-noise}

In this experiment, we test the capability of the emergent communication protocols trained in the games presented above to adapt to new game variations where we focus on adding noise to conceal information in other components of the input. In this regard, we model a new noise dynamic by adding random information to the objects to discriminate:
\begin{enumerate*}
  \item the target image provided to the Speaker, and
  \item the candidates given to the Listener.
\end{enumerate*}
We sample noise from a Gaussian distribution and add it to the output of the frozen image encoder. As such, in this game adaptation, we modify the target image to \(\vx'=f(\vx)+\vvarepsilon\) and the candidates to \(\vx_i=f(\vx_i)+\vvarepsilon\), where \(\vvarepsilon\sim\mathcal{N}(\vzero,\sigma^2\mI)\) and \(\vx_i\in\sC\). Unless otherwise noted, we set \(\sigma=1\).

We can view this new modification as an out-of-distribution task since, during training, agents only deal with noise in the communication channel, where the obstruction happens by masking some of the message tokens. \Cref{fig:compare-lg-noisy-input-1024,fig:compare-lg-noisy-input-4096} depict the results obtained for all game variants at inference time when we introduce noise to the input images. When testing with a deterministic channel (\(\lambda_\text{test}=0\)), LG (RL) and all NLG versions have similar performance, where the mean accuracy is around \(0.7\) and \(0.5\) for 1024 and 4096 candidates, respectively. We can see a slight degradation in performance across all variants compared to the previous experiment setting in \Cref{sec:eval-comm} (images without any perturbations). Nonetheless, these results indicate a positive transfer capability where the language protocols are general enough, allowing effective agent reasoning even with partially hidden input distributions. Additionally, we notice that adding noise in the communication channel during training (NLG variant) does not provide improved benefits to the communication protocol to deal with other noise types, as there is no gain in performance when \(\lambda_\text{test}=0\), comparing against LG (RL).

For the experiments where noise is also present in the communication channel, \(\lambda_\text{test}>0\), we observe similar results as in \Cref{sec:eval-comm}, where NLG versions vastly surpass both LG (RL) and LG (S) since the former emergent protocols can efficiently retrieve applicable information to reason about, from the noisy messages. Finally, the results obtained by LG (S) were considerably lower than all other variants, again claiming the superiority of having a Listener as an RL agent, see \Cref{sec:eval-comm}. Please refer to \Cref{app:add-results} for additional results on the ImageNet and CelebA datasets.



\section{Conclusion \& Future Work}

In this work, we focus on designing agent systems that can learn language protocols without prior knowledge, where communication evolves grounded on experience to solve the task at hand. We explore EC from a language evolution perspective to analyze a particular linguistic concept called conversational repair. Conversational repair appears in human languages as a mechanism to detect and resolve miscommunication and misinformation during social interactions. Mainly, we focus on the implicit repair mechanism, where the interlocutor sending the information deliberately communicates in such a way as to prevent misinformation and avoid future interactions on correcting it. Our analysis shows that implicit conversational repair can also emerge in artificial designs when there is enough disruptive environmental pressure, where sending redundant information facilitates solving the task more effectively.

For future work, several ideas can be explored. One possible research idea passes to merge explicit and implicit repair mechanisms. In this scenario, agents would have to coordinate between using implicit mechanisms to prevent misinformation or starting a posterior dialogue (an explicit mechanism) when the Listener cannot extract useful information from the message sent. We argue that this coordination mechanism can emerge naturally in sufficiently complex environments, where the Listener develops the capacity to leverage the likelihood of success and the cost of playing more communication rounds. Another exciting research direction focuses on developing a universal repair mechanism system with the objective of merging various language protocols, grounded in different tasks, into a universal and general language protocol, focusing on the capability to be used by new agents and in out-of-distribution tasks.

\begin{acks}
This work was supported by national funds through Fundação para a Ciência e a Tecnologia (FCT) with reference UIDB/50021/2020 - DOI: 10.54499/UIDB/50021/2020, project Center for Responsible AI with reference C628696807-00454142, and project RELEvaNT PTDC/CCI-COM/5060/2021. The first author acknowledges the FCT PhD grant 2022.14163.BD.
\end{acks}



\bibliographystyle{ACM-Reference-Format} 
\balance
\bibliography{refs}


\begin{thebibliography}{56}


\ifx \showCODEN    \undefined \def \showCODEN     #1{\unskip}     \fi
\ifx \showDOI      \undefined \def \showDOI       #1{#1}\fi
\ifx \showISBNx    \undefined \def \showISBNx     #1{\unskip}     \fi
\ifx \showISBNxiii \undefined \def \showISBNxiii  #1{\unskip}     \fi
\ifx \showISSN     \undefined \def \showISSN      #1{\unskip}     \fi
\ifx \showLCCN     \undefined \def \showLCCN      #1{\unskip}     \fi
\ifx \shownote     \undefined \def \shownote      #1{#1}          \fi
\ifx \showarticletitle \undefined \def \showarticletitle #1{#1}   \fi
\ifx \showURL      \undefined \def \showURL       {\relax}        \fi
\providecommand\bibfield[2]{#2}
\providecommand\bibinfo[2]{#2}
\providecommand\natexlab[1]{#1}
\providecommand\showeprint[2][]{arXiv:#2}

\bibitem[\protect\citeauthoryear{Albert and De~Ruiter}{Albert and
  De~Ruiter}{2018}]%
        {albert2018repair}
\bibfield{author}{\bibinfo{person}{Saul Albert} {and} \bibinfo{person}{Jan~P
  De~Ruiter}.} \bibinfo{year}{2018}\natexlab{}.
\newblock \showarticletitle{Repair: the interface between interaction and
  cognition}.
\newblock \bibinfo{journal}{\emph{Topics in cognitive science}}
  \bibinfo{volume}{10}, \bibinfo{number}{2} (\bibinfo{year}{2018}),
  \bibinfo{pages}{279--313}.
\newblock


\bibitem[\protect\citeauthoryear{Bisk, Holtzman, Thomason, Andreas, Bengio,
  Chai, Lapata, Lazaridou, May, Nisnevich, et~al\mbox{.}}{Bisk
  et~al\mbox{.}}{2020}]%
        {bisk2020experience}
\bibfield{author}{\bibinfo{person}{Yonatan Bisk}, \bibinfo{person}{Ari
  Holtzman}, \bibinfo{person}{Jesse Thomason}, \bibinfo{person}{Jacob Andreas},
  \bibinfo{person}{Yoshua Bengio}, \bibinfo{person}{Joyce Chai},
  \bibinfo{person}{Mirella Lapata}, \bibinfo{person}{Angeliki Lazaridou},
  \bibinfo{person}{Jonathan May}, \bibinfo{person}{Aleksandr Nisnevich},
  {et~al\mbox{.}}} \bibinfo{year}{2020}\natexlab{}.
\newblock \showarticletitle{Experience grounds language}.
\newblock \bibinfo{journal}{\emph{arXiv preprint arXiv:2004.10151}}
  (\bibinfo{year}{2020}).
\newblock


\bibitem[\protect\citeauthoryear{Bogin, Geva, and Berant}{Bogin
  et~al\mbox{.}}{2018}]%
        {bogin2018emergence}
\bibfield{author}{\bibinfo{person}{Ben Bogin}, \bibinfo{person}{Mor Geva},
  {and} \bibinfo{person}{Jonathan Berant}.} \bibinfo{year}{2018}\natexlab{}.
\newblock \showarticletitle{Emergence of communication in an interactive world
  with consistent speakers}.
\newblock \bibinfo{journal}{\emph{arXiv preprint arXiv:1809.00549}}
  (\bibinfo{year}{2018}).
\newblock


\bibitem[\protect\citeauthoryear{Chaabouni, Kharitonov, Bouchacourt, Dupoux,
  and Baroni}{Chaabouni et~al\mbox{.}}{2020}]%
        {chaabouni-etal-2020-compositionality}
\bibfield{author}{\bibinfo{person}{Rahma Chaabouni}, \bibinfo{person}{Eugene
  Kharitonov}, \bibinfo{person}{Diane Bouchacourt}, \bibinfo{person}{Emmanuel
  Dupoux}, {and} \bibinfo{person}{Marco Baroni}.}
  \bibinfo{year}{2020}\natexlab{}.
\newblock \showarticletitle{Compositionality and Generalization In Emergent
  Languages}. In \bibinfo{booktitle}{\emph{Proceedings of the 58th Annual
  Meeting of the Association for Computational Linguistics}},
  \bibfield{editor}{\bibinfo{person}{Dan Jurafsky}, \bibinfo{person}{Joyce
  Chai}, \bibinfo{person}{Natalie Schluter}, {and} \bibinfo{person}{Joel
  Tetreault}} (Eds.). \bibinfo{publisher}{Association for Computational
  Linguistics}, \bibinfo{address}{Online}, \bibinfo{pages}{4427--4442}.
\newblock
\urldef\tempurl%
\url{https://doi.org/10.18653/v1/2020.acl-main.407}
\showDOI{\tempurl}


\bibitem[\protect\citeauthoryear{Chaabouni, Kharitonov, Dupoux, and
  Baroni}{Chaabouni et~al\mbox{.}}{2019}]%
        {chaabouni2019anti}
\bibfield{author}{\bibinfo{person}{Rahma Chaabouni}, \bibinfo{person}{Eugene
  Kharitonov}, \bibinfo{person}{Emmanuel Dupoux}, {and} \bibinfo{person}{Marco
  Baroni}.} \bibinfo{year}{2019}\natexlab{}.
\newblock \showarticletitle{Anti-efficient encoding in emergent communication}.
\newblock \bibinfo{journal}{\emph{Advances in Neural Information Processing
  Systems}}  \bibinfo{volume}{32} (\bibinfo{year}{2019}).
\newblock


\bibitem[\protect\citeauthoryear{Chaabouni, Strub, Altch{\'e}, Tarassov,
  Tallec, Davoodi, Mathewson, Tieleman, Lazaridou, and Piot}{Chaabouni
  et~al\mbox{.}}{2022}]%
        {chaabouni2022emergent}
\bibfield{author}{\bibinfo{person}{Rahma Chaabouni}, \bibinfo{person}{Florian
  Strub}, \bibinfo{person}{Florent Altch{\'e}}, \bibinfo{person}{Eugene
  Tarassov}, \bibinfo{person}{Corentin Tallec}, \bibinfo{person}{Elnaz
  Davoodi}, \bibinfo{person}{Kory~Wallace Mathewson}, \bibinfo{person}{Olivier
  Tieleman}, \bibinfo{person}{Angeliki Lazaridou}, {and} \bibinfo{person}{Bilal
  Piot}.} \bibinfo{year}{2022}\natexlab{}.
\newblock \showarticletitle{Emergent Communication at Scale}. In
  \bibinfo{booktitle}{\emph{International Conference on Learning
  Representations}}.
\newblock
\urldef\tempurl%
\url{https://openreview.net/forum?id=AUGBfDIV9rL}
\showURL{%
\tempurl}


\bibitem[\protect\citeauthoryear{Chane-Sane, Schmid, and Laptev}{Chane-Sane
  et~al\mbox{.}}{2021}]%
        {chane2021goal}
\bibfield{author}{\bibinfo{person}{Elliot Chane-Sane},
  \bibinfo{person}{Cordelia Schmid}, {and} \bibinfo{person}{Ivan Laptev}.}
  \bibinfo{year}{2021}\natexlab{}.
\newblock \showarticletitle{Goal-conditioned reinforcement learning with
  imagined subgoals}. In \bibinfo{booktitle}{\emph{International Conference on
  Machine Learning}}. PMLR, \bibinfo{pages}{1430--1440}.
\newblock


\bibitem[\protect\citeauthoryear{Cherry}{Cherry}{1966}]%
        {cherry1966human}
\bibfield{author}{\bibinfo{person}{Colin Cherry}.}
  \bibinfo{year}{1966}\natexlab{}.
\newblock \showarticletitle{On human communication}.
\newblock  (\bibinfo{year}{1966}).
\newblock


\bibitem[\protect\citeauthoryear{Choi, Lazaridou, and de~Freitas}{Choi
  et~al\mbox{.}}{2018}]%
        {choicompositional}
\bibfield{author}{\bibinfo{person}{Edward Choi}, \bibinfo{person}{Angeliki
  Lazaridou}, {and} \bibinfo{person}{Nando de Freitas}.}
  \bibinfo{year}{2018}\natexlab{}.
\newblock \showarticletitle{Compositional Obverter Communication Learning from
  Raw Visual Input}. In \bibinfo{booktitle}{\emph{International Conference on
  Learning Representations}}.
\newblock


\bibitem[\protect\citeauthoryear{Dao and Gu}{Dao and Gu}{2024}]%
        {dao2024transformers}
\bibfield{author}{\bibinfo{person}{Tri Dao} {and} \bibinfo{person}{Albert Gu}.}
  \bibinfo{year}{2024}\natexlab{}.
\newblock \showarticletitle{Transformers are SSMs: Generalized models and
  efficient algorithms through structured state space duality}.
\newblock \bibinfo{journal}{\emph{arXiv preprint arXiv:2405.21060}}
  (\bibinfo{year}{2024}).
\newblock


\bibitem[\protect\citeauthoryear{Dessi, Kharitonov, and Marco}{Dessi
  et~al\mbox{.}}{2021}]%
        {dessi2021interpretable}
\bibfield{author}{\bibinfo{person}{Roberto Dessi}, \bibinfo{person}{Eugene
  Kharitonov}, {and} \bibinfo{person}{Baroni Marco}.}
  \bibinfo{year}{2021}\natexlab{}.
\newblock \showarticletitle{Interpretable agent communication from scratch
  (with a generic visual processor emerging on the side)}. In
  \bibinfo{booktitle}{\emph{Advances in Neural Information Processing
  Systems}}, \bibfield{editor}{\bibinfo{person}{M.~Ranzato},
  \bibinfo{person}{A.~Beygelzimer}, \bibinfo{person}{Y.~Dauphin},
  \bibinfo{person}{P.S. Liang}, {and} \bibinfo{person}{J.~Wortman Vaughan}}
  (Eds.), Vol.~\bibinfo{volume}{34}. \bibinfo{publisher}{Curran Associates,
  Inc.}, \bibinfo{pages}{26937--26949}.
\newblock
\urldef\tempurl%
\url{https://proceedings.neurips.cc/paper_files/paper/2021/file/e250c59336b505ed411d455abaa30b4d-Paper.pdf}
\showURL{%
\tempurl}


\bibitem[\protect\citeauthoryear{Dubey, Jauhri, Pandey, Kadian, Al-Dahle,
  Letman, Mathur, Schelten, Yang, Fan, et~al\mbox{.}}{Dubey
  et~al\mbox{.}}{2024}]%
        {dubey2024llama}
\bibfield{author}{\bibinfo{person}{Abhimanyu Dubey}, \bibinfo{person}{Abhinav
  Jauhri}, \bibinfo{person}{Abhinav Pandey}, \bibinfo{person}{Abhishek Kadian},
  \bibinfo{person}{Ahmad Al-Dahle}, \bibinfo{person}{Aiesha Letman},
  \bibinfo{person}{Akhil Mathur}, \bibinfo{person}{Alan Schelten},
  \bibinfo{person}{Amy Yang}, \bibinfo{person}{Angela Fan}, {et~al\mbox{.}}}
  \bibinfo{year}{2024}\natexlab{}.
\newblock \showarticletitle{The llama 3 herd of models}.
\newblock \bibinfo{journal}{\emph{arXiv preprint arXiv:2407.21783}}
  (\bibinfo{year}{2024}).
\newblock


\bibitem[\protect\citeauthoryear{Foerster, Assael, De~Freitas, and
  Whiteson}{Foerster et~al\mbox{.}}{2016}]%
        {foerster2016learning}
\bibfield{author}{\bibinfo{person}{Jakob Foerster},
  \bibinfo{person}{Ioannis~Alexandros Assael}, \bibinfo{person}{Nando
  De~Freitas}, {and} \bibinfo{person}{Shimon Whiteson}.}
  \bibinfo{year}{2016}\natexlab{}.
\newblock \showarticletitle{Learning to communicate with deep multi-agent
  reinforcement learning}.
\newblock \bibinfo{journal}{\emph{Advances in neural information processing
  systems}}  \bibinfo{volume}{29} (\bibinfo{year}{2016}).
\newblock


\bibitem[\protect\citeauthoryear{Graesser, Cho, and Kiela}{Graesser
  et~al\mbox{.}}{2019}]%
        {graesser2019emergent}
\bibfield{author}{\bibinfo{person}{Laura Graesser}, \bibinfo{person}{Kyunghyun
  Cho}, {and} \bibinfo{person}{Douwe Kiela}.} \bibinfo{year}{2019}\natexlab{}.
\newblock \showarticletitle{Emergent linguistic phenomena in multi-agent
  communication games}. In \bibinfo{booktitle}{\emph{2019 Conference on
  Empirical Methods in Natural Language Processing and 9th International Joint
  Conference on Natural Language Processing, EMNLP-IJCNLP 2019}}. Association
  for Computational Linguistics, \bibinfo{pages}{3700--3710}.
\newblock


\bibitem[\protect\citeauthoryear{Grill, Strub, Altch{\'e}, Tallec, Richemond,
  Buchatskaya, Doersch, Avila~Pires, Guo, Gheshlaghi~Azar, et~al\mbox{.}}{Grill
  et~al\mbox{.}}{2020}]%
        {grill2020bootstrap}
\bibfield{author}{\bibinfo{person}{Jean-Bastien Grill},
  \bibinfo{person}{Florian Strub}, \bibinfo{person}{Florent Altch{\'e}},
  \bibinfo{person}{Corentin Tallec}, \bibinfo{person}{Pierre Richemond},
  \bibinfo{person}{Elena Buchatskaya}, \bibinfo{person}{Carl Doersch},
  \bibinfo{person}{Bernardo Avila~Pires}, \bibinfo{person}{Zhaohan Guo},
  \bibinfo{person}{Mohammad Gheshlaghi~Azar}, {et~al\mbox{.}}}
  \bibinfo{year}{2020}\natexlab{}.
\newblock \showarticletitle{Bootstrap your own latent-a new approach to
  self-supervised learning}.
\newblock \bibinfo{journal}{\emph{Advances in neural information processing
  systems}}  \bibinfo{volume}{33} (\bibinfo{year}{2020}),
  \bibinfo{pages}{21271--21284}.
\newblock


\bibitem[\protect\citeauthoryear{Guo, Ren, Havrylov, Frank, Titov, and
  Smith}{Guo et~al\mbox{.}}{2019}]%
        {guo2019emergence}
\bibfield{author}{\bibinfo{person}{Shangmin Guo}, \bibinfo{person}{Yi Ren},
  \bibinfo{person}{Serhii Havrylov}, \bibinfo{person}{Stella Frank},
  \bibinfo{person}{Ivan Titov}, {and} \bibinfo{person}{Kenny Smith}.}
  \bibinfo{year}{2019}\natexlab{}.
\newblock \showarticletitle{The emergence of compositional languages for
  numeric concepts through iterated learning in neural agents}.
\newblock \bibinfo{journal}{\emph{arXiv preprint arXiv:1910.05291}}
  (\bibinfo{year}{2019}).
\newblock


\bibitem[\protect\citeauthoryear{Havrylov and Titov}{Havrylov and
  Titov}{2017}]%
        {havrylov2017emergence}
\bibfield{author}{\bibinfo{person}{Serhii Havrylov} {and} \bibinfo{person}{Ivan
  Titov}.} \bibinfo{year}{2017}\natexlab{}.
\newblock \showarticletitle{Emergence of language with multi-agent games:
  Learning to communicate with sequences of symbols}.
\newblock \bibinfo{journal}{\emph{Advances in neural information processing
  systems}}  \bibinfo{volume}{30} (\bibinfo{year}{2017}).
\newblock


\bibitem[\protect\citeauthoryear{Hayashi, Raymond, and Sidnell}{Hayashi
  et~al\mbox{.}}{2013}]%
        {hayashi2013conversational}
\bibfield{author}{\bibinfo{person}{Makoto Hayashi}, \bibinfo{person}{Geoffrey
  Raymond}, {and} \bibinfo{person}{Jack Sidnell}.}
  \bibinfo{year}{2013}\natexlab{}.
\newblock \bibinfo{booktitle}{\emph{Conversational repair and human
  understanding}}.
\newblock Number~30. \bibinfo{publisher}{Cambridge University Press}.
\newblock


\bibitem[\protect\citeauthoryear{He, Zhang, Ren, and Sun}{He
  et~al\mbox{.}}{2016}]%
        {he2016deep}
\bibfield{author}{\bibinfo{person}{Kaiming He}, \bibinfo{person}{Xiangyu
  Zhang}, \bibinfo{person}{Shaoqing Ren}, {and} \bibinfo{person}{Jian Sun}.}
  \bibinfo{year}{2016}\natexlab{}.
\newblock \showarticletitle{Deep residual learning for image recognition}. In
  \bibinfo{booktitle}{\emph{Proceedings of the IEEE conference on computer
  vision and pattern recognition}}. \bibinfo{pages}{770--778}.
\newblock


\bibitem[\protect\citeauthoryear{Hochreiter and Schmidhuber}{Hochreiter and
  Schmidhuber}{1997}]%
        {hochreiter1997long}
\bibfield{author}{\bibinfo{person}{Sepp Hochreiter} {and}
  \bibinfo{person}{J{\"u}rgen Schmidhuber}.} \bibinfo{year}{1997}\natexlab{}.
\newblock \showarticletitle{Long short-term memory}.
\newblock \bibinfo{journal}{\emph{Neural computation}} \bibinfo{volume}{9},
  \bibinfo{number}{8} (\bibinfo{year}{1997}), \bibinfo{pages}{1735--1780}.
\newblock


\bibitem[\protect\citeauthoryear{Jang, Gu, and Poole}{Jang
  et~al\mbox{.}}{2016}]%
        {jang2016categorical}
\bibfield{author}{\bibinfo{person}{Eric Jang}, \bibinfo{person}{Shixiang Gu},
  {and} \bibinfo{person}{Ben Poole}.} \bibinfo{year}{2016}\natexlab{}.
\newblock \showarticletitle{Categorical reparameterization with
  gumbel-softmax}.
\newblock \bibinfo{journal}{\emph{arXiv preprint arXiv:1611.01144}}
  (\bibinfo{year}{2016}).
\newblock


\bibitem[\protect\citeauthoryear{Jiang, Sablayrolles, Mensch, Bamford, Chaplot,
  Casas, Bressand, Lengyel, Lample, Saulnier, et~al\mbox{.}}{Jiang
  et~al\mbox{.}}{2023}]%
        {jiang2023mistral}
\bibfield{author}{\bibinfo{person}{Albert~Q Jiang}, \bibinfo{person}{Alexandre
  Sablayrolles}, \bibinfo{person}{Arthur Mensch}, \bibinfo{person}{Chris
  Bamford}, \bibinfo{person}{Devendra~Singh Chaplot}, \bibinfo{person}{Diego
  de~las Casas}, \bibinfo{person}{Florian Bressand}, \bibinfo{person}{Gianna
  Lengyel}, \bibinfo{person}{Guillaume Lample}, \bibinfo{person}{Lucile
  Saulnier}, {et~al\mbox{.}}} \bibinfo{year}{2023}\natexlab{}.
\newblock \showarticletitle{Mistral 7B}.
\newblock \bibinfo{journal}{\emph{arXiv preprint arXiv:2310.06825}}
  (\bibinfo{year}{2023}).
\newblock


\bibitem[\protect\citeauthoryear{Jorge, K{\aa}geb{\"a}ck, Johansson, and
  Gustavsson}{Jorge et~al\mbox{.}}{2016}]%
        {jorge2016learning}
\bibfield{author}{\bibinfo{person}{Emilio Jorge}, \bibinfo{person}{Mikael
  K{\aa}geb{\"a}ck}, \bibinfo{person}{Fredrik~D Johansson}, {and}
  \bibinfo{person}{Emil Gustavsson}.} \bibinfo{year}{2016}\natexlab{}.
\newblock \showarticletitle{Learning to play guess who? and inventing a
  grounded language as a consequence}.
\newblock \bibinfo{journal}{\emph{arXiv preprint arXiv:1611.03218}}
  (\bibinfo{year}{2016}).
\newblock


\bibitem[\protect\citeauthoryear{Konda and Tsitsiklis}{Konda and
  Tsitsiklis}{1999}]%
        {konda1999actor}
\bibfield{author}{\bibinfo{person}{Vijay Konda} {and} \bibinfo{person}{John
  Tsitsiklis}.} \bibinfo{year}{1999}\natexlab{}.
\newblock \showarticletitle{Actor-Critic Algorithms}. In
  \bibinfo{booktitle}{\emph{Advances in Neural Information Processing
  Systems}}, \bibfield{editor}{\bibinfo{person}{S.~Solla},
  \bibinfo{person}{T.~Leen}, {and} \bibinfo{person}{K.~M\"{u}ller}} (Eds.),
  Vol.~\bibinfo{volume}{12}. \bibinfo{publisher}{MIT Press}.
\newblock
\urldef\tempurl%
\url{https://proceedings.neurips.cc/paper_files/paper/1999/file/6449f44a102fde848669bdd9eb6b76fa-Paper.pdf}
\showURL{%
\tempurl}


\bibitem[\protect\citeauthoryear{Kuci{\'n}ski, Korbak, Ko{\l}odziej, and
  Mi{\l}o{\'s}}{Kuci{\'n}ski et~al\mbox{.}}{2021}]%
        {kucinski2021catalytic}
\bibfield{author}{\bibinfo{person}{{\L}ukasz Kuci{\'n}ski},
  \bibinfo{person}{Tomasz Korbak}, \bibinfo{person}{Pawe{\l} Ko{\l}odziej},
  {and} \bibinfo{person}{Piotr Mi{\l}o{\'s}}.} \bibinfo{year}{2021}\natexlab{}.
\newblock \showarticletitle{Catalytic role of noise and necessity of inductive
  biases in the emergence of compositional communication}.
\newblock \bibinfo{journal}{\emph{Advances in Neural Information Processing
  Systems}}  \bibinfo{volume}{34} (\bibinfo{year}{2021}),
  \bibinfo{pages}{23075--23088}.
\newblock


\bibitem[\protect\citeauthoryear{Lazaridou, Hermann, Tuyls, and
  Clark}{Lazaridou et~al\mbox{.}}{2018}]%
        {lazaridou2018emergence}
\bibfield{author}{\bibinfo{person}{Angeliki Lazaridou},
  \bibinfo{person}{Karl~Moritz Hermann}, \bibinfo{person}{Karl Tuyls}, {and}
  \bibinfo{person}{Stephen Clark}.} \bibinfo{year}{2018}\natexlab{}.
\newblock \showarticletitle{Emergence of Linguistic Communication from
  Referential Games with Symbolic and Pixel Input}. In
  \bibinfo{booktitle}{\emph{6th International Conference on Learning
  Representations, ICLR 2018-Conference Track Proceedings}}.
\newblock


\bibitem[\protect\citeauthoryear{Lazaridou, Peysakhovich, and Baroni}{Lazaridou
  et~al\mbox{.}}{2017}]%
        {lazaridou2017multiagent}
\bibfield{author}{\bibinfo{person}{Angeliki Lazaridou},
  \bibinfo{person}{Alexander Peysakhovich}, {and} \bibinfo{person}{Marco
  Baroni}.} \bibinfo{year}{2017}\natexlab{}.
\newblock \showarticletitle{Multi-Agent Cooperation and the Emergence of
  (Natural) Language}. In \bibinfo{booktitle}{\emph{International Conference on
  Learning Representations}}.
\newblock
\urldef\tempurl%
\url{https://openreview.net/forum?id=Hk8N3Sclg}
\showURL{%
\tempurl}


\bibitem[\protect\citeauthoryear{Lemon}{Lemon}{2022}]%
        {lemon2022conversational}
\bibfield{author}{\bibinfo{person}{Oliver Lemon}.}
  \bibinfo{year}{2022}\natexlab{}.
\newblock \showarticletitle{Conversational grounding in emergent
  communication--data and divergence}. In \bibinfo{booktitle}{\emph{Emergent
  Communication Workshop at ICLR 2022}}.
\newblock


\bibitem[\protect\citeauthoryear{Lewis}{Lewis}{1979}]%
        {Lewis1979}
\bibfield{author}{\bibinfo{person}{David Lewis}.}
  \bibinfo{year}{1979}\natexlab{}.
\newblock \showarticletitle{Scorekeeping in a language game}.
\newblock \bibinfo{journal}{\emph{Journal of Philosophical Logic}}
  \bibinfo{volume}{8}, \bibinfo{number}{1} (\bibinfo{date}{01 Jan}
  \bibinfo{year}{1979}), \bibinfo{pages}{339--359}.
\newblock
\showISSN{1573-0433}
\urldef\tempurl%
\url{https://doi.org/10.1007/BF00258436}
\showDOI{\tempurl}


\bibitem[\protect\citeauthoryear{Li and Bowling}{Li and Bowling}{2019a}]%
        {li2019ease}
\bibfield{author}{\bibinfo{person}{Fushan Li} {and} \bibinfo{person}{Michael
  Bowling}.} \bibinfo{year}{2019}\natexlab{a}.
\newblock \showarticletitle{Ease-of-teaching and language structure from
  emergent communication}.
\newblock \bibinfo{journal}{\emph{Advances in neural information processing
  systems}}  \bibinfo{volume}{32} (\bibinfo{year}{2019}).
\newblock


\bibitem[\protect\citeauthoryear{Li and Bowling}{Li and Bowling}{2019b}]%
        {NEURIPS2019_b0cf188d}
\bibfield{author}{\bibinfo{person}{Fushan Li} {and} \bibinfo{person}{Michael
  Bowling}.} \bibinfo{year}{2019}\natexlab{b}.
\newblock \showarticletitle{Ease-of-Teaching and Language Structure from
  Emergent Communication}. In \bibinfo{booktitle}{\emph{Advances in Neural
  Information Processing Systems}},
  \bibfield{editor}{\bibinfo{person}{H.~Wallach},
  \bibinfo{person}{H.~Larochelle}, \bibinfo{person}{A.~Beygelzimer},
  \bibinfo{person}{F.~d\textquotesingle Alch\'{e}-Buc},
  \bibinfo{person}{E.~Fox}, {and} \bibinfo{person}{R.~Garnett}} (Eds.),
  Vol.~\bibinfo{volume}{32}. \bibinfo{publisher}{Curran Associates, Inc.}
\newblock
\urldef\tempurl%
\url{https://proceedings.neurips.cc/paper_files/paper/2019/file/b0cf188d74589db9b23d5d277238a929-Paper.pdf}
\showURL{%
\tempurl}


\bibitem[\protect\citeauthoryear{Liu, Luo, Wang, and Tang}{Liu
  et~al\mbox{.}}{2015}]%
        {liu2015faceattributes}
\bibfield{author}{\bibinfo{person}{Ziwei Liu}, \bibinfo{person}{Ping Luo},
  \bibinfo{person}{Xiaogang Wang}, {and} \bibinfo{person}{Xiaoou Tang}.}
  \bibinfo{year}{2015}\natexlab{}.
\newblock \showarticletitle{Deep Learning Face Attributes in the Wild}. In
  \bibinfo{booktitle}{\emph{Proceedings of International Conference on Computer
  Vision (ICCV)}}.
\newblock


\bibitem[\protect\citeauthoryear{Loshchilov and Hutter}{Loshchilov and
  Hutter}{2017}]%
        {loshchilov2017decoupled}
\bibfield{author}{\bibinfo{person}{Ilya Loshchilov} {and}
  \bibinfo{person}{Frank Hutter}.} \bibinfo{year}{2017}\natexlab{}.
\newblock \showarticletitle{Decoupled weight decay regularization}.
\newblock \bibinfo{journal}{\emph{arXiv preprint arXiv:1711.05101}}
  (\bibinfo{year}{2017}).
\newblock


\bibitem[\protect\citeauthoryear{Mordatch and Abbeel}{Mordatch and
  Abbeel}{2018}]%
        {mordatch2018emergence}
\bibfield{author}{\bibinfo{person}{Igor Mordatch} {and} \bibinfo{person}{Pieter
  Abbeel}.} \bibinfo{year}{2018}\natexlab{}.
\newblock \showarticletitle{Emergence of grounded compositional language in
  multi-agent populations}. In \bibinfo{booktitle}{\emph{Proceedings of the
  AAAI conference on artificial intelligence}}, Vol.~\bibinfo{volume}{32}.
\newblock


\bibitem[\protect\citeauthoryear{Nikolaus}{Nikolaus}{2023}]%
        {nikolaus2023emergent}
\bibfield{author}{\bibinfo{person}{Mitja Nikolaus}.}
  \bibinfo{year}{2023}\natexlab{}.
\newblock \showarticletitle{Emergent Communication with Conversational Repair}.
  In \bibinfo{booktitle}{\emph{The Twelfth International Conference on Learning
  Representations}}.
\newblock


\bibitem[\protect\citeauthoryear{Nowak and Krakauer}{Nowak and
  Krakauer}{1999}]%
        {nowak1999evolution}
\bibfield{author}{\bibinfo{person}{Martin~A Nowak} {and}
  \bibinfo{person}{David~C Krakauer}.} \bibinfo{year}{1999}\natexlab{}.
\newblock \showarticletitle{The evolution of language}.
\newblock \bibinfo{journal}{\emph{Proceedings of the National Academy of
  Sciences}} \bibinfo{volume}{96}, \bibinfo{number}{14} (\bibinfo{year}{1999}),
  \bibinfo{pages}{8028--8033}.
\newblock


\bibitem[\protect\citeauthoryear{Oord, Li, and Vinyals}{Oord
  et~al\mbox{.}}{2018}]%
        {oord2018representation}
\bibfield{author}{\bibinfo{person}{Aaron van~den Oord}, \bibinfo{person}{Yazhe
  Li}, {and} \bibinfo{person}{Oriol Vinyals}.} \bibinfo{year}{2018}\natexlab{}.
\newblock \showarticletitle{Representation learning with contrastive predictive
  coding}.
\newblock \bibinfo{journal}{\emph{arXiv preprint arXiv:1807.03748}}
  (\bibinfo{year}{2018}).
\newblock


\bibitem[\protect\citeauthoryear{OpenAI}{OpenAI}{2023}]%
        {openai2023gpt4}
\bibfield{author}{\bibinfo{person}{OpenAI}.} \bibinfo{year}{2023}\natexlab{}.
\newblock \bibinfo{title}{GPT-4 Technical Report}.
\newblock
\newblock
\showeprint[arxiv]{2303.08774}~[cs.CL]


\bibitem[\protect\citeauthoryear{Painter-Wakefield and Parr}{Painter-Wakefield
  and Parr}{2012}]%
        {10.5555/3042573.3042686}
\bibfield{author}{\bibinfo{person}{Christopher Painter-Wakefield} {and}
  \bibinfo{person}{Ronald Parr}.} \bibinfo{year}{2012}\natexlab{}.
\newblock \showarticletitle{Greedy algorithms for sparse reinforcement
  learning}. In \bibinfo{booktitle}{\emph{Proceedings of the 29th International
  Coference on International Conference on Machine Learning}} (Edinburgh,
  Scotland) \emph{(\bibinfo{series}{ICML'12})}. \bibinfo{publisher}{Omnipress},
  \bibinfo{address}{Madison, WI, USA}, \bibinfo{pages}{867–874}.
\newblock
\showISBNx{9781450312851}


\bibitem[\protect\citeauthoryear{Pascanu, Mikolov, and Bengio}{Pascanu
  et~al\mbox{.}}{2013}]%
        {pascanu2013difficulty}
\bibfield{author}{\bibinfo{person}{Razvan Pascanu}, \bibinfo{person}{Tomas
  Mikolov}, {and} \bibinfo{person}{Yoshua Bengio}.}
  \bibinfo{year}{2013}\natexlab{}.
\newblock \showarticletitle{On the difficulty of training recurrent neural
  networks}. In \bibinfo{booktitle}{\emph{International conference on machine
  learning}}. Pmlr, \bibinfo{pages}{1310--1318}.
\newblock


\bibitem[\protect\citeauthoryear{Premack and Woodruff}{Premack and
  Woodruff}{1978}]%
        {premack1978does}
\bibfield{author}{\bibinfo{person}{David Premack} {and} \bibinfo{person}{Guy
  Woodruff}.} \bibinfo{year}{1978}\natexlab{}.
\newblock \showarticletitle{Does the chimpanzee have a theory of mind?}
\newblock \bibinfo{journal}{\emph{Behavioral and brain sciences}}
  \bibinfo{volume}{1}, \bibinfo{number}{4} (\bibinfo{year}{1978}),
  \bibinfo{pages}{515--526}.
\newblock


\bibitem[\protect\citeauthoryear{Rawlik, Toussaint, and Vijayakumar}{Rawlik
  et~al\mbox{.}}{2012}]%
        {rawlik2012stochastic}
\bibfield{author}{\bibinfo{person}{Konrad Rawlik}, \bibinfo{person}{Marc
  Toussaint}, {and} \bibinfo{person}{Sethu Vijayakumar}.}
  \bibinfo{year}{2012}\natexlab{}.
\newblock \showarticletitle{On stochastic optimal control and reinforcement
  learning by approximate inference}.
\newblock \bibinfo{journal}{\emph{Proceedings of Robotics: Science and Systems
  VIII}} (\bibinfo{year}{2012}).
\newblock


\bibitem[\protect\citeauthoryear{Ren, Guo, Labeau, Cohen, and Kirby}{Ren
  et~al\mbox{.}}{2020}]%
        {Ren2020Compositional}
\bibfield{author}{\bibinfo{person}{Yi Ren}, \bibinfo{person}{Shangmin Guo},
  \bibinfo{person}{Matthieu Labeau}, \bibinfo{person}{Shay~B. Cohen}, {and}
  \bibinfo{person}{Simon Kirby}.} \bibinfo{year}{2020}\natexlab{}.
\newblock \showarticletitle{Compositional languages emerge in a neural iterated
  learning model}. In \bibinfo{booktitle}{\emph{International Conference on
  Learning Representations}}.
\newblock
\urldef\tempurl%
\url{https://openreview.net/forum?id=HkePNpVKPB}
\showURL{%
\tempurl}


\bibitem[\protect\citeauthoryear{Rita, Strub, Grill, Pietquin, and Dupoux}{Rita
  et~al\mbox{.}}{2022}]%
        {rita2022on}
\bibfield{author}{\bibinfo{person}{Mathieu Rita}, \bibinfo{person}{Florian
  Strub}, \bibinfo{person}{Jean-Bastien Grill}, \bibinfo{person}{Olivier
  Pietquin}, {and} \bibinfo{person}{Emmanuel Dupoux}.}
  \bibinfo{year}{2022}\natexlab{}.
\newblock \showarticletitle{On the role of population heterogeneity in emergent
  communication}. In \bibinfo{booktitle}{\emph{International Conference on
  Learning Representations}}.
\newblock
\urldef\tempurl%
\url{https://openreview.net/forum?id=5Qkd7-bZfI}
\showURL{%
\tempurl}


\bibitem[\protect\citeauthoryear{Russakovsky, Deng, Su, Krause, Satheesh, Ma,
  Huang, Karpathy, Khosla, Bernstein, Berg, and Fei-Fei}{Russakovsky
  et~al\mbox{.}}{2015}]%
        {ILSVRC15}
\bibfield{author}{\bibinfo{person}{Olga Russakovsky}, \bibinfo{person}{Jia
  Deng}, \bibinfo{person}{Hao Su}, \bibinfo{person}{Jonathan Krause},
  \bibinfo{person}{Sanjeev Satheesh}, \bibinfo{person}{Sean Ma},
  \bibinfo{person}{Zhiheng Huang}, \bibinfo{person}{Andrej Karpathy},
  \bibinfo{person}{Aditya Khosla}, \bibinfo{person}{Michael Bernstein},
  \bibinfo{person}{Alexander~C. Berg}, {and} \bibinfo{person}{Li Fei-Fei}.}
  \bibinfo{year}{2015}\natexlab{}.
\newblock \showarticletitle{{ImageNet Large Scale Visual Recognition
  Challenge}}.
\newblock \bibinfo{journal}{\emph{International Journal of Computer Vision
  (IJCV)}} \bibinfo{volume}{115}, \bibinfo{number}{3} (\bibinfo{year}{2015}),
  \bibinfo{pages}{211--252}.
\newblock
\urldef\tempurl%
\url{https://doi.org/10.1007/s11263-015-0816-y}
\showDOI{\tempurl}


\bibitem[\protect\citeauthoryear{Schegloff, Jefferson, and Sacks}{Schegloff
  et~al\mbox{.}}{1977}]%
        {schegloff1977preference}
\bibfield{author}{\bibinfo{person}{Emanuel~A Schegloff}, \bibinfo{person}{Gail
  Jefferson}, {and} \bibinfo{person}{Harvey Sacks}.}
  \bibinfo{year}{1977}\natexlab{}.
\newblock \showarticletitle{The preference for self-correction in the
  organization of repair in conversation}.
\newblock \bibinfo{journal}{\emph{Language}} \bibinfo{volume}{53},
  \bibinfo{number}{2} (\bibinfo{year}{1977}), \bibinfo{pages}{361--382}.
\newblock


\bibitem[\protect\citeauthoryear{Schulman, Wolski, Dhariwal, Radford, and
  Klimov}{Schulman et~al\mbox{.}}{2017}]%
        {schulman2017proximal}
\bibfield{author}{\bibinfo{person}{John Schulman}, \bibinfo{person}{Filip
  Wolski}, \bibinfo{person}{Prafulla Dhariwal}, \bibinfo{person}{Alec Radford},
  {and} \bibinfo{person}{Oleg Klimov}.} \bibinfo{year}{2017}\natexlab{}.
\newblock \showarticletitle{Proximal policy optimization algorithms}.
\newblock \bibinfo{journal}{\emph{arXiv preprint arXiv:1707.06347}}
  (\bibinfo{year}{2017}).
\newblock


\bibitem[\protect\citeauthoryear{Sukhbaatar, Fergus, et~al\mbox{.}}{Sukhbaatar
  et~al\mbox{.}}{2016}]%
        {sukhbaatar2016learning}
\bibfield{author}{\bibinfo{person}{Sainbayar Sukhbaatar}, \bibinfo{person}{Rob
  Fergus}, {et~al\mbox{.}}} \bibinfo{year}{2016}\natexlab{}.
\newblock \showarticletitle{Learning multiagent communication with
  backpropagation}.
\newblock \bibinfo{journal}{\emph{Advances in neural information processing
  systems}}  \bibinfo{volume}{29} (\bibinfo{year}{2016}).
\newblock


\bibitem[\protect\citeauthoryear{Sutton and Barto}{Sutton and Barto}{2018}]%
        {sutton2018reinforcement}
\bibfield{author}{\bibinfo{person}{Richard~S Sutton} {and}
  \bibinfo{person}{Andrew~G Barto}.} \bibinfo{year}{2018}\natexlab{}.
\newblock \bibinfo{booktitle}{\emph{Reinforcement learning: An introduction}}.
\newblock \bibinfo{publisher}{MIT press}.
\newblock


\bibitem[\protect\citeauthoryear{Tucker, Li, Agrawal, Hughes, Sycara, Lewis,
  and Shah}{Tucker et~al\mbox{.}}{2021}]%
        {tucker2021emergent}
\bibfield{author}{\bibinfo{person}{Mycal Tucker}, \bibinfo{person}{Huao Li},
  \bibinfo{person}{Siddharth Agrawal}, \bibinfo{person}{Dana Hughes},
  \bibinfo{person}{Katia Sycara}, \bibinfo{person}{Michael Lewis}, {and}
  \bibinfo{person}{Julie~A Shah}.} \bibinfo{year}{2021}\natexlab{}.
\newblock \showarticletitle{Emergent discrete communication in semantic
  spaces}.
\newblock \bibinfo{journal}{\emph{Advances in Neural Information Processing
  Systems}}  \bibinfo{volume}{34} (\bibinfo{year}{2021}),
  \bibinfo{pages}{10574--10586}.
\newblock


\bibitem[\protect\citeauthoryear{Ueda and Washio}{Ueda and Washio}{2021}]%
        {ueda-washio-2021-relationship}
\bibfield{author}{\bibinfo{person}{Ryo Ueda} {and} \bibinfo{person}{Koki
  Washio}.} \bibinfo{year}{2021}\natexlab{}.
\newblock \showarticletitle{On the Relationship between {Z}ipf{'}s Law of
  Abbreviation and Interfering Noise in Emergent Languages}. In
  \bibinfo{booktitle}{\emph{Proceedings of the 59th Annual Meeting of the
  Association for Computational Linguistics and the 11th International Joint
  Conference on Natural Language Processing: Student Research Workshop}}.
  \bibinfo{publisher}{Association for Computational Linguistics},
  \bibinfo{address}{Online}, \bibinfo{pages}{60--70}.
\newblock
\urldef\tempurl%
\url{https://doi.org/10.18653/v1/2021.acl-srw.6}
\showDOI{\tempurl}


\bibitem[\protect\citeauthoryear{Vieillard, Kozuno, Scherrer, Pietquin, Munos,
  and Geist}{Vieillard et~al\mbox{.}}{2020}]%
        {vieillard2020leverage}
\bibfield{author}{\bibinfo{person}{Nino Vieillard}, \bibinfo{person}{Tadashi
  Kozuno}, \bibinfo{person}{Bruno Scherrer}, \bibinfo{person}{Olivier
  Pietquin}, \bibinfo{person}{Remi Munos}, {and} \bibinfo{person}{Matthieu
  Geist}.} \bibinfo{year}{2020}\natexlab{}.
\newblock \showarticletitle{Leverage the Average: an Analysis of KL
  Regularization in Reinforcement Learning}. In
  \bibinfo{booktitle}{\emph{Advances in Neural Information Processing
  Systems}}, \bibfield{editor}{\bibinfo{person}{H.~Larochelle},
  \bibinfo{person}{M.~Ranzato}, \bibinfo{person}{R.~Hadsell},
  \bibinfo{person}{M.F. Balcan}, {and} \bibinfo{person}{H.~Lin}} (Eds.),
  Vol.~\bibinfo{volume}{33}. \bibinfo{publisher}{Curran Associates, Inc.},
  \bibinfo{pages}{12163--12174}.
\newblock
\urldef\tempurl%
\url{https://proceedings.neurips.cc/paper_files/paper/2020/file/8e2c381d4dd04f1c55093f22c59c3a08-Paper.pdf}
\showURL{%
\tempurl}


\bibitem[\protect\citeauthoryear{Wagner, Reggia, Uriagereka, and
  Wilkinson}{Wagner et~al\mbox{.}}{2003}]%
        {wagner2003progress}
\bibfield{author}{\bibinfo{person}{Kyle Wagner}, \bibinfo{person}{James~A
  Reggia}, \bibinfo{person}{Juan Uriagereka}, {and} \bibinfo{person}{Gerald~S
  Wilkinson}.} \bibinfo{year}{2003}\natexlab{}.
\newblock \showarticletitle{Progress in the simulation of emergent
  communication and language}.
\newblock \bibinfo{journal}{\emph{Adaptive Behavior}} \bibinfo{volume}{11},
  \bibinfo{number}{1} (\bibinfo{year}{2003}), \bibinfo{pages}{37--69}.
\newblock


\bibitem[\protect\citeauthoryear{Williams}{Williams}{1992}]%
        {williams1992simple}
\bibfield{author}{\bibinfo{person}{Ronald~J Williams}.}
  \bibinfo{year}{1992}\natexlab{}.
\newblock \showarticletitle{Simple statistical gradient-following algorithms
  for connectionist reinforcement learning}.
\newblock \bibinfo{journal}{\emph{Machine learning}}  \bibinfo{volume}{8}
  (\bibinfo{year}{1992}), \bibinfo{pages}{229--256}.
\newblock


\bibitem[\protect\citeauthoryear{Winograd}{Winograd}{1972}]%
        {winograd1972understanding}
\bibfield{author}{\bibinfo{person}{Terry Winograd}.}
  \bibinfo{year}{1972}\natexlab{}.
\newblock \showarticletitle{Understanding natural language}.
\newblock \bibinfo{journal}{\emph{Cognitive psychology}} \bibinfo{volume}{3},
  \bibinfo{number}{1} (\bibinfo{year}{1972}), \bibinfo{pages}{1--191}.
\newblock


\bibitem[\protect\citeauthoryear{Zipf}{Zipf}{2013}]%
        {zipf2013psycho}
\bibfield{author}{\bibinfo{person}{George~Kingsley Zipf}.}
  \bibinfo{year}{2013}\natexlab{}.
\newblock \bibinfo{booktitle}{\emph{The psycho-biology of language: An
  introduction to dynamic philology}}.
\newblock \bibinfo{publisher}{Routledge}.
\newblock


\end{thebibliography}


\clearpage

\appendix

\section{Related Work} \label{app:rw}
One of the core properties of emergent communication is that agents learn a communication protocol on their own, where coordination is needed to solve the underlying task. The study of emergent communication with neural agents started with continuous communication channels during training. \citet{sukhbaatar2016learning} propose a communication channel that shares continuous vectors where each agent receives a combination of all messages broadcasted by all other agents. \citet{jorge2016learning} proposed a recurrent version of the Lewis Game with continuous messages. While such works have good performances, they benefit from having differential communication channels, making it possible for gradients to pass through. More recent works used a formulation of the Lewis Game to study emergent communication, where messages contain discrete tokens but still allow for gradients to pass through the communication channel~\citet{havrylov2017emergence,mordatch2018emergence,guo2019emergence,chaabouni-etal-2020-compositionality,rita2022on}. As such, these methods employ a DIAL paradigm~\citep{foerster2016learning}, allowing for an end-to-end training scheme (across agents) by sampling discrete tokens using straight-through Gumbel-Softmax estimator~\citep{jang2016categorical}. From a language evolution perspective, these approaches do not fully align with the properties of human communication, which is discrete and undifferentiable.

Other approaches, similarly to our work, close the gap to human language by using discrete channels to communicate at training and execution times, meaning gradients do not flow through the channel. In this case, we have a RIAL approach~\citep{foerster2016learning} where each agent perceives others as part of the environment. In most cases, these works rely on Reinforce~\citep{williams1992simple} or on an actor-critic~\citep{konda1999actor} variation to model the Speaker and Listener, where both agents try to maximize the game's reward. Primarily,~\citet{foerster2016learning,lazaridou2017multiagent} developed discrete communication channels composed of only one symbol. The former study designed games where two agents simultaneously have the Speaker and Listener roles to classify images. Nevertheless, the architecture having independent agents performed poorly for the mentioned task, when compared to a DIAL approach. The latter work implemented a simpler version of the Lewis Game where the Listener discriminates between two images. The Speaker also has information about the images that the Listener will receive. \citet{choicompositional} further extended the latter game design creating agents that can handle messages composed of multiple discrete symbols, called the \textit{Obverter} technique. The authors accomplish this by modeling the Speaker to choose the message that maximizes the Speaker's understanding. This assumption roots the theory of mind~\citep{premack1978does}, where the Speaker assumes the Listener's mind and its own are identical. Although this work is an improvement from past works regarding the emulation of human language, there are still severe limitations. For example, the environment considered has only two effective degrees of freedom that must be modeled (shape and color of simple 3D shapes), limiting severely the input diversity.

New extensions to the Obverter focus on studying specific properties of human language, like compositionality or pragmatics. \citet{Ren2020Compositional} developed an iterated learning strategy for the Lewis Game, trying to create highly compositional languages as a consequence of developing a new language protocol by having several distinct learning phases for each agent. From a linguistics perspective, compositionality is crucial since it encourages the expression of complex concepts through simpler ones. Another approach tries to leverage a population of agents to study its effect on simplifying the emerged language~\citep{graesser2019emergent}. The authors use a simple visual task where agents communicate binary messages through a fixed number of rounds to match an image to a caption. The image depicts a simple shape with a specific color. Since each agent observes only part of the image, they must cooperate to solve the task. \citet{NEURIPS2019_b0cf188d} also consider a variation of the Lewis Game with a discrete communication channel. The main objective is to give another perspective on how to evaluate the structure of the resulting communication protocol, giving experimental evidence that compositional languages are easier to teach to new Listeners. This additional external pressure surfaces when the Speaker interacts with new Listeners during training. The proposed experiments continue in the same line of simplicity since inputs are categorical values with two attributes, messages contain only two tokens, and the number of candidates the Listener discriminates is only five objects.

As a succeeding work, \citet{chaabouni2022emergent} proposed scaling several dimensions of the Lewis Game to create a setup closer to simulating human communication. The scaled dimensions are: the number of candidates received by the Listener, the dataset of images used, and the number of learning agents. Scaling such dimensions makes the referential game more complex, promoting the generality and validity of the experimental results. Moreover, this study also suggests that compositionality is not a natural emergent factor of generalization of a language as the environment becomes more complex (e.g., using real-world images). This work can be seen as the initial starting point of our proposed work. In particular, we follow this game setup and extensively make a more challenging environment where we add an RL agent as the Listener that only has information about the game's outcome and not which candidate is the correct guess (as in the original implementation). This modification alone brings advantages to the generalization capability of the communication protocol, where the game accuracy proportionally increases with the number of candidates when testing on unseen images. Additionally, we introduce a stochastic communication channel aiming to increase environmental pressure to study a specific linguistics property, denoted conversational repair mechanisms, see Sec. 1. As a result, the agents seek to create a protocol where information is given redundantly to overcome the noise effect. This type of robust communication is a form of an implicit repair mechanism.

Closely related to our work,~\citet{nikolaus2023emergent} explore explicit conversational repair mechanisms. The method proposed also introduced noise in the communication channel. As a way to study other initiate repair mechanisms, the authors propose adding a feedback loop where information flows from the Listener to the Speaker. Although this is a positive development, several simplifications are present, which increase the dissimilarity to human languages. Not only is the message sent by the Listener a single binary token but this feedback loop is also triggered after every token message the Speaker sends. As such, this communication loop breaks the turn-taking aspect of the human dialogue, and the feedback information received by the Speaker is extremely reduced. Comparing against our work, both studies have their distinct focus since each explores a different conversational repair mechanism. Nonetheless, our experiment is more complicated since the only feedback the Speaker receives is the outcome of the game. As such, the coordination of a communication protocol becomes more challenging, where only by trial and error can the Speaker understand that giving redundant information is advantageous in such noisy conditions.

Other works in the literature also consider noise in the communication channel. \citet{tucker2021emergent,kucinski2021catalytic} apply a DIAL scheme, where the noise is sampled and added in the continuous space before applying the Gumbel-Softmax trick. As such, the problem is simplified by learning a continuous latent space robust to noise. More closely related to our work is the study presented by~\citet{ueda-washio-2021-relationship}. In such work, we are in an emergent discrete communication setting where the authors use a noisy communication channel to test the Zipf's law of abbreviation~\citep{zipf2013psycho}. As such, the goal is to investigate conditions for which common messages become shorter. In this context, the authors propose an adversarial setting to try to deceive the Listener by giving other plausible messages (different from the Speaker's message), which, as training progresses, promotes the usage of shorter messages. Our method differs from this work since we assume the communication channel can suffer external perturbations, mimicking the loss of information. In our case, we apply noise by changing the corresponding token to a pre-defined token called the \textit{unknown} token. Therefore, we are interested in evaluating the communication protocols' robustness to handle different noise levels at test time. Additionally, our referential games are more complex since we use datasets with natural images for discrimination instead of categorical inputs.

\section{Lewis Game Variants} \label{app:lg}

\begin{figure}[!t]
    \centering
    \includegraphics[width=0.7\linewidth,keepaspectratio]{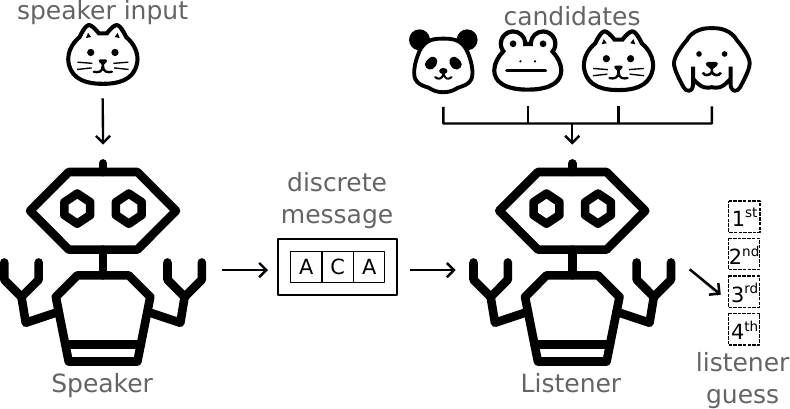}
    \caption{Visual Representation of the Lewis Game (LG). In this illustration, the message, \(\vm\), contains three tokens (\(N=3\)).}
    \label{fig:emcom-lg}
\end{figure}

\Cref{fig:emcom-lg} exposes a sketch for the LG (S) and LG (RL). The message sent to the Listener remains fully observable, without any external noise.

\section{Agent Architecture for LG (S)} \label{app:list-arch}
In this section, we detail the agent architecture proposed by~\citet{chaabouni2022emergent}. We denote this architecture as LG (S), meaning the Listener implements a supervised agent, see \Cref{sec:eval-variants}. We use this game as a frame of reference to benchmark our novel games and agent architectures.

\begin{figure*}[!t]
  \begin{minipage}[!t]{0.49\textwidth}
    \centering
    \includegraphics[width=\linewidth,keepaspectratio]{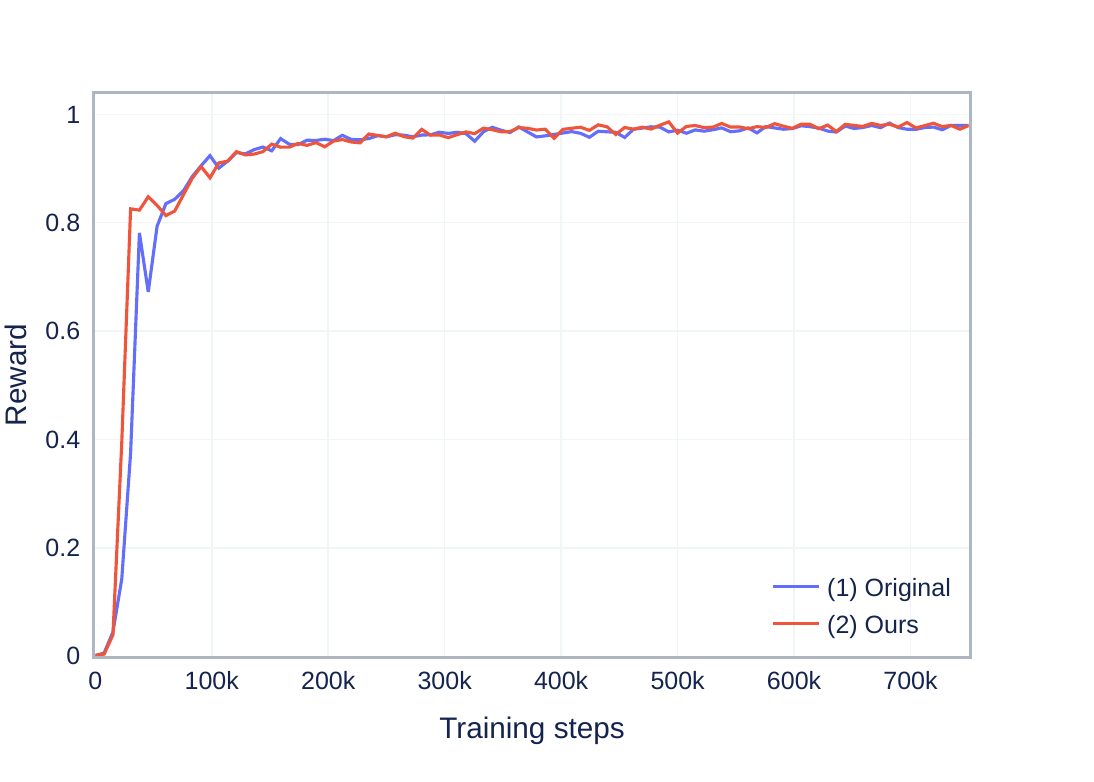}
    \captionof{figure}{Accuracy during training of two implementations of LG (S), (1) Original~\cite{chaabouni2022emergent} and (2) Ours (\cref{app:list-arch}). The dataset used was ImageNet and \(|\sC|=1024\).}
    \label{fig:compare-lg-ss}
  \end{minipage}
  \hfill
  \begin{minipage}[!t]{0.49\textwidth}
        \linespread{0.6}\selectfont\centering
        \centering
        \captionof{table}{Test accuracy with SD for two implementations of LG (S), (1) Original~\cite{chaabouni2022emergent} and (2) Ours (\cref{app:list-arch}), using ImageNet dataset. During training \(|\sC|=1024\).}
        \label{table:compare-lg-ss}
        \begin{tabular}[t]{llrr}
        \toprule
        Game & Implementation & \multicolumn{2}{c}{\(|\sC|\) (test)} \\[1ex]\cmidrule(r){3-4}
         & & \multicolumn{1}{c}{\(1024\)} & \multicolumn{1}{c}{\(4096\)} \\
        \midrule
        LG {\scriptsize(S)} & Original (1)& 0.96 & 0.88 \\[1.1ex]
        LG {\scriptsize(S)} & Ours (2) & 0.96 & 0.88 \\
        \bottomrule
        \end{tabular}
    \end{minipage}
  \end{figure*}

The speaker architecture remains the same as in LG (RL) and NLG (\Cref{sec:meth-arch}). The Listener agent is the only one suffering modifications between games, which take place in the head module. Hence, the base modules that process the message and candidates also remain unchanged between games. We refer to \Cref{sec:meth-arch} for a thorough description of the unchanged modules. The head module of the Supervised agent sends the attention logits through a softmax to convert them into a distribution. The corresponding learning procedure invokes the InfoNCE loss~\cite{oord2018representation} to attract the message representation to the representation of the right candidate while, at the same time, repulsing all other candidates. Furthermore, our LG (S) implementation differs slightly from the original implementation~\citep{chaabouni2022emergent}, where we place the \(\tanh\) activation function before giving the message (\(\vl_m\)) and candidate representations (\(\vl_{j}\)) to the attention mechanism to match the Listener's architecture used in the LG (RL). \cref{fig:compare-lg-ss} illustrates the training procedure for both implementations, where we can see similar data efficiency since both curves jump to a mean reward (accuracy) of 0.8 before the \SI{50}{\kilo{}} steps. Additionally, as the training step increases, both implementations converge to similar values (around \(0.98\)). We also introduce test accuracies in \cref{table:compare-lg-ss}, where both implementations also yield identical results.

\section{Architecture Implementation} \label{app:arch}
We now provide concrete network implementations for the Speaker and Listener agents playing LG (S), LG (RL), and NLG. As detailed previously, the Speaker network architecture remains unchanged for all games. Additionally, since the Listener implementation may vary between games, we first describe common submodules used in every architecture. This set of networks encodes the message and candidates, as described in \Cref{app:list-arch}. Afterward, we introduce the Listener's head implementation for each LG variant.

\subsection{Speaker} \label{app:arch-impl-speaker}
The network architecture implemented for the Speaker agent will receive as input an image and output a message composed of \(N\) discrete tokens, each retrieved from the same fixed vocabulary:
\begin{itemize}
    \item \(f\): A frozen ResNet-50~\citep{he2016deep} model trained with BYOL algorithm~\citep{grill2020bootstrap} on ImageNet dataset~\citep{ILSVRC15}. The output size of the resulting features is \num{2048}. For more details on the weights, see~\citep{chaabouni2022emergent}.
    \item \(g(\cdot\;;\vtheta)\): A single linear layer to reduce the features' dimensionality from \num{2048} to \num{512} in order to fit in the next layer, an LSTM~\citep{hochreiter1997long}. Additionally, we divide the resulting vector into two equal parts, denoting the initial hidden, \(\vz_{0,\vtheta}\), and cell values ,\(\vc_{0,\vtheta}\), of the LSTM.
    \item \(e(\cdot\;;\vtheta)\): Embedding layer to convert discrete tokens into continuous vectors. The embedding layer receives the discrete token \(m_t\), as an integer, and outputs a feature vector \(e(m_t;\vtheta)\) of size \num{10}. The number of embeddings of \(e(\cdot\;;\vtheta)\) is \(|\sW| + 1\).
    \item \(h(\cdot\;;\vtheta)\): Recurrent layer, implemented as a single LSTM of size \num{256}. The initial hidden and cell states are the output of \(g(\cdot\;;\vtheta)\). Additionally, \(h\) processes the continuous version of each message token \(e(m_{t-1};\vtheta)\) iteratively. Furthermore, the computed hidden value \(\vz_{t,\vtheta}\), in each iteration \(t\), will be the input to the value and critics heads, \(\pi_S(\cdot|\vz_{t,\vtheta})\) and \(v(\cdot\;;\vtheta)\), respectively.
    \item \(\pi_S(\cdot|\vz_{t,\vtheta};\vtheta)\): For a given iteration \(t\), the actor policy head uses a linear layer (output size of \(|\sW|\)) followed by the \(\text{softmax}\) function to convert the hidden state \(\vz_{t,\vtheta}\) into a categorical distribution \(\text{Cat}(|\sW|,\pi_S(\cdot|\vz_{t,\vtheta};\vtheta))\), where each value indicates the probability of choosing each token as the next one to add to the message, \(m_{t+1}\).
    \item \(v(\cdot\;;\vtheta)\): Critic value head to estimate the expected cumulative reward, in this case, the reward received for the game \(R(\vx,\hat{\vx})\), by implementing a linear layer with an output size of \num{1}.
\end{itemize}

\subsection{Listener}
The Listener architecture has two input entries to acquire the Speaker's message and the set of candidate images. This architecture suffers internal modifications to adapt to the specifications of the LG variant to play. Moreover, it is also possible to create different Listener architectures to solve the same game as we show in~\cref{sec:eval-variants}, where we have S and RL Listener architectures to play the LG. Hence, we first detail the implementation of common modules used in every Listener architecture. These sub-modules are the message and candidates' sub-modules used to extract information from the message and candidates, respectively. Afterward, we detail the sub-module specific to each different Listener, labeled as the \emph{Listener's head}.

\subsubsection{Common sub-modules}
We describe every network implementation contained in the common Listener's sub-modules (message and candidates sub-modules):
\begin{itemize}
    \item \(f\) \emph{(candidates)}: A frozen ResNet-50 trained with BYOL on the ImageNet dataset, where \(f\) converts the image into a vector with \num{2048} features. \(f\) is the same network used in the Speaker architecture, see~\cref{app:arch-impl-speaker}.
    \item \(c(\cdot\;;\vphi)\)  \emph{(candidates)}: This single linear layer, followed by the \(\tanh\) function, receives each candidate's feature vector, \(f(\vx_i)\), and reduces its dimensionality from \num{2048} to \num{256}. We denote the output for each candidate \(i\) as \(\vl_{i}=c(f(\vx_i);\vphi)\). 
    \item \(e(\cdot\;;\vphi)\) \emph{(message)}: A single embedding layer to convert a discrete message token \(m_t\) into a continuous vector of size \num{10}. The number of embeddings of \(e(\cdot\;;\vphi)\) is \(|\sW|\).
    \item \(h(\cdot\;;\vphi)\) \emph{(message)}: An LSTM layer with a size of \num{512}, receiving the continuous featuers of each message token \(e(m_t;\vphi)\) iteratively. The initial hidden \(\vz_{0,\vphi}\) and cell states \(\vc_{0,\vphi}\) are both \(\vzero\).
    \item \(g(\cdot\;;\vphi)\)  \emph{(message)}: Linear layer followed by the \(\tanh\) function to reduce the dimensionality of the last hidden state value \(\vz_{N,\phi}\) from a vector size \num{512} to \num{256}, denoted as \(\vl_\text{m}\). 
\end{itemize}

\subsubsection{LG (S)} \label{app:arch-list-ss}
Upon the message and candidate sub-modules outputting the respective hidden values, \(\vl_\text{m}\) and \(\vl_{j}\), the Listener architecture computes a single value score between \(\vl_\text{m}\) and each \(\vl_{j}\), using the cosine similarity function, \(\text{cos\_sim}=\vl_\text{m}\cdot \vl_{j} / (\|\vl_\text{m}\| \|\vl_{j}\|)\). Following this step, the similarity outputs pass through a \(\text{softmax}\) function, yielding a categorical distribution suitable to apply InfoNCE loss~\citep{oord2018representation}.

\subsubsection{LG (RL) \& NLG}
The Listener's head architecture for both LG (RL) and NLG has the same structure:
\begin{itemize}
    \item \(\vs\): Non-parametrizable function to combine information coming from the received message and candidate set. The attention mechanism \(s\) happens through a dot product between \(\vl_\text{m}\) and each \(\vl_{j}\), defined as \(\vs=\left[\begin{matrix}\vl_\text{m}\cdot \vl_{1} & \ldots & \vl_\text{m}\cdot \vl_{|\sC|}\end{matrix}\right]^T\).
    \item \(\pi_L(\cdot|\vs)\): \(\pi_L\) corresponds to the Listener's actor head and contains the \(\text{softmax}\) function to convert the logits \(\vs\) into a categorical distribution \(\text{Cat}(|\sC|,\pi_L(\cdot|\vs))\). Each distribution value maps the probability of choosing the candidate with the same index \(\pi_L(j|\vs_\phi))\), where \(j\in\{1,\ldots,|\sC|\}\).
    \item \(v(\cdot\;;\vphi)\): The Listener's critic head \(v\) receives the attention logits \(\vs\) and passes them through a multi-layer perceptron (MLP), outputting a one-dimension value corresponding to a prediction of the reward \(R(\vx,\hat{\vx})\) for the associated game. The hidden and output sizes of the MLP are \((\max(|\sC|/4,4),\allowbreak \max(|\sC| / 16,\allowbreak2),\allowbreak1)\), and the activation function used between layers is the \(\text{ReLU}\) function.
\end{itemize}

\subsection{Hyperparameters}
In this subsection, we detail the hyperparameters used to instantiate all LG experiments, exposed in \Cref{table:common-hyperparams,table:nlg-mrilg-hyperparams}.

\begin{table}[t]
\centering
\caption{Common hyperparameters for all games: LG (S), LG (RL), and NLG. The values of hyperparameters described as a set, \(\{\cdot\}\), means we run experiments with each value in the set.}
\label{table:common-hyperparams}
\begin{tabular}{lc}
\toprule
hyperparameter & value\\
\midrule
training steps & \SI{750}{\kilo{}} \\
\(|\sC|\) & \(\{\SI{16}{},\SI{64}{},\SI{256}{},\SI{1024}{}\}\) \\
\(|\mathcal{W}|\) & \SI{20}{} \\
\(T\) & \SI{10}{} \\
\(\alpha_{S,A}\) & 1 \\
\(\alpha_{S,C}\) & 1 \\
\(\alpha_{S,\mathcal{H}}\) & \SI{1e-4}{} \\
\(\alpha_{S,A}\) & \SI{0.5}{} \\
\(\alpha_{L,A}\) & 1 \\
\(\alpha_{L,C}\) & \SI{1e-3}{} \\
\(\alpha_{L,\mathcal{H}}\) & \SI{1e-4}{} \\
Speaker's optim & adam \\
Listener's optim & adam \\
Speaker's optim lr & \SI{1e-4}{} \\
Listener's optim lr & \SI{5e-5}{} \\
\(\eta\) & \SI{0.99}{}\\
\(\gamma\) & \SI{0.99}{}\\
\bottomrule
\end{tabular}
\end{table}

\begin{table}[t]
\centering
\caption{Hyperparameters exclusive to NLG. The values of hyperparameters described as a set, \(\{\cdot\}\), means we run experiments with each value in the set.}
\label{table:nlg-mrilg-hyperparams}
\begin{tabular}{lc}
\toprule
hyperparameter & value\\
\midrule
\(\lambda_\text{init}\) & \SI{0}{} \\
\(\lambda\) & \(\{\SI{0.25}{},\SI{0.5}{},\SI{0.75}{}\}\) \\
noise schedule & linear \\
noise schedule steps & \SI{300}{\kilo{}} \\
\bottomrule
\end{tabular}
\end{table}

\section{Training Details}
Following the concise introduction about the learning strategy of both agents (\cref{sec:meth-learn}), we now add complementary information detailing the loss functions used. We also propose an ablation study regarding essential architectural and training procedure choices given at the end of \cref{sec:meth-learn}.

\subsection{Learning Strategy} \label{app:learn-strat}
We model both agents as RL agents for the novel LG variants proposed in this study, LG (RL) and NLG. We also follow a RIAL procedure~\citep{foerster2016learning}, where each agent perceives others as part of the environment, meaning no gradients flow between agents, allowing us to completely isolate the loss function of each agent. We will detail, first, the Speaker's loss function and, secondly, the loss function used by the Listener agent.

\subsubsection{Speaker}
The Speaker's objective will converge on creating messages \(\vm=\left(m_t\right)_{t=1}^N\) in such a way as to facilitate the mapping between the generated message and the right candidate (\(\hat{\vx}=\vx\)), chosen by the Listener. The Speaker, parametrized by \(\theta\), generates messages iteratively, where the following message token results from sampling its actor's stochastic policy, \(m_t\sim\pi_S(\cdot|\vx,(m_{t'})_{t'=1}^{t-1})\), conditioned on the target input image \(\vx\) and previously sampled tokens \((m_{t'})_{t'=1}^{t-1}\). As a result, the Speaker will attempt to find the best policy to maximize the expected reward:
\begin{equation*}
    J\left(\vtheta\right) = \mathbb{E}_{\vx\sim\mathcal{U}\left(\sX\right)}\left[\mathbb{E}_{\pi_S(\cdot|\vx)}\left[R\left(\vx,\hat{\vx}\right)\right]\right],
\end{equation*}
where \(\mathbb{E}_{\vx\sim\mathcal{U}\left(\sX\right)}\) considers the expectation for \(\vx\) over the dataset \(\sX\). Since \(\mathcal{U}(\sX)\) is a discrete uniform distribution, each possible image \(\vx\) has the same likelihood of being sampled. Additionally, \(\mathbb{E}_{\pi_S(\cdot|\vx)}\) is over all possible outputs generated by \(\pi_S\), in this case, all sequences of messages \(\vm\), for each \(\vx\).

For a given target image \(\vx\), we define the expected reward as \(V^{\pi_S}(\vx,\hat{\vx})=\mathbb{E}_{\pi_S(\cdot|\vx)}\left[R\left(\vx,\hat{\vx}\right)\right]\). Using the policy gradient theorem, we can derive the gradient for the policy \(\pi_S\) as:
\begin{multline}
    \partial_\vtheta V^{\pi_S}\left(\vx,\hat{\vx}\right) = \mathbb{E}_{\pi_S(\cdot|\vx)}\Bigg[\\
    \sum_{t=1}^{N}R\left(\vx,\hat{\vx}\right)\partial_\vtheta\log{\pi_S\left(m_t|\vx,\left(m_{t'}\right)_{t'=1}^{t-1}\right)}\Bigg]. \label{eq:pol-grad-speaker-no-baseline}
\end{multline}
As we can see in \Eqref{eq:pol-grad-speaker-no-baseline}, we set the discounted cumulative reward as \(R(\vx,\hat{\vx})\). This is true when the Speaker's discount factor, \(\gamma_S\), is set to \(1\), which is our case. This is a valid assumption since messages have a fixed length where only the final Speaker's state (a message with \(N\) tokens) has an implication in the game result, given that it is when the message is sent to the Listener. As such, the Speaker only wants to maximize this final reward without considering delay through time. Moreover, to reduce the overall variance, we also subtract a baseline to \(R(\vx,\hat{\vx})\). In this case, the policy gradient becomes:
\begin{multline}
    \partial_\vtheta V^{\pi_S}\left(\vx,\hat{\vx}\right) =    \mathbb{E}_{\pi_S(\cdot|\vx)}\Bigg[\\
    \sum_{t=1}^{N}\left(R\left(\vx,\hat{\vx}\right)-V_{t-1}^{\pi_S}\left(\vx,\hat{\vx}\right)\right)\partial_\vtheta\log{\pi_S\left(m_t|\vx,\left(m_{t'}\right)_{t'=1}^{t-1}\right)}\Bigg], \label{eq:pol-grad-speaker}
\end{multline}
where \(V_{t-1}^{\pi_S}(\vx,\hat{\vx})=\mathbb{E}_{\pi_S(\cdot|\vx)}[R(\vx,\hat{\vx})|(m_{t'})_{t'=1}^{t-1}]\) contains the value currently conditioned on information gathered until timestep \(t-1\).

To approximate \eqref{eq:pol-grad-speaker}, the Speaker's learning strategy will minimize the following two losses. First, the critic's head \(v(\vz_{t,\vtheta};\vtheta)\) will adjust towards matching \(V_{t-1}^{\pi_S}\left(\vx,\hat{\vx}\right)\) by minimizing the critic loss \(L_\text{S,C}\left(\vtheta\right)\):
\begin{equation*}
    L_\text{S,C}\left(\vtheta\right) = \frac{1}{|\sX'|}\sum_{x\in\sX'}\sum_{t=1}^N\left(R\left(\vx,\hat{\vx}\right)-v\left(\vz_{t,\vtheta};\vtheta\right)\right)^2,
\end{equation*}
where \(\sX'\subset\sX\) is a random batch of the original dataset \(\sX\).

Secondly, the actor's policy loss \(L_\text{S,A}\left(\vtheta\right)\) will minimize the negative of the expected total reward:
%
\begin{multline*}
    L_\text{S,A}\left(\vtheta\right)=-\frac{1}{|\sX'|}\sum_{x\in\sX'}\sum_{t=1}^N\big[\\
    \text{sg}\left(R\left(\vx,\hat{\vx}\right)-v\left(\vz_{t,\vtheta};\vtheta\right)\right)\log{\pi_S\left(m_t|\vz_{t,\vtheta}\right)}\big],
\end{multline*}
where \(\text{sg}\) is the \emph{stop-gradient} function. Additionally, the Speaker's actor head \(\pi_S(\cdot|\vz_{t,\vtheta})\) estimates \(\pi_S(\cdot|\vx,(m_{t'})_{t'=1}^{t-1})\), which is a valid approximation since the conditioned variable, \(\vz_{t,\vtheta}\), jointly encodes information about the target image \(\vx\) and the tokens already present in the message \((m_{t'})_{t'=1}^{t-1}\), due to its recurrent nature.

Furthermore, as detailed in \cref{sec:meth-learn}, the Speaker adds two additional loss terms to optimize together with \eqref{eq:pol-grad-speaker}. One such supplementary loss term is the KL divergence, \(L_{\text{S,KL}}(\vtheta)\), which aims to minimize the relative entropy between the actor's policy \(\pi_S\) and a target version of this policy \(\overline{\pi}_S\), computed as an EMA, \(\overline{\vtheta}\leftarrow (1-\eta)\vtheta+\eta\overline{\vtheta}\), where \(\eta\) is a constant. As shown in previous studies~\citep{schulman2017proximal,vieillard2020leverage}, using an additional KL loss term helps achieve better performance and, especially, stabilizes training which considerably helps in our case. We define \(L_{\text{S,KL}}(\vtheta)\) as:
\begin{multline*}
    L_{\text{S,KL}}\left(\vtheta\right) = \frac{1}{|\sX'|}\sum_{x\in\sX'}\sum_{t=1}^{N} \mathbb{E}_{m\sim\pi_S\left(\cdot|\vz_{t,\vtheta}\right)}\vast[\\
    \pi_S\left(m|\vz_{t,\vtheta}\right)\log\frac{\pi_S\left(m|\vz_{t,\vtheta}\right)}{\overline{\pi}_S\left(m|\vz_{t,\overline{\vtheta}}\right)}\vast].
\end{multline*}
The other additional loss term is an entropy loss term \(L_{\text{S},\mathcal{H}}(\vtheta)\), where the objective passes to increase the actor's policy entropy to incentivize the exploration of new actions (tokens to create the message). Given \(\pi_S\), the entropy loss term minimizes the following negative sampled version of the entropy:
\begin{multline*}
    L_{\text{S},\mathcal{H}}\left(\vtheta\right) = \frac{1}{|\sX'|}\sum_{x\in\sX'}\sum_{t=1}^N \mathbb{E}_{m\sim\pi_S\left(\cdot|\vz_{t,\vtheta}\right)}\big[\\
    \pi_S\left(m|\vz_{t,\vtheta}\right)\log{\pi_S\left(m|\vz_{t,\vtheta}\right)}\big].
\end{multline*}

\subsubsection{Listener}
The Listener's policy, \(\pi_L\), aims at detecting the candidate \(\sC_j\) that corresponds to the target image \(\vx\) received by the Speaker. This reasoning process is also conditioned on the Speaker's message \(\vm\) to guide the discrimination of the candidates to the right one, see \Cref{sec:meth-nlg}. As such, we define the Listener's objective as the maximization of the expected reward of the game:
%
\begin{multline}
    J\left(\vphi\right) = \mathbb{E}_{\vx\sim\mathcal{U}\left(\sX\right),\sC\sim\mathcal{U}\left(\sX\right),\vm\sim\pi_S(\cdot|\vx)}\big[\\
    \mathbb{E}_{\hat{\vx}\sim\pi_L(\cdot|\vm,\sC)}\left[R\left(\vx,\hat{\vx}\right)\right]\big],
\label{eq:list_exp_rwd}
\end{multline}
where the expectation \(\mathbb{E}_{\vx\sim\mathcal{U}(\sX),\sC\sim\mathcal{U}\left(\sX\right),\mM^{(N-1)}\sim\pi_S(\cdot|\vx)}\) considers the space containing all possible combinations of the target image \(\vx\) sampled from a discrete uniform distribution over the training dataset \(\mathcal{U}(\sX)\), set of candidates \(\sC\) also sampled independently from the same distribution \(\mathcal{U}(\sX)\), and all possible messages \(\vm\) generated by the Speaker's policy \(\pi_S\). Note that \(\sC\sim\mathcal{U}(\sX)\) is an abuse of notation (used for readability), where in reality, we have \(\sC'_1,\ldots,\sC'_{|\sC|-1}\stackrel{\text{i.i.d.}}{\sim}\mathcal{U}(\sX)\) and \(\sC=\{\vx\}\cup\sC'\), to make sure the target image \(\vx\) is also in \(\sC\). The second expectation \(\mathbb{E}_{\hat{\vx}\sim\pi_L(\cdot|\vm,\sC)}\) of \eqref{eq:list_exp_rwd} is over all possible actions for the listener's policy when fixing the target image \(\vx\), candidates \(\sC\), and message \(\vm\). We derive this expectation using policy gradient method to get:
\begin{alignat*}{2}
    \partial_\vphi V^{\pi_L}\left(\vx,\vm,\sC\right) = \mathbb{E}_{\pi_L}\left[ R\left(\vx,\hat{\vx}\right)\partial_\vphi\log{\left(\pi_L\left(\hat{\vx}|\vm,\sC\right)\right)}\right].
\end{alignat*}

Similarly to the Speaker, the main objective of the Listener's training encompasses minimizing a critic (value) loss \(L_\text{L,C}(\vphi)\) and an actor's policy loss term \(L_\text{L,A}(\vphi)\). Regarding the value loss term \(L_\text{L,C}(\vphi)\), the function takes the form:
\begin{equation*}
    L_\text{L,C}(\vphi) = \frac{1}{|\sX'|}\sum_{x\in\sX'}\left(R\left(\vx,\hat{\vx}\right)-v\left(\vs;\vphi\right)\right)^2.
\end{equation*}
Additionally, to derive the original policy \(\pi_L(\cdot|\vm,\sC)\), the Listener's actor head \(\pi_L(\cdot|\vs)\) targets the minimization of the negative of the expected rewards:
\begin{align*}
    L_\text{L,A}(\vphi) =
    -\frac{1}{|\sX'|}\sum_{x\in\sX'}\text{sg}\left(R\left(\vx,\hat{\vx}\right)-v\left(\vs;\phi\right)\right)\log\pi_L\left(\hat{\vx}|\vs\right),
\end{align*}
where \(\vs\) encodes information from the message sent by the Speaker and the candidates' set, see \Cref{sec:meth-arch}. At last, we also add an entropy loss term \(L_{\text{L},\mathcal{H}}(\vphi)\) to the Listener's loss to prevent early stagnation to specific actions:
\begin{align*}
    L_{\text{L},\mathcal{H}}\left(\vphi\right) = \frac{1}{|\sX'|}\sum_{x\in\sX'}\mathbb{E}_{\hat{\vx}\sim\pi_L\left(\cdot|\vs\right)}\left[\pi_L\left(\hat{\vx}|\vs\right)\log{\pi_L\left(\hat{\vx}|\vs\right)}\right].
\end{align*}

\subsection{Datasets} \label{app:data}
The datasets used to evaluate the proposed games are the ImageNet~\citep{ILSVRC15} and CelebA~\citep{liu2015faceattributes}. We use the datasets provided by \citet{chaabouni2022emergent}, where the authors preprocess the data, as we will explain next. The ImageNet dataset contains mainly RGB images of objects and animals clustered over 1000 labels. The training of the LG and ETL experiments uses 99\% of the original training data of ImageNet and the official validation set as the test set. The CelebA dataset contains RGB images of celebrity faces denoting 10177 different identities. This dataset also describes binary attributes for each image, like smiling, hair color, glasses, etc. Additionally, a new dataset split is performed to ensure there is overlapping for all identities between all sets (train, validation, and test), see \citet{chaabouni2022emergent}.

Finally, images from both datasets are down-sampled using bicubic sampling, forcing the shorted side to have 256 pixels, and then a center crop of 224x224 pixels is applied. The channel axis also suffers normalization using mean and SD obtained from the ImageNet train set~\citep{he2016deep}. Afterward, a ResNet architecture pre-trained with BYOL \citep{grill2020bootstrap} on ImageNet outputs the representation used in all experiments, see \Cref{app:arch-impl-speaker}.

\subsection{Computational Resources}
All experiments present in this work are computationally tractable with standard GPU hardware. Each experiment takes, at most, \SI{7}{\hour} to run on a single GPU (NVIDIA GeForce RTX 3090), with a peak of GPU memory of less than \SI{12}{\gibi\byte}.

\subsection{Scheduling Noise} \label{app:noise-schedule}
\begin{figure*}[!t]
    \begin{center}
    \begin{subfigure}{.49\textwidth}
      \centering
      \centerline{\includegraphics[width=\linewidth]{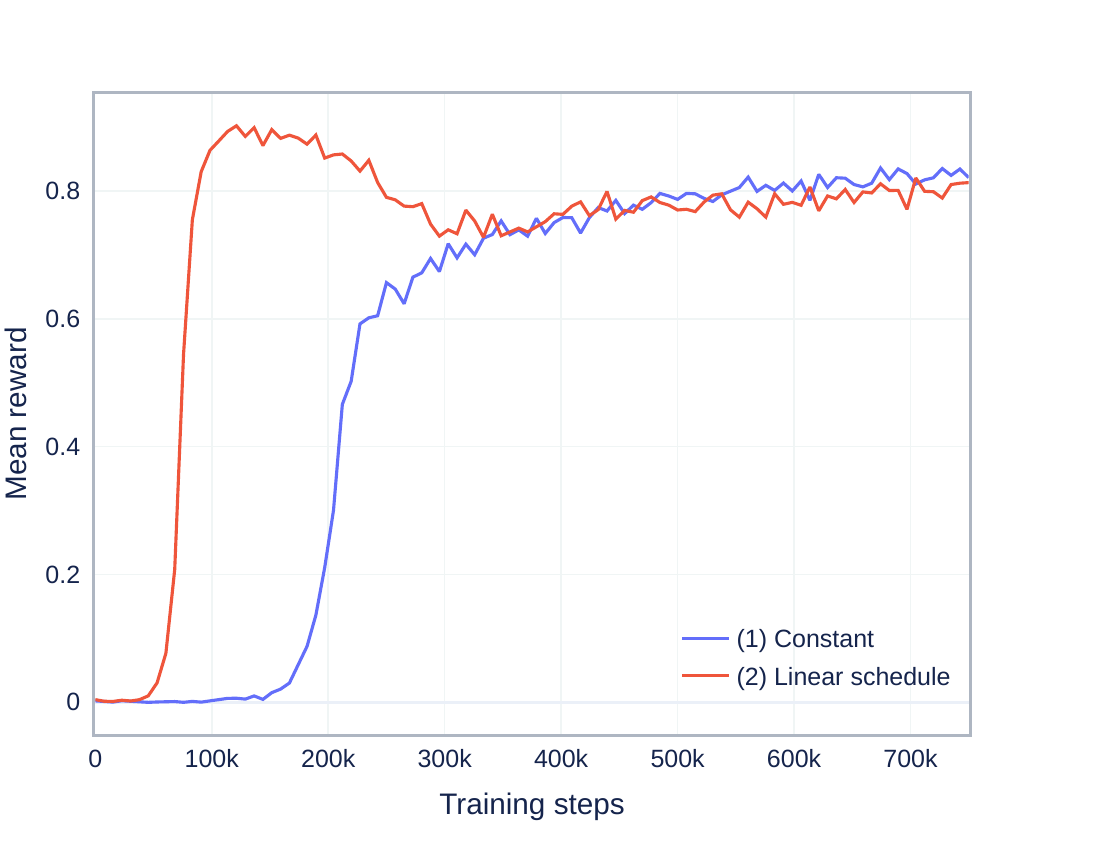}}
      \caption{}
      \label{fig:compare-mrilg-noise-sched-rwd}
    \end{subfigure}
    \hfill
    \begin{subfigure}{.49\textwidth}
      \centering
      \centerline{\includegraphics[width=\linewidth]{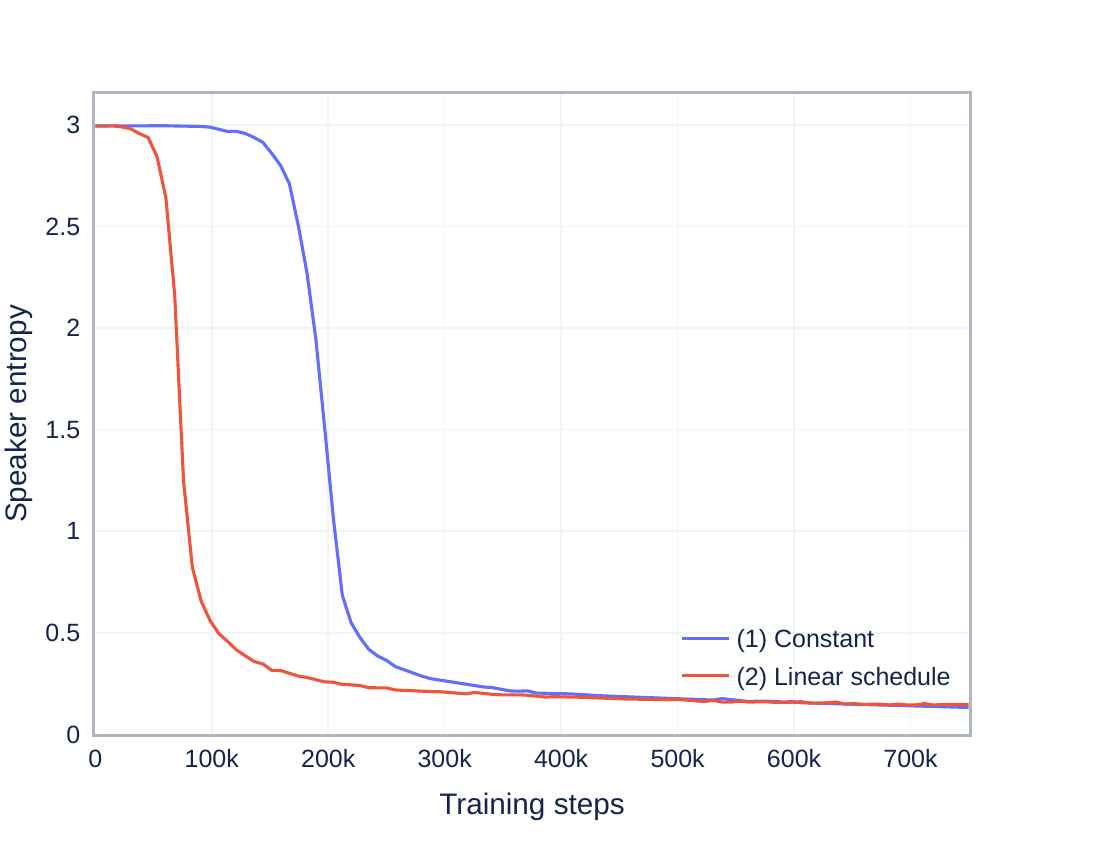}}
      \caption{}
      \label{fig:compare-mrilg-noise-sched-ent}
    \end{subfigure}
    \end{center}
    \caption{Mean reward (\subref{fig:compare-mrilg-noise-sched-rwd}) and the mean value for the Speaker's policy entropy (\subref{fig:compare-mrilg-noise-sched-ent}) on the ImageNet dataset of different implementations of NLG (RL), with \(|\sC|=1024\). The LG (RL) implementations differ on how to schedule the noise \(\lambda\) value during training: (1) \(\lambda\) remains constant at the end value \(0.5\); (2) \(\lambda\) linearly scales from \(0\) to \(0.5\) during the first \SI{300}{\kilo{}} steps, staying at this value afterward.}
    \label{fig:compare-mrilg-noise-sched}
    \vspace{-2ex}
\end{figure*}

We analyze different approaches to schedule the addition of noise to the communication channel for NLG. Our objective resides on understanding if there is an impact on performance if the noise is introduced gradually instead of fixing it at the established value from the start of training. As such, we introduce two ways to schedule the noise threshold during training:
\begin{enumerate*}
  \item the noise value is fixed at its final value \(\lambda\) during the entire training procedure;
  \item a scheduler linearly scales the noise from \(0\) to \(\lambda\) during the first \(40\%\) of the training steps.
\end{enumerate*}
\Cref{fig:compare-mrilg-noise-sched} illustrates both experiments, (1) and (2), when fixing \(\lambda=0.5\). We observe that by gradually scheduling the noise, instead of fixing it at the specified end value, the mean reward leaves the zone near \(0\) considerably earlier, meaning the agents have an easier time starting coordinating on a shared communication protocol, see \Cref{fig:compare-mrilg-noise-sched-rwd}. Having the noise increasing gradually allows for the pair of agents to first focus on creating a common language and, only after, slowly adjust to the noise in the message. Note that introducing noise makes the environment highly stochastic since the noise introduced will mask random tokens of the message. Additionally, the Listener is the only one capable of perceiving this modification, which makes the coordination between the pair more challenging. \cref{fig:compare-mrilg-noise-sched-ent} also corroborates our analysis, where in (1), the Speaker's entropy only starts decreasing after \SI{120}{\kilo{}} steps, as opposed to (2), where the same happens just after \SI{50}{\kilo{}} steps. As such, a communication protocol starts to emerge much sooner in (2), where the Speaker begins conveying more stable and regular messages. Nonetheless and as we observed in \Cref{fig:compare-mrilg-noise-sched-rwd}, scheduling the noise threshold does not affect the final mean reward obtained, but only helps with sample efficiency.

\section{Evaluation's Additional Results} \label{app:add-results}

\begin{figure*}[!t]
\begin{minipage}[!t]{1\textwidth}
\centering
\includegraphics[width=0.55\linewidth]{figures/legend}
\caption*{}
\end{minipage}
\begin{minipage}[!t]{1\textwidth}
\vspace{-7ex}
\centering
\includegraphics[width=1\linewidth]{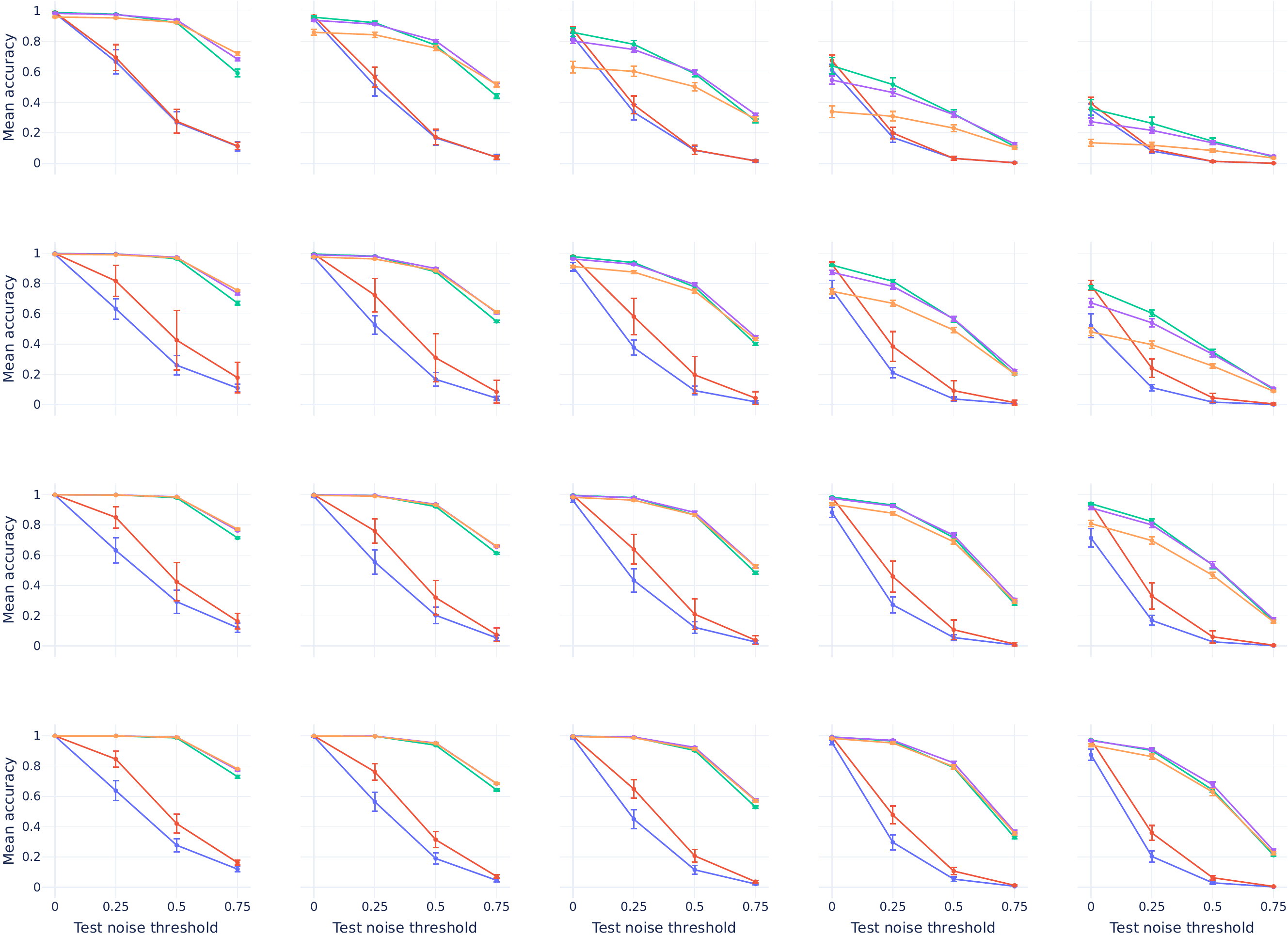}
\caption{Mean test accuracy for all LG variants on the ImageNet dataset. Each row contains experiments trained with a particular candidate size, \(|\sC|\). From the first to the last row, the candidate sizes are \(\{16, 64, 256, 1024\}\). Each column depicts the test accuracy obtained by all experiments when setting the candidate size, \(|\sC_\text{test}|\), to a specific value at test time (the test candidate set size can be different from the value used during training). From the first to the last column, the test candidate sizes are \(\{16, 64, 256, 1024, 4096\}\). We report the average (plus SD) over 10 seeds.}
\label{fig:compare-lg-imagenet-all}
\end{minipage}
\end{figure*}

\begin{figure*}[!t]
\begin{minipage}[!t]{1\textwidth}
\centering
\includegraphics[width=0.55\linewidth]{figures/legend}
\caption*{}
\end{minipage}
\begin{minipage}[!t]{1\textwidth}
\vspace{-7ex}
\centering
\includegraphics[width=1\linewidth]{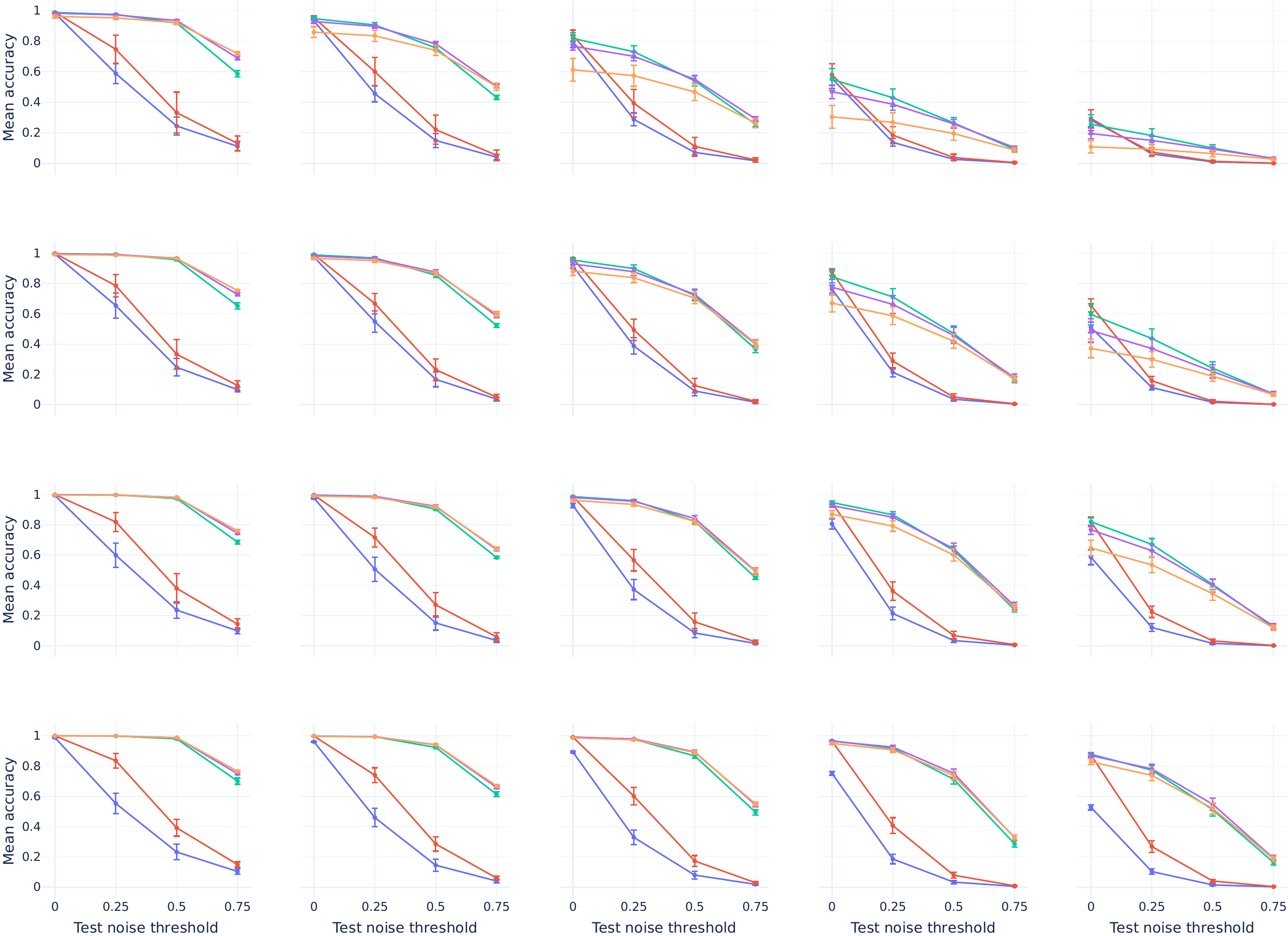}
\caption{Mean test accuracy for all LG variants on the CelebA dataset. Each row contains experiments trained with a particular candidate size, \(|\sC|\). From the first to the last row, the candidate sizes are \(\{16, 64, 256, 1024\}\). Each column depicts the test accuracy obtained by all experiments when setting the candidate size, \(|\sC_\text{test}|\), to a specific value at test time (the test candidate set size can be different from the value used during training). From the first to the last column, the test candidate sizes are \(\{16, 64, 256, 1024, 4096\}\). We report the average (plus SD) over 10 seeds.}
\label{fig:compare-lg-celeba-all}
\end{minipage}
\end{figure*}

\begin{figure*}[!t]
\begin{minipage}[!t]{1\textwidth}
\centering
\includegraphics[width=0.55\linewidth]{figures/legend}
\caption*{}
\end{minipage}
\begin{minipage}[!t]{1\textwidth}
\vspace{-7ex}
\centering
\includegraphics[width=1\linewidth]{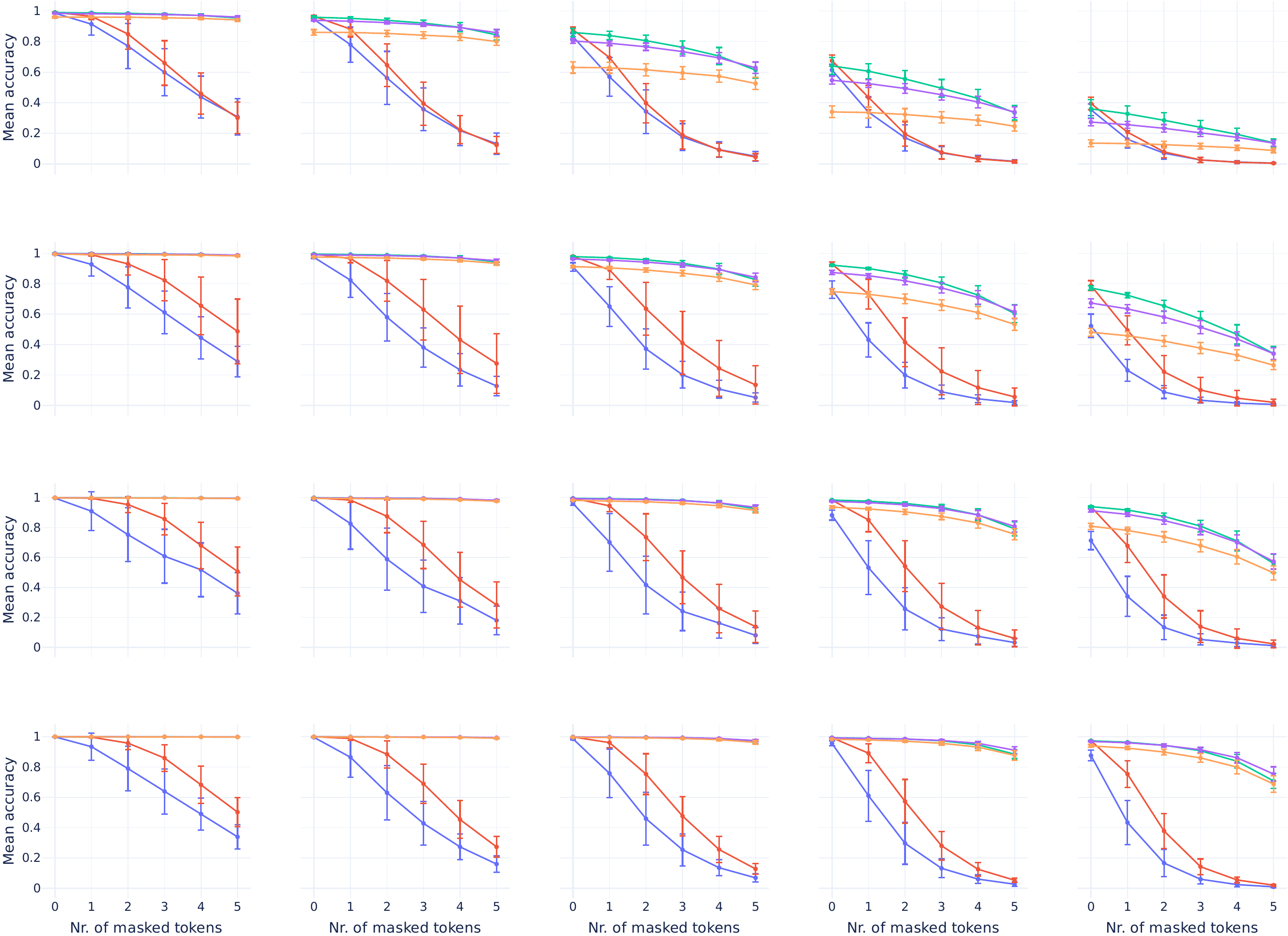}
\caption{Mean test accuracy, when randomly masking a fixed number of message tokens (see \Cref{sec:eval-mess-struct}), for all LG variants on the ImageNet dataset. Each row contains experiments trained with a particular candidate size, \(|\sC|\). From the first to the last row, the candidate sizes are \(\{16, 64, 256, 1024\}\). Each column depicts the test accuracy obtained by all experiments when setting the candidate size, \(|\sC_\text{test}|\), to a specific value at test time (the test candidate set size can be different from the value used during training). From the first to the last column, the test candidate sizes are \(\{16, 64, 256, 1024, 4096\}\). We report the average (plus SD) over 10 seeds and 10 different combinations of masked tokens (except when no tokens are masked).}
\label{fig:compare-lg-mess-imagenet-all}
\end{minipage}
\end{figure*}

\begin{figure*}[!t]
\begin{minipage}[!t]{1\textwidth}
\centering
\includegraphics[width=0.55\linewidth]{figures/legend}
\caption*{}
\end{minipage}
\begin{minipage}[!t]{1\textwidth}
\vspace{-7ex}
\centering
\includegraphics[width=1\linewidth]{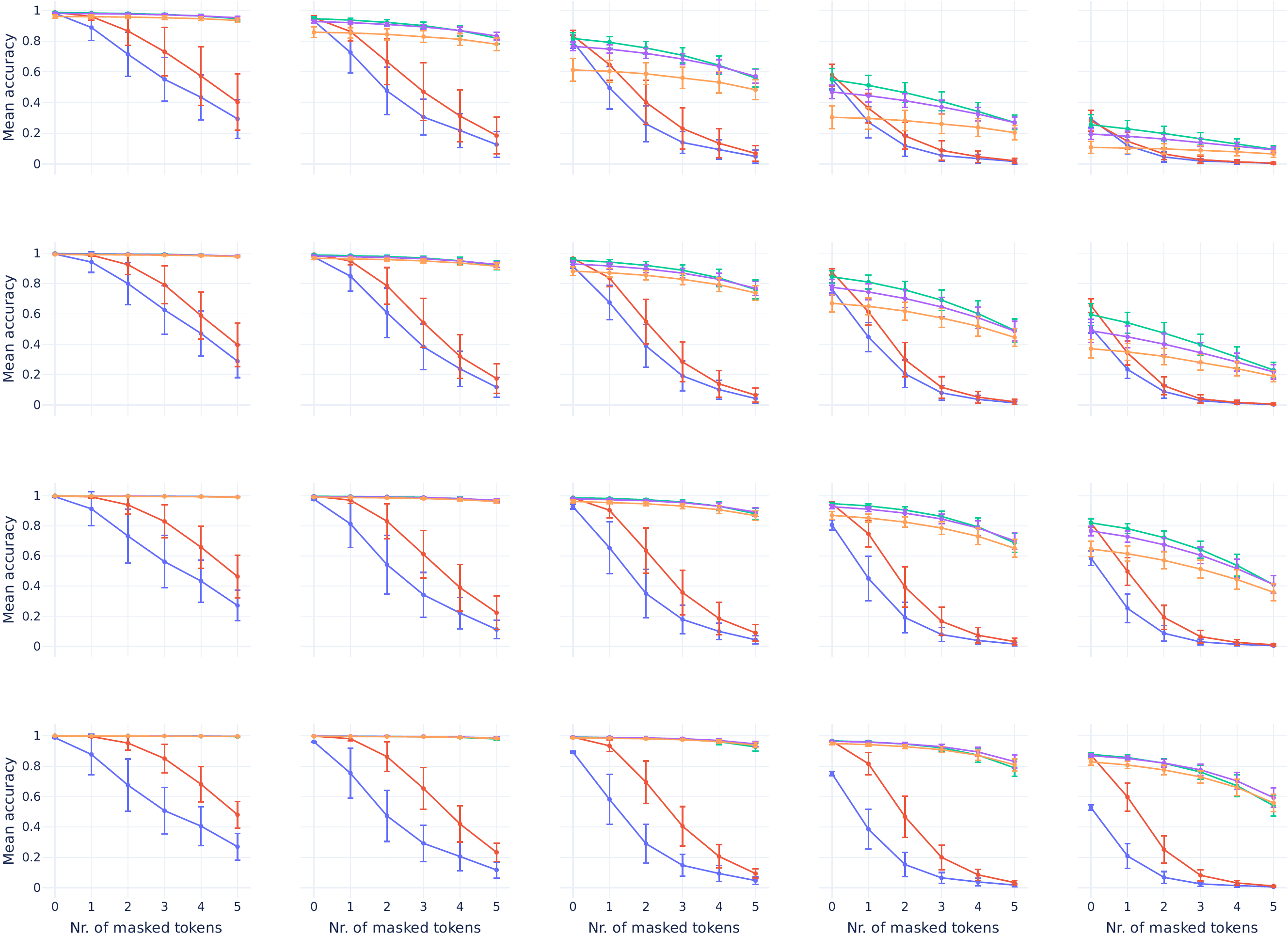}
\caption{Mean test accuracy, when randomly masking a fixed number of message tokens (see \Cref{sec:eval-mess-struct}), for all LG variants on the CelebA dataset. Each row contains experiments trained with a particular candidate size, \(|\sC|\). From the first to the last row, the candidate sizes are \(\{16, 64, 256, 1024\}\). Each column depicts the test accuracy obtained by all experiments when setting the candidate size, \(|\sC_\text{test}|\), to a specific value at test time (the test candidate set size can be different from the value used during training). From the first to the last column, the test candidate sizes are \(\{16, 64, 256, 1024, 4096\}\). We report the average (plus SD) over 10 seeds and 10 different combinations of masked tokens (except when no tokens are masked).}
\label{fig:compare-lg-mess-celeba-all}
\end{minipage}
\end{figure*}

\begin{figure*}[!t]
\begin{minipage}[!t]{1\textwidth}
\centering
\includegraphics[width=0.55\linewidth]{figures/legend}
\caption*{}
\end{minipage}
\begin{minipage}[!t]{1\textwidth}
\vspace{-7ex}
\centering
\includegraphics[width=1\linewidth]{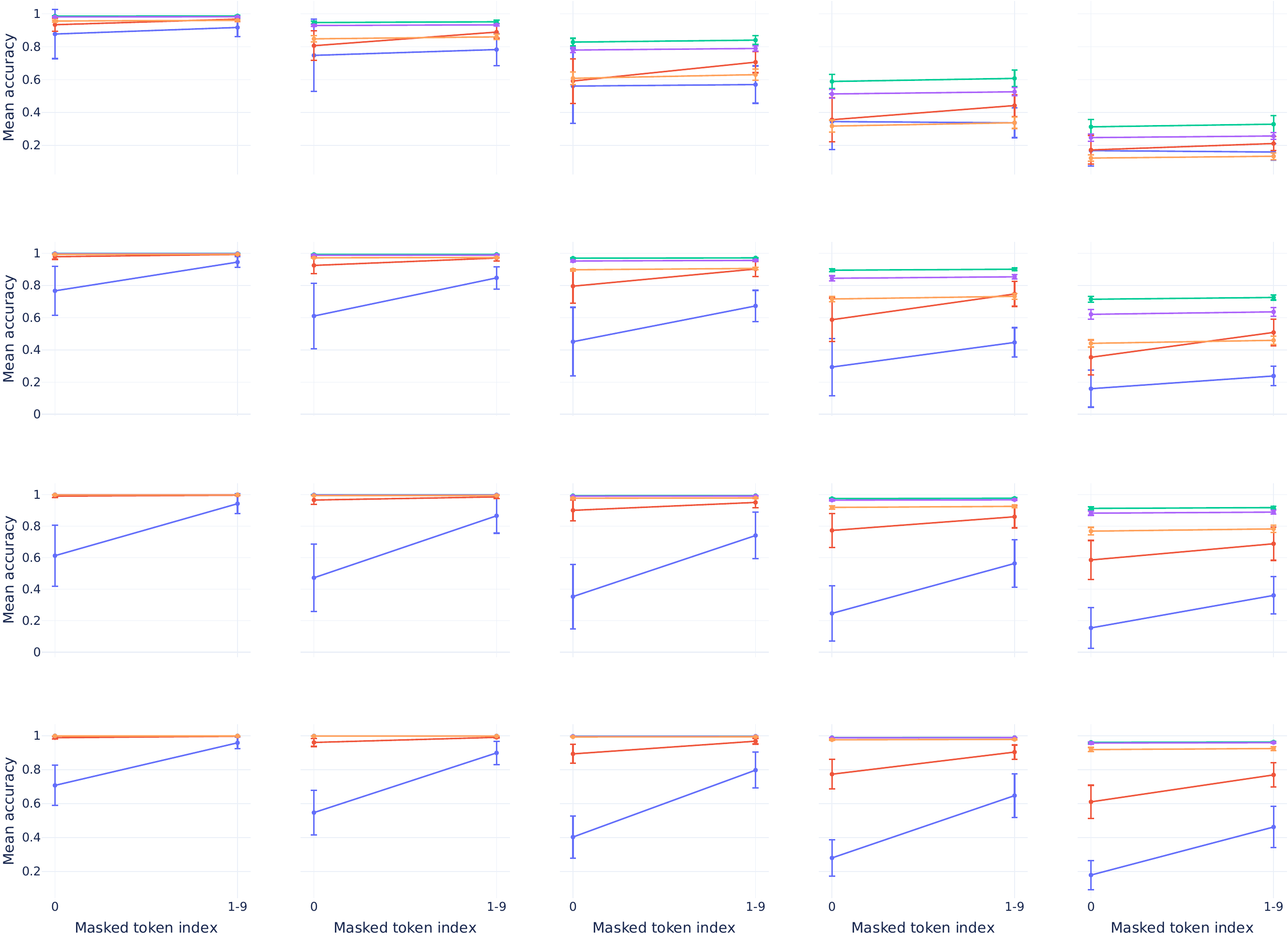}
\caption{Comparison of the mean test accuracy when masking the first message token vs.\ masking any other token, for all LG variants on the ImageNet dataset. Each row contains experiments trained with a particular candidate size, \(|\sC|\). From the first to the last row, the candidate sizes are \(\{16, 64, 256, 1024\}\). Each column depicts the test accuracy obtained by all experiments when setting the candidate size, \(|\sC_\text{test}|\), to a specific value at test time (the test candidate set size can be different from the value used during training). From the first to the last column, the test candidate sizes are \(\{16, 64, 256, 1024, 4096\}\). We report the average (plus SD) over 10 seeds.}
\label{fig:compare-lg-mess-1-imagenet-all}
\end{minipage}
\end{figure*}

\begin{figure*}[!t]
\begin{minipage}[!t]{1\textwidth}
\centering
\includegraphics[width=0.55\linewidth]{figures/legend}
\caption*{}
\end{minipage}
\begin{minipage}[!t]{1\textwidth}
\vspace{-7ex}
\centering
\includegraphics[width=1\linewidth]{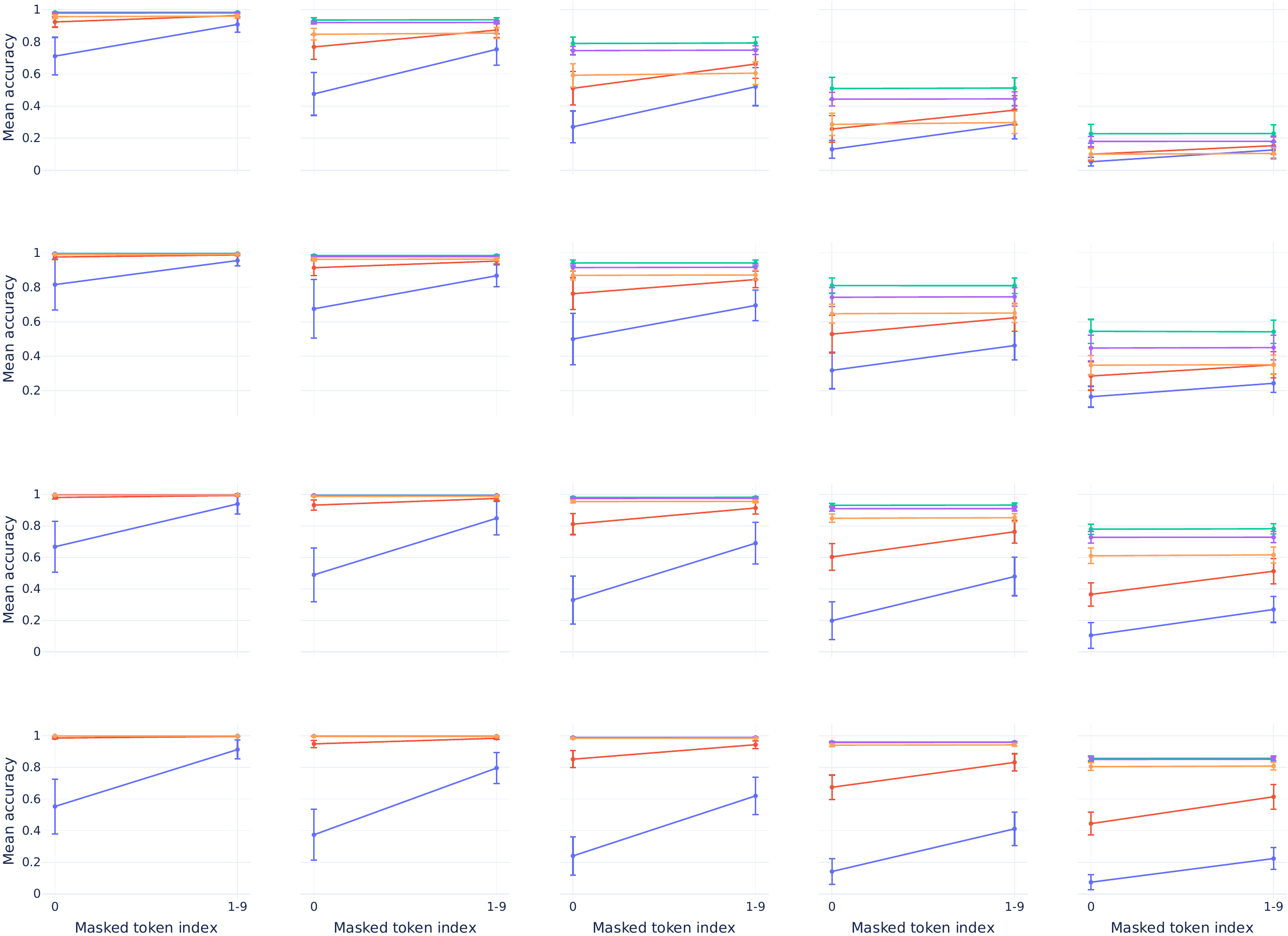}
\caption{Comparison of the mean test accuracy when masking the first message token vs.\ masking any other token, for all LG variants on the CelebA dataset. Each row contains experiments trained with a particular candidate size, \(|\sC|\). From the first to the last row, the candidate sizes are \(\{16, 64, 256, 1024\}\). Each column depicts the test accuracy obtained by all experiments when setting the candidate size, \(|\sC_\text{test}|\), to a specific value at test time (the test candidate set size can be different from the value used during training). From the first to the last column, the test candidate sizes are \(\{16, 64, 256, 1024, 4096\}\). We report the average (plus SD) over 10 seeds.}
\label{fig:compare-lg-mess-1-celeba-all}
\end{minipage}
\end{figure*}

\begin{figure*}[!t]
\begin{minipage}[!t]{1\textwidth}
\centering
\includegraphics[width=0.55\linewidth]{figures/legend}
\caption*{}
\end{minipage}
\begin{minipage}[!t]{1\textwidth}
\vspace{-7ex}
\centering
\includegraphics[width=1\linewidth]{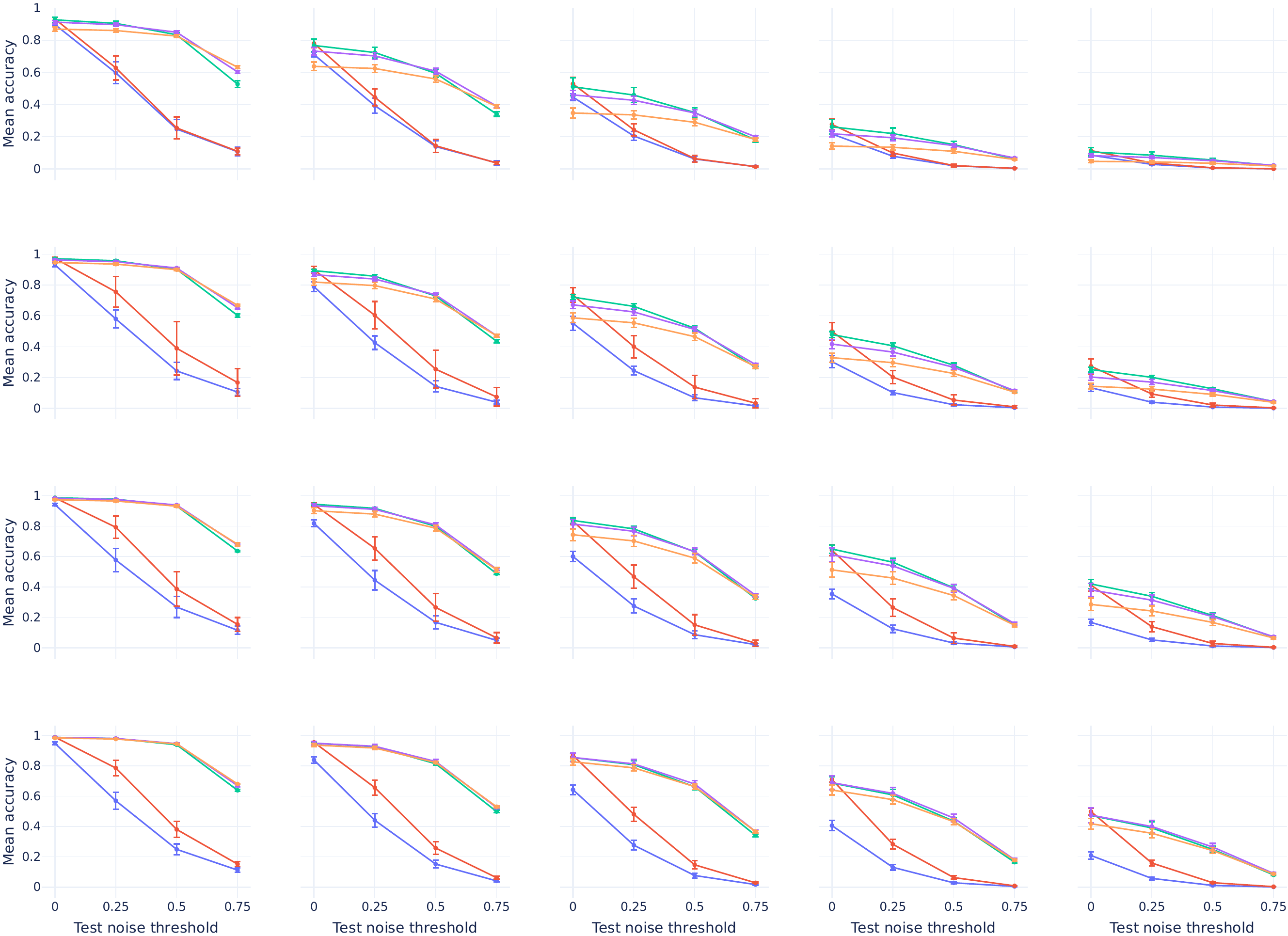}
\caption{Mean test accuracy, when noise is added to the inputs (see \Cref{sec:eval-ext-noise}), for all LG variants on the ImageNet dataset. Each row contains experiments trained with a particular candidate size, \(|\sC|\). From the first to the last row, the candidate sizes are \(\{16, 64, 256, 1024\}\). Each column depicts the test accuracy obtained by all experiments when setting the candidate size, \(|\sC_\text{test}|\), to a specific value at test time (the test candidate set size can be different from the value used during training). From the first to the last column, the test candidate sizes are \(\{16, 64, 256, 1024, 4096\}\). We report the average (plus SD) over 10 seeds.}
\label{fig:compare-lg-input-imagenet-all}
\end{minipage}
\end{figure*}

\begin{figure*}[!t]
\begin{minipage}[!t]{1\textwidth}
\centering
\includegraphics[width=0.55\linewidth]{figures/legend}
\caption*{}
\end{minipage}
\begin{minipage}[!t]{1\textwidth}
\vspace{-7ex}
\centering
\includegraphics[width=1\linewidth]{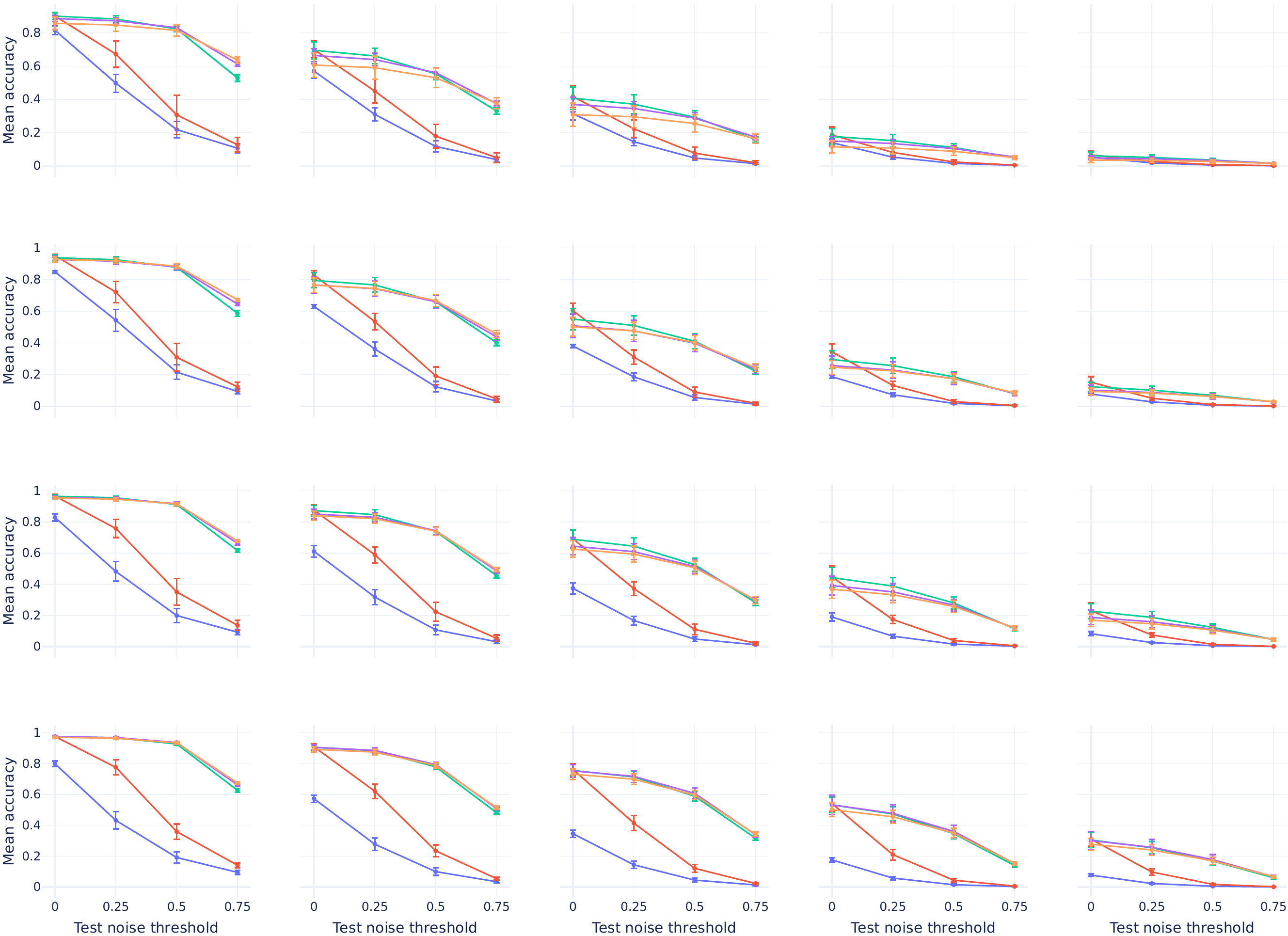}
\caption{Mean test accuracy, when noise is added to the inputs (see \Cref{sec:eval-ext-noise}), for all LG variants on the CelebA dataset. Each row contains experiments trained with a particular candidate size, \(|\sC|\). From the first to the last row, the candidate sizes are \(\{16, 64, 256, 1024\}\). Each column depicts the test accuracy obtained by all experiments when setting the candidate size, \(|\sC_\text{test}|\), to a specific value at test time (the test candidate set size can be different from the value used during training). From the first to the last column, the test candidate sizes are \(\{16, 64, 256, 1024, 4096\}\). We report the average (plus SD) over 10 seeds.}
\label{fig:compare-lg-input-celeba-all}
\end{minipage}
\end{figure*}

In this section, we present additional results to study and compare changes in performance as the number of candidates increases during training, as well as the impact of the noise level in NLG. For additional results regarding the evaluation in \Cref{sec:eval-comm}, please refer to \Cref{fig:compare-lg-imagenet-all} for the results obtained during the test phase for all game variants using the ImageNet dataset~\citep{ILSVRC15} and fixing different noise levels. The same results for the CelebA dataset~\citep{liu2015faceattributes} appear in \Cref{fig:compare-lg-celeba-all}. Regarding \Cref{sec:eval-mess-struct}, we present the results obtained in all experiments in \Cref{fig:compare-lg-mess-imagenet-all,fig:compare-lg-mess-celeba-all} as a function of the number of masked tokens, for the ImageNet and CelebA datasets, respectively. Additionally, we showcase particular results when masking a single token, aiming to show the difference in performance for the LG (S) and LG (RL) variants when the first message token is masked vs.\ any other token. \Cref{fig:compare-lg-mess-1-imagenet-all,fig:compare-lg-mess-1-celeba-all} depict these results for the ImageNet and CelebA datasets, respectively. Finally, the results obtained for all experiments where noise also appears in the inputs given to the Speaker and Listener at test time (\Cref{sec:eval-ext-noise}) appear in \Cref{fig:compare-lg-input-imagenet-all,fig:compare-lg-input-celeba-all}, for the ImageNet and CelebA datasets, respectively.

\section{Ease \& Transfer Learning} \label{app:etl}
Following~\citet{chaabouni2022emergent}, we also utilize secondary tasks to evaluate the generality and applicability of the languages learned in each LG variant. This evaluation procedure is called Ease \& Transfer Learning (ETL). The primary purpose of ETL is to evaluate if a preceding communication protocol, which was grounded in a particular task, can be used by new agents to solve new tasks. If this proposition holds, we get solid and empirical evidence about the generality of the emergent language since it can encode high-level descriptive information to solve different types of tasks related to the original environment.

ETL comprises four tasks: \emph{Discrimination}, \emph{Classification}, \emph{Attribute}, and \emph{Reconstruction}. We use the first two tasks to evaluate instances of games played with the ImageNet dataset and all four tasks when using the CelebA dataset. Due to the lack of description and reproducibility in~\citet{chaabouni2022emergent}, we thoroughly report the objective of each task accompanied by the training regime and hyper-parameters adopted.

We define new Listener agents to play the new tasks with a previously trained Speaker from one of the original LG games. Additionally, we freeze the Speaker weights, meaning the communication protocol remains fixed throughout training and evaluation of the ETL task. As such, we generate a new discrete dataset that depends on the Speaker (converting each image into a message). On this account, the only learning agent is the Listener. The Listener architecture has a common module used to process the received message that does not depend on the ETL task. This module is equal to the one implemented in the Listener architecture used in the LG, see \Cref{sec:meth-arch}. The difference between architectures occurs in the Listener's head module, as we will describe when introducing each task.

Similar to the tests employed for the LG games, we add a new dimension for the ETL tasks where we perturb the communication channel by adding noise. Accordingly, by varying this new dimension, we can analyze how it affects the generalization capabilities of the communication protocols. We will now describe each ETL task in detail, and afterward, the evaluation performed.

\subsection{Discrimination Task} \label{app:etl-discr}
This task is similar to the original LG task, where we add noise to the candidates and the Speaker's input, see \Cref{sec:eval-ext-noise}. The noise follows a Gaussian distribution with mean \(\mu\) and a standard deviation \(\sigma\), see \Cref{table:etl-classif-hyperparams}. Since we now have a fixed dataset, we regard this task as a supervised problem where we design the Listener to have a similar architecture to the one used in the original LG (S). As such, we define a distance loss to train the Listener to recognize similarities between messages and the correct candidate and, simultaneously, dissimilarities to opposing ones. To create a more challenging task, we set the size of the candidates' set, \(|\sC|\), to 4096.

\subsubsection{Listener's Head Architecture}
We plug the head defined for the LG (S) (\Cref{app:arch-list-ss}), where a distance function computes similarities between the message and each candidate. Then, a softmax function converts similarities into a categorical distribution to train using InfoNCE loss~\citep{oord2018representation}.

\subsubsection{Hyperparameters}
The hyperparameters adjusted for the Discrimination task appear in \Cref{table:etl-discr-hyperparams}.

\begin{table}[t]
\centering
\caption{Hyperparameters modified (from \Cref{table:etl-common-hyperparams}) for the Discrimination task.}
\label{table:etl-discr-hyperparams}
\begin{tabular}{lc}
\toprule
hyperparameter & value\\
\midrule
\(\mu\) & \SI{0}{} \\
\(\sigma\) & \SI{0.5}{} \\
\(|\sC|\) & \SI{4096}{} \\
batch size & \SI{4096}{} (\(=|\sC|\)) \\
\bottomrule
\end{tabular}
\end{table}

\subsection{Classification Task} \label{app:etl-classif}
With the classification task, we aim to evaluate whether the emerging language has any valuable information to identify the category of an image. In this task, the Speaker receives an image and encodes a message to the Listener. Afterward, the Listener tries to determine the class of the image while receiving only the message sent by the Speaker as input. For the ImageNet dataset, we use the label of each image as the target and the identity for the CelebA dataset.

\subsubsection{Listener's Head Architecture} \label{app:etl-classif-list}
Since the objective of the Classification task is to identify the class of the image received by the Speaker, the Listener's head architecture consists of only one MLP where the final layer has the same size as the number of classes (1000 and 10177, for the ImageNet and CelebA datasets, respectively) and is then used for prediction after applying softmax. The MLP mentioned above also has one hidden layer of size 256 and uses ReLU as the activation function.

\subsubsection{Hyperparameters}
\Cref{table:etl-classif-hyperparams} exposes the hyperparameters modified explicity for the Classification task.

\begin{table}[t]
\centering
\caption{Hyperparameters modified (from \Cref{table:etl-common-hyperparams}) for the Classification task.}
\label{table:etl-classif-hyperparams}
\begin{tabular}{lc}
\toprule
hyperparameter & value\\
\midrule
training steps (for ImageNet dataset only)& \SI{30}{\kilo{}} \\
\bottomrule
\end{tabular}
\end{table}

\subsection{Attribute Task} \label{app:etl-attr}
This task is very similar to the classification task, where instead of predicting the image class, the Listener tries to predict secondary binary information related to facial attributes, such as gender, hair color, age, etc.

\subsubsection{Listener's Head Architecture}
The Listener architecture is similar to the one described for the classification task, \cref{app:etl-classif-list}. In this case, we have an independent model to classify each attribute, where the loss adopted is the binary cross-entropy loss.

\subsection{Reconstruction Task} \label{app:etl-reconst}
The final ETL task is the Reconstruction task, where the primary purpose is reconstructing the original image. The Listener's head promotes a convolutional decoder to reconstruct the original input image (Speaker's input) by taking into account only the discrete message sent by the Speaker. See \Cref{app:etl-reconst-list} for a complete description of the Listener's head architecture.

\subsubsection{Listener's Head Architecture} \label{app:etl-reconst-list}
The Listener's head architecture encapsulates a decoder to reconstruct the original image. Since the hidden message features \(\vz_{T,\phi}\) are the input of this module and have only one dimension, the first step addresses the redimension of \(\vz_{T,\phi}\) into a 3D shape. As such, a single linear layer of size \(2048\) upsamples \(\vz_{T,\phi}\), where the resulting dimension is \((4, 4, 128)\) (with layout \texttt{(height, width, channel)}) after reshaping, as \(128\times4\times4=2048\). Afterward, a decoding procedure aims to upsample the features' first two dimensions. The procedure occurs four times sequentially, and at each step, we upsample the dimensions of the current features by doubling the width and height, using nearest neighbor interpolation. Next, the upsampled features are given to a convolution layer followed by a ReLU to reduce the number of channels as part of the reconstruction process. The number of output channels is \(64, 32, 16, 16\) for each convolution, respectively. Finally, a concluding convolution layer, followed by \(\text{tanh}\), outputs the decoded image with dimensions \((64, 64, 3)\). With \(\text{tanh}\) as the final activation, the reconstructed image is the standardized version of the original one after normalization using the ImageNet coefficients~\citep{he2016deep}. Furthermore, the parameters for all convolutions are equal: kernel size of \((3\times3)\), padding \texttt{same}, stride \(1\), and without bias. Additionally, we use the MSE loss to train the Listener.

Finally, we employ the same optimizer parameters as the original work~\citep{chaabouni2022emergent}, where gradients are first clipped by their global norm when it is above a maximum threshold \(G_\text{max}\)~\citep{pascanu2013difficulty}. Then, \texttt{adam} with weight decay regularization~\citep{loshchilov2017decoupled} computes the update rule. For more information on the hyper-parameters used, see \Cref{app:etl-reconst-params}.

\subsubsection{Hyperparameters }\label{app:etl-reconst-params}
Please refer to \Cref{table:etl-recons-hyperparams} to consult the modified hyperparameters for the Reconstrunction task.

\begin{table}[t]
\centering
\caption{Hyperparameters modified (from \Cref{table:etl-common-hyperparams}) for the Reconstruction task.}
\label{table:etl-recons-hyperparams}
\begin{tabular}{lc}
\toprule
hyperparameter & value\\
\midrule
\(G_\text{max}\) & 500 \\
Listener's optimizer & \texttt{adam w/ weight decay reg.} \\
Listener's optimizer \texttt{lr} & \SI{3e-4}{} \\
Listener's optimizer \texttt{b1} & \SI{0.9}{} \\
Listener's optimizer \texttt{b2} & \SI{0.9}{} \\
Listener's optimizer \texttt{weight decay} & \SI{0.01}{} \\
\bottomrule
\end{tabular}
\end{table}

\subsection{Common Hyperparameters} \label{app:etl-params}

\Cref{table:etl-common-hyperparams} describes the hyperparameters commonly used by all ETL tasks.

\begin{table}[t]
\centering
\caption{Default hyperparameters used in all ETL tasks. Note that each ETL task can change some parameter values, as specified in \Cref{app:etl-discr,app:etl-classif,app:etl-reconst}}
\label{table:etl-common-hyperparams}
\begin{tabular}{lc}
\toprule
hyperparameter & value\\
\midrule
training steps & \SI{10}{\kilo{}} \\
batch size & \SI{128}{} \\
\(|\mathcal{W}|\) & \SI{20}{} \\
\(T\) & \SI{10}{} \\
Listener's optimizer & \texttt{adam} \\
Listener's optimizer lr & \SI{1e-3}{} \\
\bottomrule
\end{tabular}
\end{table}

\subsection{Evaluation}
In this sub-section, we present all results obtained by evaluating all ETL tasks regarding each LG variant trained. Similarly to the primary evaluation, \Cref{sec:eval}, we describe the results incrementally, where we start by comparing both implementations (S and RL) for the original LG. Afterward, we introduce noise, adding NLG to the comparison. Additionally, we compare the influence of the candidate set size in all LG variants. Regarding the metrics adopted, we report accuracy for all tasks except for the Reconstruction task, where we report the obtained final loss.

\subsubsection{SS \& RL implementations for LG} \label{app:etl-eval-lg}
We start the ETL evaluation by showcasing and comparing the performance of both prototypes of the LG (S and RL) when training without noise. We report the test performance for all applicable ETL tasks for the ImageNet and CelebA datasets in \Cref{table:compare_etl_lg_imagenet,table:compare_etl_lg_celeba}, respectively. As we will see next, our newly proposed variant, LG (RL), reports better results than LG (S).

Looking at the discrimination task, we continue to observe the same behavior encountered in the LG evaluation, \Cref{sec:eval}, where the RL version is superior to S. From this new perspective, we obtain further evidence that the communication protocol from the RL variant is more general than the S counterpart. In this case, the messages from the RL version are more appropriate to describe the noisy image, which facilitates the selection process by the new Listener agent. As such, the communication protocol from LG (RL) is more capable of dealing with other noise sources, such as detecting noise in the Speaker's input image.

Regarding the classification task, we see deficient performance for both datasets, with around \(0.14\) and \(0.01\) test accuracy for ImageNet and CelebA, respectively. Not only that, but we also observe the same results for all LG variants testing with or without noise. In \Cref{app:etl-classif-analysis}, we propose a simple explanation for these results that emerge from the messages' content and structure.

Taking a closer look at the results for the CelebA dataset, we observe similar and considerably high results (around \(0.85\)) for the attribute task, which does not depend on the LG version, S or RL.
We can then assess that there is a specific batch of images where the Listener agent cannot decode any relevant information from the message to accurately classify the attribute.

For the final task, Reconstruction, we showcase the loss function obtained in the test set. We observe that LG (RL) obtains lower loss than LG (S). To complement this analysis, \Cref{fig:etl-reconstruct-all-lg-s,fig:etl-reconstruct-all-lg-rl} show the reconstruction of several images for the LG (S) and LG (RL) variants, respectively. Both versions can reconstruct some particularities of the original image such as sunglasses, hat, hair color, and gender. On the other hand, information about orientation, age, or skin tone are not encoded in the messages. Additionally, we can clearly observe different facial layouts and different hair color, reconstructed by LG (S) and LG (RL), for the same image.

\begin{table*}[t]
\linespread{0.6}\selectfont\centering
\caption{Test accuracy, with SD, of every ETL task trained using, the ImageNet dataset and over 10 seeds. During the LG training (before ETL), we fixed \(|\sC|\) to \(1024\).}
\label{table:compare_etl_lg_imagenet}
\centering
\begin{tabular}[t]{lrrr}
\toprule
Game & \multicolumn{1}{c}{\(|\sC|\)} & \multicolumn{2}{c}{ETL tasks} \\[1ex]\cmidrule(r){3-4}
 & & \multicolumn{1}{c}{Discrimination} & \multicolumn{1}{c}{Classification} \\
\midrule
LG {\scriptsize(S)} & \(1024\) & \longcell{\(0.58\)\\{\tiny(\(0.05\))}} & \longcell{\(0.14\)\\{\tiny(\(0.01\))}} \\[2.2ex]
LG {\scriptsize(RL)} & \(1024\) & \longcell{\(0.88\)\\{\tiny(\(0.01\))}} & \longcell{\(0.14\)\\{\tiny(\(0.01\))}} \\
\bottomrule
\end{tabular}
\end{table*}

\begin{table*}[t]
\linespread{0.6}\selectfont\centering
\caption{Test accuracy, with SD, of every ETL task trained, using the CelebA dataset and over 10 seeds. During the LG training (before ETL), we fixed \(|\sC|\) to \(1024\).}
\label{table:compare_etl_lg_celeba}
\centering
\begin{tabular}[t]{lrrrrr}
\toprule
Game & \multicolumn{1}{c}{\(|\sC|\)} & \multicolumn{4}{c}{ETL tasks} \\[1ex]\cmidrule(r){3-6}
 & & \multicolumn{1}{c}{Discrimination} & \multicolumn{1}{c}{Classification} & \multicolumn{1}{c}{Attribute} & \multicolumn{1}{c}{Reconstruction}\\
\midrule
LG {\scriptsize(S)} & \(1024\) & \longcell{\(0.44\)\\{\tiny(\(0.05\))}} & \longcell{\(0.01\)\\{\tiny(\(0.00\))}} & \longcell{\(0.86\)\\{\tiny(\(0.00\))}} & \longcell{\(5814\)\\{\tiny(\(74\))}} \\[2.2ex]
LG {\scriptsize(RL)} & \(1024\) & \longcell{\(0.62\)\\{\tiny(\(0.08\))}} & \longcell{\(0.01\)\\{\tiny(\(0.00\))}} & \longcell{\(0.85\)\\{\tiny(\(0.00\))}} & \longcell{\(5671\)\\{\tiny(\(192\))}} \\
\bottomrule
\end{tabular}
\end{table*}

\subsubsection{Introducing Noise} \label{app:etl-eval-noise}
To evaluate the robustness of the proposed LG variants to message perturbations, we add a new train/test regime to ETL where the communication channel disturbs messages sent by the Speaker by adding random noise, similar to NLG. To have an insightful analysis, we propose comparing LG (RL) and NLG in the case of train/test regimes without and with noise. We disregard LG (S) since LG (RL) achieved better results, see \Cref{app:etl-eval-lg}. Additionally, to have a contained analysis, we fixed the noise level to \(0.5\), and the candidate set size to 1024. The exhaustive version of these results appears at \Cref{table:etl_imagenet_0_eval,table:etl_celeba_0_eval,table:etl_imagenet_025_eval,table:etl_celeba_025_eval,table:etl_imagenet_05_eval,table:etl_celeba_05_eval,table:etl_imagenet_075_eval,table:etl_celeba_075_eval}.

The results for the ImageNet dataset appear in \Cref{table:compare_etl_noise_imagenet_det,table:compare_etl_noise_imagenet_noise} for the train/test regime without and with noise, respectively. Regarding the Discrimination task in the deterministic procedure (without noise), the accuracy obtained for NLG is slightly lower than in the LG (RL), around \(0.03\) below. Despite this minor decline, we continue to obtain evidence that the NLG generate robust communication protocols of similar efficiency to those induced by LG (RL). When introducing noise in the communication channel, we observe a performance increase, around \(0.05\), for NLG against LG (RL). Recall that the discrimination task adds continuous noise to the Speaker's input and candidates. Having noisy inputs creates a different type of uncertainty that the communication protocol is not optimized for. Even having robust agents to message noise practically does not impact the creation of more descriptive messages when a significant part of the noise comes from another source, in this case, combined with the input.

We perform a similar comparison for the CelebA dataset deployed in \Cref{table:compare_etl_noise_celeba_det,table:compare_etl_noise_celeba_noise}, for the deterministic and noisy regime, respectively. Focusing on the Discrimination task and the deterministic regime, LG (RL) and NLG have statistically equal accuracy values with mean \(0.62\), and \(0.60\), respectively.
When training and testing with noise, we observe a considerable drop in accuracy, around \(0.35\), for all variants. Similarly to what we discovered for the ImageNet dataset, the noise introduced to the inputs turns the discriminative task more challenging than the original game.

Regarding the Attribute task, all values vary between \(0.84\) and \(0.86\). As such, the performances for every LG variant are highly identical. We hypothesize that there is a considerable collection of samples where the identification of each attribute is accessible and effortlessly contemplated in the communication protocols. For the other minority of samples, the communication protocols cannot encode information about the attributes. For this reason, we observe a sharp rupture in accuracy for all methods, aroung \(85\%\).

\begin{figure*}[!t]
    \begin{center}
    \begin{subfigure}{1.\textwidth}
      \centering
      \centerline{\includegraphics[width=1.02\linewidth]{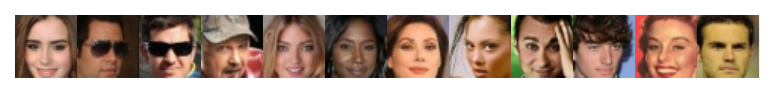}}
      \caption{Original CelebA images, randomly selected.}
      \label{fig:etl-reconstruct-original-ss}
    \end{subfigure}
    \begin{subfigure}{1\textwidth}
      \centering
      \centerline{\includegraphics[width=1.02\linewidth]{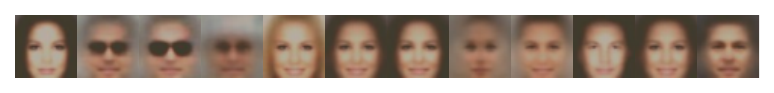}}
      \caption{LG (S), train/test ETL Reconstruction task without noise.}
      \label{fig:etl-reconstruct-lg-ss}
    \end{subfigure}
    \begin{subfigure}{1\textwidth}
    \centering
    \centerline{\includegraphics[width=1.02\linewidth]{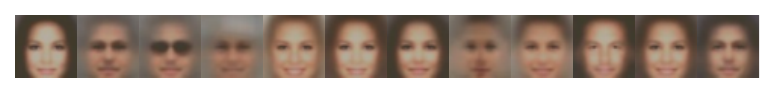}}
    \caption{LG (S), train/test ETL Reconstruction task with a noise level of \(0.25\).}
    \label{fig:etl-reconstruct-lg-ss-0.25}
    \end{subfigure}
    \begin{subfigure}{1\textwidth}
    \centering
    \centerline{\includegraphics[width=1.02\linewidth]{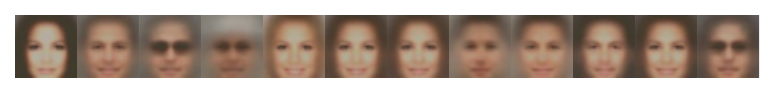}}
    \caption{LG (S), train/test ETL Reconstruction task with a noise level of \(0.5\).}
    \label{fig:etl-reconstruct-lg-ss-0.5}
    \end{subfigure}
    \begin{subfigure}{1\textwidth}
    \centering
    \centerline{\includegraphics[width=1.02\linewidth]{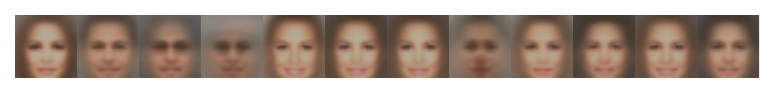}}
    \caption{LG (S), train/test ETL Reconstruction task with a noise level of \(0.75\).}
    \label{fig:etl-reconstruct-lg-ss-0.75}
    \end{subfigure}
    \end{center}
    \caption{Random outputs of ETL's Reconstruction task when using a Speaker trained in the LG (S). We report reconstructions when training and testing without (\Cref{sub@fig:etl-reconstruct-lg-ss}) and with noise (\Cref{sub@fig:etl-reconstruct-lg-ss-0.25,sub@fig:etl-reconstruct-lg-ss-0.5,sub@fig:etl-reconstruct-lg-ss-0.75}), in the ETL task. The faces shown in each column are a reconstruction of the original images available in the matching columns of \Cref{sub@fig:etl-reconstruct-original-ss}.}
    \label{fig:etl-reconstruct-all-lg-s}
    \vspace{-2ex}
\end{figure*}

\begin{figure*}[!t]
    \begin{center}
    \begin{subfigure}{1.\textwidth}
      \centering
      \centerline{\includegraphics[width=1.02\linewidth]{figures/reconstruct/original.pdf}}
      \caption{Original CelebA images, randomly selected.}
      \label{fig:etl-reconstruct-original-rl}
    \end{subfigure}
    \begin{subfigure}{1\textwidth}
        \centering
        \centerline{\includegraphics[width=1.02\linewidth]{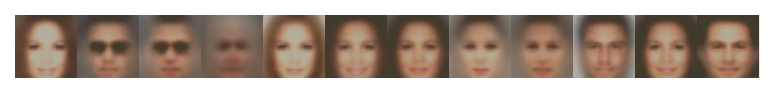}}
        \caption{LG (RL), train/test ETL Reconstruction task without noise.}
        \label{fig:etl-reconstruct-lg-rl}
    \end{subfigure}
    \begin{subfigure}{1\textwidth}
      \centering
      \centerline{\includegraphics[width=1.02\linewidth]{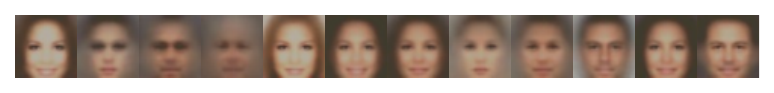}}
      \caption{LG (RL), train/test ETL Reconstruction task with a noise level of \(0.25\).}
      \label{fig:etl-reconstruct-lg-rl-0.25}
    \end{subfigure}
    \begin{subfigure}{1\textwidth}
      \centering
      \centerline{\includegraphics[width=1.02\linewidth]{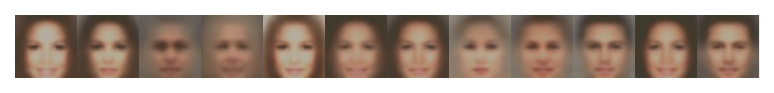}}
      \caption{LG (RL), train/test ETL Reconstruction task with a noise level of \(0.5\).}
      \label{fig:etl-reconstruct-lg-rl-0.5}
    \end{subfigure}
    \begin{subfigure}{1\textwidth}
      \centering
      \centerline{\includegraphics[width=1.02\linewidth]{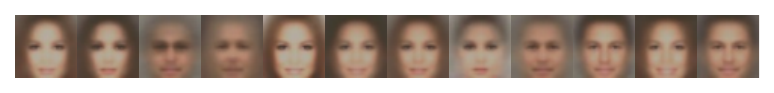}}
      \caption{LG (RL), train/test ETL Reconstruction task with a noise level of \(0.75\).}
      \label{fig:etl-reconstruct-lg-rl-0.75}
    \end{subfigure}
    \end{center}
    \caption{Random outputs of ETL's Reconstruction task when using a Speaker trained in the LG (RL). We report reconstructions when training and testing without (\Cref{sub@fig:etl-reconstruct-lg-rl}) and with noise (\Cref{sub@fig:etl-reconstruct-lg-rl-0.25,sub@fig:etl-reconstruct-lg-rl-0.5,sub@fig:etl-reconstruct-lg-rl-0.75}), in the ETL task. The faces shown in each column are a reconstruction of the original images available in the matching columns of \Cref{sub@fig:etl-reconstruct-original-rl}.}
    \label{fig:etl-reconstruct-all-lg-rl}
    \vspace{-2ex}
\end{figure*}

\begin{figure*}[!t]
    \begin{center}
    \begin{subfigure}{1.\textwidth}
      \centering
      \centerline{\includegraphics[width=1.02\linewidth]{figures/reconstruct/original.pdf}}
      \caption{Original CelebA images, randomly selected.}
      \label{fig:etl-nlg-reconstruct-original}
    \end{subfigure}
    \begin{subfigure}{1\textwidth}
      \centering
      \centerline{\includegraphics[width=1.02\linewidth]{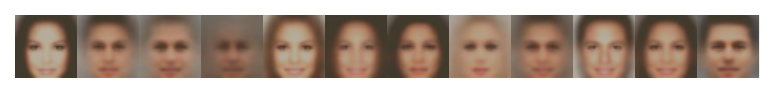}}
      \caption{NLG with \(\lambda=0.25\), train/test ETL Reconstruction task without noise.}
      \label{fig:etl-reconstruct-nlg-0.25}
    \end{subfigure}
    \begin{subfigure}{1\textwidth}
      \centering
      \centerline{\includegraphics[width=1.02\linewidth]{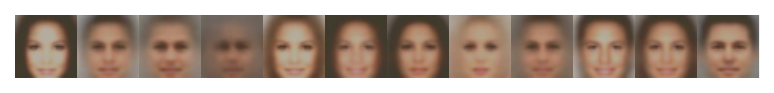}}
      \caption{NLG with \(\lambda=0.25\), train/test ETL Reconstruction task with a noise level of \(0.25\).}
      \label{fig:etl-reconstruct-nlg-0.25-0.25}
    \end{subfigure}
    \begin{subfigure}{1\textwidth}
      \centering
      \centerline{\includegraphics[width=1.02\linewidth]{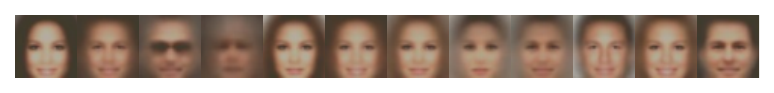}}
      \caption{NLG with \(\lambda=0.5\), train/test ETL Reconstruction task without noise.}
      \label{fig:etl-reconstruct-nlg-0.5}
    \end{subfigure}
    \begin{subfigure}{1\textwidth}
      \centering
      \centerline{\includegraphics[width=1.02\linewidth]{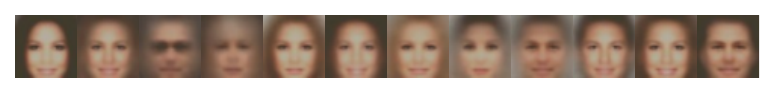}}
      \caption{NLG with \(\lambda=0.5\), train/test ETL Reconstruction task with a noise level of \(0.75\).}
      \label{fig:etl-reconstruct-nlg-0.5-0.5}
    \end{subfigure}
    \begin{subfigure}{1\textwidth}
      \centering
      \centerline{\includegraphics[width=1.02\linewidth]{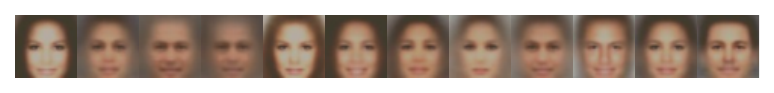}}
      \caption{NLG with \(\lambda=0.75\), train/test ETL Reconstruction task without noise.}
      \label{fig:etl-reconstruct-nlg-0.75}
    \end{subfigure}
    \begin{subfigure}{1\textwidth}
      \centering
      \centerline{\includegraphics[width=1.02\linewidth]{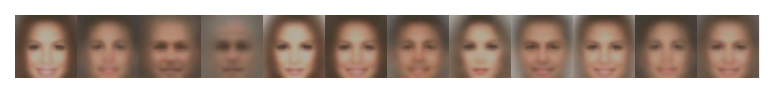}}
      \caption{NLG with \(\lambda=0.75\), train/test ETL Reconstruction task with a noise level of \(0.25\).}
      \label{fig:etl-reconstruct-nlg-0.75-0.75}
    \end{subfigure}
    \end{center}
    \caption{Random outputs of ETL's Reconstruction task when using a Speaker trained in the NLG. We report reconstructions when training and testing without (\Cref{sub@fig:etl-reconstruct-nlg-0.25,sub@fig:etl-reconstruct-nlg-0.5,sub@fig:etl-reconstruct-nlg-0.75}) and with noise (\Cref{sub@fig:etl-reconstruct-nlg-0.25-0.25,sub@fig:etl-reconstruct-nlg-0.5-0.5,sub@fig:etl-reconstruct-nlg-0.75-0.75}), in the ETL task. The faces shown in each column are a reconstruction of the original images available in the matching column of \Cref{sub@fig:etl-nlg-reconstruct-original}.}
    \label{fig:etl-reconstruct-all-nlg}
    \vspace{-2ex}
\end{figure*}

Finally, we compare the results obtained in the Reconstruction task. The losses for all LG variants are statistically equal for the test regime without and with noise. No LG variant demonstrated superior performance for this challenging task, which is obviously expected since the reconstructed images originate from discrete tokens (message content), making extracting unique information from each image extremely difficult. Additionally, in \Cref{fig:etl-reconstruct-all-lg-s,fig:etl-reconstruct-all-lg-rl,fig:etl-reconstruct-all-nlg}, we show some reconstructed images for the compared LG variants (LG (S), LG (RL), and NLG, respectively). The level of reconstruction detail is similar accross variants, where we can see all variants encoding hat, sunglasses, hair color, and ignoring attributes not relevant to the original discrimination task, such as skin tone and face orientation.

\begin{table*}[t]
\linespread{0.6}\selectfont\centering
\caption{Test accuracy, with SD, of every ETL task trained, using the ImageNet dataset and over 10 seeds. We do not apply noise during ETL's train and test phase. During the LG training (before ETL), we fixed \(|\sC|\) to \(1024\), and \(\lambda\) to \(0.5\).}
\label{table:compare_etl_noise_imagenet_det}
\centering
\begin{tabular}[t]{lrrrrr}
\toprule
Game & \(\lambda\) & \multicolumn{1}{c}{\(|\sC|\)} & \multicolumn{2}{c}{ETL tasks} \\[1ex]\cmidrule(r){4-5}
 & & & \multicolumn{1}{c}{Discrimination} & \multicolumn{1}{c}{Classification} \\
\midrule
LG {\scriptsize(RL)} & \(0\) & \(1024\) & \longcell{\(0.88\)\\{\tiny(\(0.01\))}} & \longcell{\(0.14\)\\{\tiny(\(0.01\))}} \\[2.2ex]
NLG & \(0.5\) & \(1024\) & \longcell{\(0.85\)\\{\tiny(\(0.03\))}} & \longcell{\(0.14\)\\{\tiny(\(0.02\))}} \\[2.2ex]
\bottomrule
\end{tabular}
\end{table*}

\begin{table*}[t]
\linespread{0.6}\selectfont\centering
\caption{Test accuracy, with SD, of every ETL task trained, using the CelebA dataset and over 10 seeds. During the LG training (before ETL), we fixed \(|\sC|\) to \(1024\). We do not apply noise during ETL's train and test phase. During the LG training (before ETL), we fixed \(|\sC|\) to \(1024\), and \(\lambda\) to \(0.5\).}
\label{table:compare_etl_noise_celeba_det}
\centering
\begin{tabular}[t]{lrrrrrr}
\toprule
Game & \multicolumn{1}{c}{\(\lambda\)} & \multicolumn{1}{c}{\(|\sC|\)} & \multicolumn{4}{c}{ETL tasks} \\[1ex]\cmidrule(r){4-7}
 & & & \multicolumn{1}{c}{Discrimination} & \multicolumn{1}{c}{Classification} & \multicolumn{1}{c}{Attribute} & \multicolumn{1}{c}{Reconstruction}\\
\midrule
LG {\scriptsize(RL)} & \(0\) & \(1024\) & \longcell{\(0.62\)\\{\tiny(\(0.08\))}} & \longcell{\(0.01\)\\{\tiny(\(0.00\))}} & \longcell{\(0.85\)\\{\tiny(\(0.00\))}} & \longcell{\(5671\)\\{\tiny(\(192\))}} \\[2.2ex]
NLG & \(0.5\) & \(1024\) & \longcell{\(0.60\)\\{\tiny(\(0.10\))}} & \longcell{\(0.00\)\\{\tiny(\(0.00\))}} & \longcell{\(0.85\)\\{\tiny(\(0.00\))}} & \longcell{\(5659\)\\{\tiny(\(129\))}} \\[2.2ex]
\bottomrule
\end{tabular}
\end{table*}

\begin{table*}[t]
\linespread{0.6}\selectfont\centering
\caption{Test accuracy, with SD, of every ETL task trained, using the ImageNet dataset and over 10 seeds. We fix the noise during ETL's train and test phase at \(0.5\). During the LG training (before ETL), we fixed \(|\sC|\) to \(1024\), and \(\lambda\) to \(0.5\).}
\label{table:compare_etl_noise_imagenet_noise}
\centering
\begin{tabular}[t]{lrrrr}
\toprule
Game & \(\lambda\) & \multicolumn{1}{c}{\(|\sC|\)} & \multicolumn{2}{c}{ETL tasks} \\[1ex]\cmidrule(r){4-5}
 & & & \multicolumn{1}{c}{Discrimination} & \multicolumn{1}{c}{Classification} \\
\midrule
LG {\scriptsize(RL)} & \(0\) & \(1024\) & \longcell{\(0.36\)\\{\tiny(\(0.02\))}} & \longcell{\(0.07\)\\{\tiny(\(0.00\))}} \\[2.2ex]
NLG & \(0.5\) & \(1024\) & \longcell{\(0.41\)\\{\tiny(\(0.02\))}} & \longcell{\(0.08\)\\{\tiny(\(0.01\))}} \\[2.2ex]
\bottomrule
\end{tabular}
\end{table*}

\begin{table*}[t]
\linespread{0.6}\selectfont\centering
\caption{Test accuracy, with SD, of every ETL task trained, using the CelebA dataset and over 10 seeds. We fix the noise during ETL's train and test phase at \(0.5\). During the LG training (before ETL), we fixed \(|\sC|\) to \(1024\), and \(\lambda\) to \(0.5\).}
\label{table:compare_etl_noise_celeba_noise}
\centering
\begin{tabular}[t]{lrrrrrr}
\toprule
Game & \multicolumn{1}{c}{\(\lambda\)} & \multicolumn{1}{c}{\(|\sC|\)} & \multicolumn{4}{c}{ETL tasks} \\[1ex]\cmidrule(r){4-7}
 & & & \multicolumn{1}{c}{Discrimination} & \multicolumn{1}{c}{Classification} & \multicolumn{1}{c}{Attribute} & \multicolumn{1}{c}{Reconstruction}\\
\midrule
LG {\scriptsize(RL)} & \(0\) & \(1024\) & \longcell{\(0.26\)\\{\tiny(\(0.03\))}} & \longcell{\(0.00\)\\{\tiny(\(0.00\))}} & \longcell{\(0.85\)\\{\tiny(\(0.00\))}} & \longcell{\(5871\)\\{\tiny(\(155\))}} \\[2.2ex]
NLG & \(0.5\) & 1024 & \longcell{\(0.31\)\\{\tiny(\(0.04\))}} & \longcell{\(0.00\)\\{\tiny(\(0.00\))}} & \longcell{\(0.85\)\\{\tiny(\(0.00\))}} & \longcell{\(5833\)\\{\tiny(\(103\))}} \\[2.2ex]
\bottomrule
\end{tabular}
\end{table*}

\subsubsection{Varying Number of Candidates}
Finally, we study the last axis of the predominant hyper-parameters, which is the number of candidates. We do a quick analysis since the results comply with the main evaluation (\Cref{sec:eval-cand}), where increasing the number of candidates is essential to improve performance. By increasing the number of candidates, the Listener receives a broader variety of samples, facilitating learning and recalling unique and valuable information to discriminate all candidates. Consequently, a more sophisticated communication protocol is necessary as the number of candidates increases since the discriminative task becomes more difficult. For this evaluation and similar to the previous ones, we fix the noise at \(0.5\). The results are for the following game variants: LG (RL), and NLG.

\Cref{table:compare_etl_cand_imagenet,table:compare_etl_cand_imagenet_noise} display the results for the ImageNet dataset employing the deterministic and noisy regimes, respectively. Looking at the Discrimination task, we observe an increase in test accuracy for all tasks as the number of candidates (used in the original RL task) scales up. Another important and related observation is that the gap between accuracies decreases in the noisy train/test regime. For example, in the regular regime (deterministic), the gap between 256 and 1024 candidates for LG (RL) and NLG variants is about \(0.08\) for both. On the other hand, the gaps for the same variants in the noisy regime are \(0.01\) for both noiseless and noisy regimes.

A similar outcome occurs for the results obtained when training with the CelebA dataset, \Cref{table:compare_etl_cand_celeba,table:compare_etl_cand_celeba_noise} display the results for the deterministic and noisy regimes, respectively. In the case of the Discrimination and Reconstruction tasks applied in both train/test regimes, there is a clear benefit of having a Speaker trained in a more complex RL task (increased number of candidates). As such, we observe that the test accuracy increases for the Discrimination task as the number of candidates increases for all LG variants. Comparetively, the loss obtained at evaluation time for the Reconstruction task also decreases as the number of candidates increases. The performance gap is also more visible in the deterministic regime, where no message noise is involved. As an illustration, consider the Discrimination task where the growth in accuracy for the NLG variant, when increasing the number of candidates from 256 to 1024, is \(0.17\) and \(0.11\), referring to the deterministic and noise regimes, respectively. Comparatively, looking at the Reconstruction task and the same LG variant, the increment in the test loss from 256 to 1024 candidates is \(128\) and \(135\) for the deterministic and noisy train/test regimes, respectively. Finally, addressing the Attribute task, we observe no significant improvements in performance when expanding the number of candidates. Similarly to the previous evaluation analyzes (\Cref{app:etl-eval-noise}), the test accuracy stays fixed at \(0.85\)/\(0.86\).

\begin{table*}[t]
\linespread{0.6}\selectfont\centering
\caption{Test accuracy, with SD, of every ETL task trained, using the ImageNet dataset and over 10 seeds. We do not apply noise during ETL's train and test phase. During the LG training (before ETL), we fixed \(\lambda\) to \(0.5\).}
\label{table:compare_etl_cand_imagenet}
\centering
\begin{tabular}[t]{lrrrr}
\toprule
Game & \multicolumn{1}{c}{\(\lambda\)} & \multicolumn{1}{c}{\(|\sC|\)} & \multicolumn{2}{c}{ETL tasks} \\[1ex]\cmidrule(r){4-5}
 & & & \multicolumn{1}{c}{Discrimination} & \multicolumn{1}{c}{Classification} \\
\midrule
LG {\scriptsize(RL)} & \(0\) & \(16\) & \longcell{\(0.37\)\\{\tiny(\(0.06\))}} & \longcell{\(0.07\)\\{\tiny(\(0.01\))}} \\[2.2ex]
LG {\scriptsize(RL)} & \(0\) & \(64\) & \longcell{\(0.64\)\\{\tiny(\(0.09\))}} & \longcell{\(0.07\)\\{\tiny(\(0.01\))}} \\[2.2ex]
LG {\scriptsize(RL)} & \(0\) & \(256\) & \longcell{\(0.80\)\\{\tiny(\(0.04\))}} & \longcell{\(0.13\)\\{\tiny(\(0.01\))}} \\[2.2ex]
LG {\scriptsize(RL)} & \(0\) & \(1024\) & \longcell{\(0.88\)\\{\tiny(\(0.01\))}} & \longcell{\(0.14\)\\{\tiny(\(0.01\))}} \\[2.2ex]
NLG & \(0.5\) & \(16\) & \longcell{\(0.30\)\\{\tiny(\(0.05\))}} & \longcell{\(0.07\)\\{\tiny(\(0.01\))}} \\[2.2ex]
NLG & \(0.5\) & \(64\) & \longcell{\(0.54\)\\{\tiny(\(0.05\))}} & \longcell{\(0.07\)\\{\tiny(\(0.01\))}} \\[2.2ex]
NLG & \(0.5\) & \(256\) & \longcell{\(0.77\)\\{\tiny(\(0.04\))}} & \longcell{\(0.13\)\\{\tiny(\(0.01\))}} \\[2.2ex]
NLG & \(0.5\) & \(1024\) & \longcell{\(0.85\)\\{\tiny(\(0.03\))}} & \longcell{\(0.14\)\\{\tiny(\(0.02\))}} \\[2.2ex]
\bottomrule
\end{tabular}
\end{table*}

\begin{table*}[t]
\linespread{0.6}\selectfont\centering
\caption{Test accuracy, with SD, of every ETL task trained, using the CelebA dataset and over 10 seeds. We do not apply noise during ETL's train and test phase. During the LG training (before ETL), we fixed \(\lambda\) to \(0.5\).}
\label{table:compare_etl_cand_celeba}
\centering
\begin{tabular}[t]{lrrrrrrr}
\toprule
Game & \(\lambda\) & \(\nu\) & \multicolumn{1}{c}{\(|\sC|\)} & \multicolumn{4}{c}{ETL tasks} \\[1ex]\cmidrule(r){5-8}
 & & & & \multicolumn{1}{c}{Discrimination} & \multicolumn{1}{c}{Classification} & \multicolumn{1}{c}{Attribute} & \multicolumn{1}{c}{Reconstruction}\\
\midrule
LG {\scriptsize(RL)}  & \multicolumn{1}{c}{-}  & \multicolumn{1}{c}{-} & \(16\) & \longcell{\(0.17\)\\{\tiny(\(0.07\))}} & \longcell{\(0.00\)\\{\tiny(\(0.00\))}} & \longcell{\(0.85\)\\{\tiny(\(0.00\))}} & \longcell{\(6050\)\\{\tiny(\(50\))}} \\[2.2ex]
LG {\scriptsize(RL)}  & \multicolumn{1}{c}{-}  & \multicolumn{1}{c}{-} & \(64\) & \longcell{\(0.34\)\\{\tiny(\(0.11\))}} & \longcell{\(0.00\)\\{\tiny(\(0.00\))}} & \longcell{\(0.85\)\\{\tiny(\(0.00\))}} & \longcell{\(6012\)\\{\tiny(\(121\))}} \\[2.2ex]
LG {\scriptsize(RL)}  & \multicolumn{1}{c}{-}  & \multicolumn{1}{c}{-} & \(256\) & \longcell{\(0.48\)\\{\tiny(\(0.09\))}} & \longcell{\(0.00\)\\{\tiny(\(0.00\))}} & \longcell{\(0.86\)\\{\tiny(\(0.00\))}} & \longcell{\(5750\)\\{\tiny(\(1544\))}} \\[2.2ex]
LG {\scriptsize(RL)}  & \multicolumn{1}{c}{-}  & \multicolumn{1}{c}{-} & \(1024\) & \longcell{\(0.65\)\\{\tiny(\(0.09\))}} & \longcell{\(0.01\)\\{\tiny(\(0.00\))}} & \longcell{\(0.86\)\\{\tiny(\(0.00\))}} & \longcell{\(5633\)\\{\tiny(\(127\))}} \\[2.2ex]
NLG & \(0.5\) & \multicolumn{1}{c}{-} & \(16\) & \longcell{\(0.16\)\\{\tiny(\(0.06\))}} & \longcell{\(0.00\)\\{\tiny(\(0.00\))}} & \longcell{\(0.85\)\\{\tiny(\(0.00\))}} & \longcell{\(6071\)\\{\tiny(\(80\))}} \\[2.2ex]
NLG & \(0.5\) & \multicolumn{1}{c}{-} & \(64\) & \longcell{\(0.24\)\\{\tiny(\(0.08\))}} & \longcell{\(0.00\)\\{\tiny(\(0.00\))}} & \longcell{\(0.85\)\\{\tiny(\(0.00\))}} & \longcell{\(5992\)\\{\tiny(\(164\))}} \\[2.2ex]
NLG & \(0.5\) & \multicolumn{1}{c}{-} & \(256\) & \longcell{\(0.37\)\\{\tiny(\(0.10\))}} & \longcell{\(0.00\)\\{\tiny(\(0.00\))}} & \longcell{\(0.85\)\\{\tiny(\(0.00\))}} & \longcell{\(5815\)\\{\tiny(\(192\))}} \\[2.2ex]
NLG & \(0.5\) & \multicolumn{1}{c}{-} & \(1024\) & \longcell{\(0.54\)\\{\tiny(\(0.01\))}} & \longcell{\(0.00\)\\{\tiny(\(0.00\))}} & \longcell{\(0.85\)\\{\tiny(\(0.00\))}} & \longcell{\(5687\)\\{\tiny(\(157\))}} \\[2.2ex]
\bottomrule
\end{tabular}
\end{table*}

\begin{table*}[t]
\linespread{0.6}\selectfont\centering
\caption{Test accuracy, with SD, of every ETL task trained, using the ImageNet dataset and over 10 seeds. We fix the noise during ETL's train and test phase at \(0.5\). During the LG training (before ETL), we fixed \(\lambda\) to \(0.5\).}
\label{table:compare_etl_cand_imagenet_noise}
\centering
\begin{tabular}[t]{lrrrrr}
\toprule
Game & \multicolumn{1}{c}{\(\lambda\)} & \multicolumn{1}{c}{\(\nu\)} & \multicolumn{1}{c}{\(|\sC|\)} & \multicolumn{2}{c}{ETL tasks} \\[1ex]\cmidrule(r){5-6}
 & & & & \multicolumn{1}{c}{Discrimination} & \multicolumn{1}{c}{Classification} \\
\midrule
LG {\scriptsize(RL)} & \multicolumn{1}{c}{-}  & \multicolumn{1}{c}{-} & \(16\) & \longcell{\(0.11\)\\{\tiny(\(0.01\))}} & \longcell{\(0.04\)\\{\tiny(\(0.00\))}} \\[2.2ex]
LG {\scriptsize(RL)} & \multicolumn{1}{c}{-}  & \multicolumn{1}{c}{-} & \(64\) & \longcell{\(0.22\)\\{\tiny(\(0.03\))}} & \longcell{\(0.06\)\\{\tiny(\(0.01\))}} \\[2.2ex]
LG {\scriptsize(RL)} & \multicolumn{1}{c}{-}  & \multicolumn{1}{c}{-} & \(256\) & \longcell{\(0.31\)\\{\tiny(\(0.02\))}} & \longcell{\(0.06\)\\{\tiny(\(0.00\))}} \\[2.2ex]
LG {\scriptsize(RL)} & \multicolumn{1}{c}{-}  & \multicolumn{1}{c}{-} & \(1024\) & \longcell{\(0.36\)\\{\tiny(\(0.02\))}} & \longcell{\(0.07\)\\{\tiny(\(0.00\))}} \\[2.2ex]
NLG & \(0.5\) & \multicolumn{1}{c}{-} & \(16\) & \longcell{\(0.13\)\\{\tiny(\(0.02\))}} & \longcell{\(0.04\)\\{\tiny(\(0.00\))}} \\[2.2ex]
NLG & \(0.5\) & \multicolumn{1}{c}{-} & \(64\) & \longcell{\(0.23\)\\{\tiny(\(0.02\))}} & \longcell{\(0.04\)\\{\tiny(\(0.00\))}} \\[2.2ex]
NLG & \(0.5\) & \multicolumn{1}{c}{-} & \(256\) & \longcell{\(0.35\)\\{\tiny(\(0.02\))}} & \longcell{\(0.08\)\\{\tiny(\(0.00\))}} \\[2.2ex]
NLG & \(0.5\) & \multicolumn{1}{c}{-} & \(1024\) & \longcell{\(0.41\)\\{\tiny(\(0.02\))}} & \longcell{\(0.08\)\\{\tiny(\(0.01\))}} \\[2.2ex]
\bottomrule
\end{tabular}
\end{table*}

\begin{table*}[t]
\linespread{0.6}\selectfont\centering
\caption{Test accuracy, with SD, of every ETL task trained, using the CelebA dataset and over 10 seeds. We fix the noise during ETL's train and test phase at \(0.5\). During the LG training (before ETL), we fixed \(\lambda\) to \(0.5\).}
\label{table:compare_etl_cand_celeba_noise}
\centering
\begin{tabular}[t]{lrrrrrrr}
\toprule
Game & \(\lambda\) & \(\nu\) & \multicolumn{1}{c}{\(|\sC|\)} & \multicolumn{4}{c}{ETL tasks} \\[1ex]\cmidrule(r){5-8}
 & & & & \multicolumn{1}{c}{Discrimination} & \multicolumn{1}{c}{Classification} & \multicolumn{1}{c}{Attribute} & \multicolumn{1}{c}{Reconstruction}\\
\midrule
LG {\scriptsize(RL)} & \(0\) & \multicolumn{1}{c}{-} & \(16\) & \longcell{\(0.06\)\\{\tiny(\(0.02\))}} & \longcell{\(0.00\)\\{\tiny(\(0.00\))}} & \longcell{\(0.85\)\\{\tiny(\(0.00\))}} & \longcell{\(6210\)\\{\tiny(\(53\))}} \\[2.2ex]
LG {\scriptsize(RL)} & \(0\) & \multicolumn{1}{c}{-} & \(64\) & \longcell{\(0.13\)\\{\tiny(\(0.03\))}} & \longcell{\(0.00\)\\{\tiny(\(0.00\))}} & \longcell{\(0.85\)\\{\tiny(\(0.00\))}} & \longcell{\(6061\)\\{\tiny(\(140\))}} \\[2.2ex]
LG {\scriptsize(RL)} & \(0\) & \multicolumn{1}{c}{-} & \(256\) & \longcell{\(0.19\)\\{\tiny(\(0.03\))}} & \longcell{\(0.00\)\\{\tiny(\(0.00\))}} & \longcell{\(0.85\)\\{\tiny(\(0.00\))}} & \longcell{\(5896\)\\{\tiny(\(117\))}} \\[2.2ex]
LG {\scriptsize(RL)} & \(0\) & \multicolumn{1}{c}{-} & \(1024\) & \longcell{\(0.26\)\\{\tiny(\(0.03\))}} & \longcell{\(0.00\)\\{\tiny(\(0.00\))}} & \longcell{\(0.85\)\\{\tiny(\(0.00\))}} & \longcell{\(5871\)\\{\tiny(\(155\))}} \\[2.2ex]
NLG & \(0.5\) & \multicolumn{1}{c}{-} & \(16\) & \longcell{\(0.06\)\\{\tiny(\(0.02\))}} & \longcell{\(0.00\)\\{\tiny(\(0.00\))}} & \longcell{\(0.85\)\\{\tiny(\(0.00\))}} & \longcell{\(6196\)\\{\tiny(\(73\))}} \\[2.2ex]
NLG & \(0.5\) & \multicolumn{1}{c}{-} & \(64\) & \longcell{\(0.11\)\\{\tiny(\(0.04\))}} & \longcell{\(0.00\)\\{\tiny(\(0.00\))}} & \longcell{\(0.85\)\\{\tiny(\(0.00\))}} & \longcell{\(6096\)\\{\tiny(\(110\))}} \\[2.2ex]
NLG & \(0.5\) & \multicolumn{1}{c}{-} & \(256\) & \longcell{\(0.20\)\\{\tiny(\(0.04\))}} & \longcell{\(0.00\)\\{\tiny(\(0.00\))}} & \longcell{\(0.85\)\\{\tiny(\(0.00\))}} & \longcell{\(5968\)\\{\tiny(\(160\))}} \\[2.2ex]
NLG & \(0.5\) & \multicolumn{1}{c}{-} & \(1024\) & \longcell{\(0.31\)\\{\tiny(\(0.04\))}} & \longcell{\(0.00\)\\{\tiny(\(0.00\))}} & \longcell{\(0.85\)\\{\tiny(\(0.00\))}} & \longcell{\(5833\)\\{\tiny(\(103\))}} \\[2.2ex]
\bottomrule
\end{tabular}
\end{table*}

\subsubsection{Particular Analysis of the Classification Task} \label{app:etl-classif-analysis}
To conclude the ETL study, we now focus on the Classification task for both datasets (ImageNet and CelebA). In all experiments, the test accuracy is not higher than \(0.15\) and \(0.01\) when considering the ImageNet and CelebA datasets, respectively. We argue that this outcome is a consequence of the encoding qualities of the communication protocol. With these results, there is strong evidence that the emerged communication protocol does not encode any particular information about the class of each image since such information is irrelevant to solve the original setting (discriminating between images). Therefore, since the Listener architecture for the Classification task only receives the message sent by the Speaker as input, and it has no helpful information to solve the task.

\begin{table*}[t]
\linespread{0.6}\selectfont\centering
\centering
\caption{Test accuracy, with SD, of every ETL task trained, using the ImageNet dataset and over 10 seeds. We do not apply noise during ETL's train and test phase.}
\label{table:etl_imagenet_0_eval}
\begin{tabular}{lrrrrrr}
\toprule
Game & \(\lambda\) & \(|\mathcal{\sC}|\) & \multicolumn{1}{c}{Discrimination} & \multicolumn{1}{c}{Classification} \\[1ex]
\midrule
LG {\scriptsize(S)} & \(0\) & \(16\) & \longcell{\(0.23\)\\{\tiny(\(0.02\))}} & \longcell{\(0.08\)\\{\tiny(\(0.01\))}} \\[2.2ex]
LG {\scriptsize(S)} & \(0\) & \(64\) & \longcell{\(0.33\)\\{\tiny(\(0.06\))}} & \longcell{\(0.11\)\\{\tiny(\(0.01\))}} \\[2.2ex]
LG {\scriptsize(S)} & \(0\) & \(256\) & \longcell{\(0.44\)\\{\tiny(\(0.05\))}} & \longcell{\(0.13\)\\{\tiny(\(0.01\))}} \\[2.2ex]
LG {\scriptsize(S)} & \(0\) & \(1024\) & \longcell{\(0.58\)\\{\tiny(\(0.05\))}} & \longcell{\(0.14\)\\{\tiny(\(0.01\))}} \\[2.2ex]
LG {\scriptsize(RL)} & \(0\) & \(16\) & \longcell{\(0.37\)\\{\tiny(\(0.06\))}} & \longcell{\(0.07\)\\{\tiny(\(0.01\))}} \\[2.2ex]
LG {\scriptsize(RL)} & \(0\) & \(64\) & \longcell{\(0.64\)\\{\tiny(\(0.09\))}} & \longcell{\(0.11\)\\{\tiny(\(0.02\))}} \\[2.2ex]
LG {\scriptsize(RL)} & \(0\) & \(256\) & \longcell{\(0.80\)\\{\tiny(\(0.04\))}} & \longcell{\(0.13\)\\{\tiny(\(0.01\))}} \\
\bottomrule
\end{tabular}
\end{table*}

\begin{table*}[t]
\linespread{0.6}\selectfont\centering
\centering
\caption{Test accuracy, with SD, of every ETL task trained, using the CelebA dataset and over 10 seeds. We do not apply noise during ETL's train and test phase.}
\label{table:etl_celeba_0_eval}
\begin{tabular}{lrrrrrr}
\toprule
Game & \(\lambda\) & \(|\mathcal{\sC}|\) & \multicolumn{1}{c}{Discrimination} & \multicolumn{1}{c}{Classification} & \multicolumn{1}{c}{Attribute} & \multicolumn{1}{c}{Reconstruction} \\[1ex]
\midrule
LG {\scriptsize(S)} & \(0\) & \(16\) & \longcell{\(0.18\)\\{\tiny(\(0.04\))}} & \longcell{\(0.00\)\\{\tiny(\(0.00\))}} & \longcell{\(0.85\)\\{\tiny(\(0.00\))}} & \longcell{\(6094\)\\{\tiny(\(84\))}} \\[2.2ex]
LG {\scriptsize(S)} & \(0\) & \(64\) & \longcell{\(0.31\)\\{\tiny(\(0.02\))}} & \longcell{\(0.00\)\\{\tiny(\(0.00\))}} & \longcell{\(0.86\)\\{\tiny(\(0.00\))}} & \longcell{\(5971\)\\{\tiny(\(78\))}} \\[2.2ex]
LG {\scriptsize(S)} & \(0\) & \(256\) & \longcell{\(0.39\)\\{\tiny(\(0.06\))}} & \longcell{\(0.01\)\\{\tiny(\(0.00\))}} & \longcell{\(0.86\)\\{\tiny(\(0.00\))}} & \longcell{\(5852\)\\{\tiny(\(122\))}} \\[2.2ex]
LG {\scriptsize(S)} & \(0\) & \(1024\) & \longcell{\(0.44\)\\{\tiny(\(0.05\))}} & \longcell{\(0.01\)\\{\tiny(\(0.00\))}} & \longcell{\(0.86\)\\{\tiny(\(0.00\))}} & \longcell{\(5814\)\\{\tiny(\(74\))}} \\[2.2ex]
LG {\scriptsize(RL)} & \(0\) & \(16\) & \longcell{\(0.15\)\\{\tiny(\(0.06\))}} & \longcell{\(0.00\)\\{\tiny(\(0.00\))}} & \longcell{\(0.85\)\\{\tiny(\(0.00\))}} & \longcell{\(6110\)\\{\tiny(\(79\))}} \\[2.2ex]
LG {\scriptsize(RL)} & \(0\) & \(64\) & \longcell{\(0.33\)\\{\tiny(\(0.08\))}} & \longcell{\(0.00\)\\{\tiny(\(0.00\))}} & \longcell{\(0.85\)\\{\tiny(\(0.00\))}} & \longcell{\(5930\)\\{\tiny(\(178\))}} \\[2.2ex]
LG {\scriptsize(RL)} & \(0\) & \(256\) & \longcell{\(0.47\)\\{\tiny(\(0.11\))}} & \longcell{\(0.00\)\\{\tiny(\(0.00\))}} & \longcell{\(0.86\)\\{\tiny(\(0.00\))}} & \longcell{\(5710\)\\{\tiny(\(167\))}} \\
\bottomrule
\end{tabular}
\end{table*}

\begin{table*}[t]
\linespread{0.6}\selectfont\centering
\centering
\caption{Test accuracy, with SD, of every ETL task trained, using the ImageNet dataset and over 10 seeds. We fix the noise during ETL's train and test phase at \(0.25\).}
\label{table:etl_imagenet_025_eval}
\begin{tabular}{lrrrrrr}
\toprule
Game & \(\lambda\) & \(|\mathcal{\sC}|\) & \multicolumn{1}{c}{Discrimination} & \multicolumn{1}{c}{Classification} \\[1ex]
\midrule
LG {\scriptsize(S)} & \(0\) & \(16\) & \longcell{\(0.13\)\\{\tiny(\(0.02\))}} & \longcell{\(0.06\)\\{\tiny(\(0.01\))}} \\[2.2ex]
LG {\scriptsize(S)} & \(0\) & \(64\) & \longcell{\(0.19\)\\{\tiny(\(0.04\))}} & \longcell{\(0.08\)\\{\tiny(\(0.01\))}} \\[2.2ex]
LG {\scriptsize(S)} & \(0\) & \(256\) & \longcell{\(0.24\)\\{\tiny(\(0.04\))}} & \longcell{\(0.09\)\\{\tiny(\(0.01\))}} \\[2.2ex]
LG {\scriptsize(S)} & \(0\) & \(1024\) & \longcell{\(0.36\)\\{\tiny(\(0.05\))}} & \longcell{\(0.10\)\\{\tiny(\(0.01\))}} \\[2.2ex]
LG {\scriptsize(RL)} & \(0\) & \(16\) & \longcell{\(0.23\)\\{\tiny(\(0.03\))}} & \longcell{\(0.06\)\\{\tiny(\(0.01\))}} \\[2.2ex]
LG {\scriptsize(RL)} & \(0\) & \(64\) & \longcell{\(0.43\)\\{\tiny(\(0.05\))}} & \longcell{\(0.08\)\\{\tiny(\(0.01\))}} \\[2.2ex]
LG {\scriptsize(RL)} & \(0\) & \(256\) & \longcell{\(0.59\)\\{\tiny(\(0.03\))}} & \longcell{\(0.10\)\\{\tiny(\(0.01\))}} \\
\bottomrule
\end{tabular}
\end{table*}

\begin{table*}[t]
\linespread{0.6}\selectfont\centering
\centering
\caption{Test accuracy, with SD, of every ETL task trained, using the CelebA dataset and over 10 seeds. We fix the noise during ETL's train and test phase at \(0.25\).}
\label{table:etl_celeba_025_eval}
\begin{tabular}{lrrrrrr}
\toprule
Game & \(\lambda\) & \(|\mathcal{\sC}|\) & \multicolumn{1}{c}{Discrimination} & \multicolumn{1}{c}{Classification} & \multicolumn{1}{c}{Attribute} & \multicolumn{1}{c}{Reconstruction} \\[1ex]
\midrule
LG {\scriptsize(S)} & \(0\) & \(16\) & \longcell{\(0.11\)\\{\tiny(\(0.02\))}} & \longcell{\(0.00\)\\{\tiny(\(0.00\))}} & \longcell{\(0.85\)\\{\tiny(\(0.00\))}} & \longcell{\(6177\)\\{\tiny(\(67\))}} \\[2.2ex]
LG {\scriptsize(S)} & \(0\) & \(64\) & \longcell{\(0.19\)\\{\tiny(\(0.02\))}} & \longcell{\(0.00\)\\{\tiny(\(0.00\))}} & \longcell{\(0.85\)\\{\tiny(\(0.00\))}} & \longcell{\(6055\)\\{\tiny(\(60\))}} \\[2.2ex]
LG {\scriptsize(S)} & \(0\) & \(256\) & \longcell{\(0.25\)\\{\tiny(\(0.05\))}} & \longcell{\(0.00\)\\{\tiny(\(0.00\))}} & \longcell{\(0.85\)\\{\tiny(\(0.00\))}} & \longcell{\(5955\)\\{\tiny(\(116\))}} \\[2.2ex]
LG {\scriptsize(S)} & \(0\) & \(1024\) & \longcell{\(0.28\)\\{\tiny(\(0.05\))}} & \longcell{\(0.00\)\\{\tiny(\(0.00\))}} & \longcell{\(0.85\)\\{\tiny(\(0.00\))}} & \longcell{\(5932\)\\{\tiny(\(67\))}} \\[2.2ex]
LG {\scriptsize(RL)} & \(0\) & \(16\) & \longcell{\(0.11\)\\{\tiny(\(0.04\))}} & \longcell{\(0.00\)\\{\tiny(\(0.00\))}} & \longcell{\(0.85\)\\{\tiny(\(0.00\))}} & \longcell{\(6145\)\\{\tiny(\(65\))}} \\[2.2ex]
LG {\scriptsize(RL)} & \(0\) & \(64\) & \longcell{\(0.24\)\\{\tiny(\(0.05\))}} & \longcell{\(0.00\)\\{\tiny(\(0.00\))}} & \longcell{\(0.85\)\\{\tiny(\(0.00\))}} & \longcell{\(5973\)\\{\tiny(\(162\))}} \\[2.2ex]
LG {\scriptsize(RL)} & \(0\) & \(256\) & \longcell{\(0.34\)\\{\tiny(\(0.07\))}} & \longcell{\(0.00\)\\{\tiny(\(0.00\))}} & \longcell{\(0.85\)\\{\tiny(\(0.00\))}} & \longcell{\(5769\)\\{\tiny(\(144\))}} \\
\bottomrule
\end{tabular}
\end{table*}

\begin{table*}[t]
\linespread{0.6}\selectfont\centering
\centering
\caption{Test accuracy, with SD, of every ETL task trained, using the ImageNet dataset and over 10 seeds. We fix the noise during ETL's train and test phase at \(0.5\).}
\label{table:etl_imagenet_05_eval}
\begin{tabular}{lrrrrrr}
\toprule
Game & \(\lambda\) & \(|\mathcal{\sC}|\) & \multicolumn{1}{c}{Discrimination} & \multicolumn{1}{c}{Classification} \\[1ex]
\midrule
LG {\scriptsize(S)} & \(0\) & \(16\) & \longcell{\(0.06\)\\{\tiny(\(0.01\))}} & \longcell{\(0.04\)\\{\tiny(\(0.00\))}} \\[2.2ex]
LG {\scriptsize(S)} & \(0\) & \(64\) & \longcell{\(0.09\)\\{\tiny(\(0.02\))}} & \longcell{\(0.05\)\\{\tiny(\(0.01\))}} \\[2.2ex]
LG {\scriptsize(S)} & \(0\) & \(256\) & \longcell{\(0.10\)\\{\tiny(\(0.03\))}} & \longcell{\(0.05\)\\{\tiny(\(0.01\))}} \\[2.2ex]
LG {\scriptsize(S)} & \(0\) & \(1024\) & \longcell{\(0.16\)\\{\tiny(\(0.03\))}} & \longcell{\(0.06\)\\{\tiny(\(0.01\))}} \\[2.2ex]
LG {\scriptsize(RL)} & \(0\) & \(16\) & \longcell{\(0.11\)\\{\tiny(\(0.01\))}} & \longcell{\(0.04\)\\{\tiny(\(0.00\))}} \\[2.2ex]
LG {\scriptsize(RL)} & \(0\) & \(64\) & \longcell{\(0.22\)\\{\tiny(\(0.03\))}} & \longcell{\(0.06\)\\{\tiny(\(0.01\))}} \\[2.2ex]
LG {\scriptsize(RL)} & \(0\) & \(256\) & \longcell{\(0.31\)\\{\tiny(\(0.02\))}} & \longcell{\(0.06\)\\{\tiny(\(0.00\))}} \\
\bottomrule
\end{tabular}
\end{table*}

\begin{table*}[t]
\linespread{0.6}\selectfont\centering
\centering
\caption{Test accuracy, with SD, of every ETL task trained, using the CelebA dataset and over 10 seeds. We fix the noise during ETL's train and test phase at \(0.5\).}
\label{table:etl_celeba_05_eval}
\begin{tabular}{lrrrrrr}
\toprule
Game & \(\lambda\) & \(|\mathcal{\sC}|\) & \multicolumn{1}{c}{Discrimination} & \multicolumn{1}{c}{Classification} & \multicolumn{1}{c}{Attribute} & \multicolumn{1}{c}{Reconstruction} \\[1ex]
\midrule
LG {\scriptsize(S)} & \(0\) & \(16\) & \longcell{\(0.05\)\\{\tiny(\(0.01\))}} & \longcell{\(0.00\)\\{\tiny(\(0.00\))}} & \longcell{\(0.84\)\\{\tiny(\(0.00\))}} & \longcell{\(6291\)\\{\tiny(\(53\))}} \\[2.2ex]
LG {\scriptsize(S)} & \(0\) & \(64\) & \longcell{\(0.09\)\\{\tiny(\(0.01\))}} & \longcell{\(0.00\)\\{\tiny(\(0.00\))}} & \longcell{\(0.85\)\\{\tiny(\(0.00\))}} & \longcell{\(6184\)\\{\tiny(\(48\))}} \\[2.2ex]
LG {\scriptsize(S)} & \(0\) & \(256\) & \longcell{\(0.12\)\\{\tiny(\(0.03\))}} & \longcell{\(0.00\)\\{\tiny(\(0.00\))}} & \longcell{\(0.85\)\\{\tiny(\(0.00\))}} & \longcell{\(6107\)\\{\tiny(\(104\))}} \\[2.2ex]
LG {\scriptsize(S)} & \(0\) & \(1024\) & \longcell{\(0.13\)\\{\tiny(\(0.03\))}} & \longcell{\(0.00\)\\{\tiny(\(0.00\))}} & \longcell{\(0.84\)\\{\tiny(\(0.00\))}} & \longcell{\(6102\)\\{\tiny(\(73\))}} \\[2.2ex]
LG {\scriptsize(RL)} & \(0\) & \(16\) & \longcell{\(0.06\)\\{\tiny(\(0.02\))}} & \longcell{\(0.00\)\\{\tiny(\(0.00\))}} & \longcell{\(0.85\)\\{\tiny(\(0.00\))}} & \longcell{\(6210\)\\{\tiny(\(53\))}} \\[2.2ex]
LG {\scriptsize(RL)} & \(0\) & \(64\) & \longcell{\(0.13\)\\{\tiny(\(0.03\))}} & \longcell{\(0.00\)\\{\tiny(\(0.00\))}} & \longcell{\(0.85\)\\{\tiny(\(0.00\))}} & \longcell{\(6061\)\\{\tiny(\(140\))}} \\[2.2ex]
LG {\scriptsize(RL)} & \(0\) & \(256\) & \longcell{\(0.19\)\\{\tiny(\(0.03\))}} & \longcell{\(0.00\)\\{\tiny(\(0.00\))}} & \longcell{\(0.85\)\\{\tiny(\(0.00\))}} & \longcell{\(5896\)\\{\tiny(\(117\))}} \\
\bottomrule
\end{tabular}
\end{table*}

\begin{table*}[t]
\linespread{0.6}\selectfont\centering
\centering
\caption{Test accuracy, with SD, of every ETL task trained, using the ImageNet dataset and over 10 seeds. We fix the noise during ETL's train and test phase at \(0.75\).}
\label{table:etl_imagenet_075_eval}
\begin{tabular}{lrrrrrr}
\toprule
Game & \(\lambda\) & \(|\mathcal{\sC}|\) & \multicolumn{1}{c}{Discrimination} & \multicolumn{1}{c}{Classification} \\[1ex]
\midrule
LG {\scriptsize(S)} & \(0\) & \(16\) & \longcell{\(0.02\)\\{\tiny(\(0.00\))}} & \longcell{\(0.02\)\\{\tiny(\(0.00\))}} \\[2.2ex]
LG {\scriptsize(S)} & \(0\) & \(64\) & \longcell{\(0.02\)\\{\tiny(\(0.01\))}} & \longcell{\(0.02\)\\{\tiny(\(0.00\))}} \\[2.2ex]
LG {\scriptsize(S)} & \(0\) & \(256\) & \longcell{\(0.02\)\\{\tiny(\(0.01\))}} & \longcell{\(0.02\)\\{\tiny(\(0.00\))}} \\[2.2ex]
LG {\scriptsize(S)} & \(0\) & \(1024\) & \longcell{\(0.04\)\\{\tiny(\(0.01\))}} & \longcell{\(0.03\)\\{\tiny(\(0.00\))}} \\[2.2ex]
LG {\scriptsize(RL)} & \(0\) & \(16\) & \longcell{\(0.03\)\\{\tiny(\(0.00\))}} & \longcell{\(0.02\)\\{\tiny(\(0.00\))}} \\[2.2ex]
LG {\scriptsize(RL)} & \(0\) & \(64\) & \longcell{\(0.06\)\\{\tiny(\(0.01\))}} & \longcell{\(0.03\)\\{\tiny(\(0.00\))}} \\[2.2ex]
LG {\scriptsize(RL)} & \(0\) & \(256\) & \longcell{\(0.08\)\\{\tiny(\(0.01\))}} & \longcell{\(0.03\)\\{\tiny(\(0.00\))}} \\
\bottomrule
\end{tabular}
\end{table*}

\begin{table*}[t]
\linespread{0.6}\selectfont\centering
\centering
\caption{Test accuracy, with SD, of every ETL task trained, using the CelebA dataset and over 10 seeds. We fix the noise during ETL's train and test phase at \(0.75\).}
\label{table:etl_celeba_075_eval}
\begin{tabular}{lrrrrrr}
\toprule
Game & \(\lambda\) & \(|\mathcal{\sC}|\) & \multicolumn{1}{c}{Discrimination} & \multicolumn{1}{c}{Classification} & \multicolumn{1}{c}{Attribute} & \multicolumn{1}{c}{Reconstruction} \\[1ex]
\midrule
LG {\scriptsize(S)} & \(0\) & \(16\) & \longcell{\(0.01\)\\{\tiny(\(0.00\))}} & \longcell{\(0.00\)\\{\tiny(\(0.00\))}} & \longcell{\(0.83\)\\{\tiny(\(0.00\))}} & \longcell{\(6438\)\\{\tiny(\(31\))}} \\[2.2ex]
LG {\scriptsize(S)} & \(0\) & \(64\) & \longcell{\(0.02\)\\{\tiny(\(0.00\))}} & \longcell{\(0.00\)\\{\tiny(\(0.00\))}} & \longcell{\(0.83\)\\{\tiny(\(0.00\))}} & \longcell{\(6357\)\\{\tiny(\(33\))}} \\[2.2ex]
LG {\scriptsize(S)} & \(0\) & \(256\) & \longcell{\(0.03\)\\{\tiny(\(0.01\))}} & \longcell{\(0.00\)\\{\tiny(\(0.00\))}} & \longcell{\(0.83\)\\{\tiny(\(0.00\))}} & \longcell{\(6311\)\\{\tiny(\(68\))}} \\[2.2ex]
LG {\scriptsize(S)} & \(0\) & \(1024\) & \longcell{\(0.03\)\\{\tiny(\(0.01\))}} & \longcell{\(0.00\)\\{\tiny(\(0.00\))}} & \longcell{\(0.83\)\\{\tiny(\(0.00\))}} & \longcell{\(6316\)\\{\tiny(\(59\))}} \\[2.2ex]
LG {\scriptsize(RL)} & \(0\) & \(16\) & \longcell{\(0.02\)\\{\tiny(\(0.00\))}} & \longcell{\(0.00\)\\{\tiny(\(0.00\))}} & \longcell{\(0.84\)\\{\tiny(\(0.00\))}} & \longcell{\(6331\)\\{\tiny(\(31\))}} \\[2.2ex]
LG {\scriptsize(RL)} & \(0\) & \(64\) & \longcell{\(0.04\)\\{\tiny(\(0.01\))}} & \longcell{\(0.00\)\\{\tiny(\(0.00\))}} & \longcell{\(0.84\)\\{\tiny(\(0.00\))}} & \longcell{\(6232\)\\{\tiny(\(104\))}} \\[2.2ex]
LG {\scriptsize(RL)} & \(0\) & \(256\) & \longcell{\(0.05\)\\{\tiny(\(0.01\))}} & \longcell{\(0.00\)\\{\tiny(\(0.00\))}} & \longcell{\(0.84\)\\{\tiny(\(0.00\))}} & \longcell{\(6122\)\\{\tiny(\(75\))}} \\
\bottomrule
\end{tabular}
\end{table*}

\section{Additional Results}
Continuing the evaluation present in \Cref{sec:eval}, we present all tests performed for each LG variants with different hyperparameters, see \Cref{table:lg_imagenet_0_test_reward,table:lg_celeba_0_test_reward,table:lg_imagenet_025_test_reward,table:lg_celeba_025_test_reward,table:lg_imagenet_05_test_reward,table:lg_celeba_05_test_reward,table:lg_imagenet_075_test_reward,table:lg_celeba_075_test_reward}.

\begin{table*}[t]
\linespread{0.6}\selectfont\centering
\centering
\caption{Test accuracy with SD for different game variants, using ImageNet dataset and over 10 seeds. During test \(\lambda_{\text{test}}\) is set to \(0\).}
\label{table:lg_imagenet_0_test_reward}
\begin{tabular}{lrrrr}
\toprule
Game & \(\lambda\) & \(|\sC|\) & \multicolumn{2}{c}{\(|\sC|\) (test)} \\[1ex]
 &  &  & \multicolumn{1}{c}{\(1024\)} & \multicolumn{1}{c}{\(4096\)} \\
\midrule
LG {\scriptsize(S)} & \(0\) & \(16\) & \longcell{\(0.61\)\\{\tiny(\(0.04\))}} & \longcell{\(0.35\)\\{\tiny(\(0.04\))}} \\[2.2ex]
LG {\scriptsize(S)} & \(0\) & \(64\) & \longcell{\(0.76\)\\{\tiny(\(0.06\))}} & \longcell{\(0.52\)\\{\tiny(\(0.08\))}} \\[2.2ex]
LG {\scriptsize(S)} & \(0\) & \(256\) & \longcell{\(0.88\)\\{\tiny(\(0.03\))}} & \longcell{\(0.71\)\\{\tiny(\(0.06\))}} \\[2.2ex]
LG {\scriptsize(S)} & \(0\) & \(1024\) & \longcell{\(0.96\)\\{\tiny(\(0.02\))}} & \longcell{\(0.88\)\\{\tiny(\(0.04\))}} \\[2.2ex]
LG {\scriptsize(RL)} & \(0\) & \(16\) & \longcell{\(0.67\)\\{\tiny(\(0.04\))}} & \longcell{\(0.39\)\\{\tiny(\(0.04\))}} \\[2.2ex]
LG {\scriptsize(RL)} & \(0\) & \(64\) & \longcell{\(0.93\)\\{\tiny(\(0.01\))}} & \longcell{\(0.79\)\\{\tiny(\(0.03\))}} \\
\bottomrule
\end{tabular}
\end{table*}

\begin{table*}[t]
\linespread{0.6}\selectfont\centering
\centering
\caption{Test accuracy with SD for different game variants, using CelebA dataset and over 10 seeds. During test \(\lambda_{\text{test}}\) is set to \(0\).}
\label{table:lg_celeba_0_test_reward}
\begin{tabular}{lrrrr}
\toprule
Game & \(\lambda\) & \(|\sC|\) & \multicolumn{2}{c}{\(|\sC|\) (test)} \\[1ex]
 &  & & \multicolumn{1}{c}{\(1024\)} & \multicolumn{1}{c}{\(4096\)} \\
\midrule
LG {\scriptsize(S)} & \(0\) & \(16\) & \longcell{\(0.56\)\\{\tiny(\(0.07\))}} & \longcell{\(0.29\)\\{\tiny(\(0.06\))}} \\[2.2ex]
LG {\scriptsize(S)} & \(0\) & \(64\) & \longcell{\(0.76\)\\{\tiny(\(0.03\))}} & \longcell{\(0.51\)\\{\tiny(\(0.03\))}} \\[2.2ex]
LG {\scriptsize(S)} & \(0\) & \(256\) & \longcell{\(0.81\)\\{\tiny(\(0.03\))}} & \longcell{\(0.59\)\\{\tiny(\(0.05\))}} \\[2.2ex]
LG {\scriptsize(S)} & \(0\) & \(1024\) & \longcell{\(0.75\)\\{\tiny(\(0.01\))}} & \longcell{\(0.53\)\\{\tiny(\(0.02\))}} \\[2.2ex]
LG {\scriptsize(RL)} & \(0\) & \(16\) & \longcell{\(0.58\)\\{\tiny(\(0.07\))}} & \longcell{\(0.28\)\\{\tiny(\(0.07\))}} \\[2.2ex]
LG {\scriptsize(RL)} & \(0\) & \(64\) & \longcell{\(0.88\)\\{\tiny(\(0.02\))}} & \longcell{\(0.65\)\\{\tiny(\(0.04\))}} \\
\bottomrule
\end{tabular}
\end{table*}

\begin{table*}[t]
\linespread{0.6}\selectfont\centering
\centering
\caption{Test accuracy with SD for different game variants, using ImageNet dataset and over 10 seeds. During test \(\lambda_{\text{test}}\) is set to \(0.25\).}
\label{table:lg_imagenet_025_test_reward}
\begin{tabular}{lrrrr}
\toprule
Game & \(\lambda\) & \(|\sC|\) & \multicolumn{2}{c}{\(|\sC|\) (test)} \\[1ex]
 &  &  & \multicolumn{1}{c}{\(1024\)} & \multicolumn{1}{c}{\(4096\)} \\
\midrule
LG {\scriptsize(S)} & \(0\) & \(16\) & \longcell{\(0.17\)\\{\tiny(\(0.03\))}} & \longcell{\(0.08\)\\{\tiny(\(0.02\))}} \\[2.2ex]
LG {\scriptsize(S)} & \(0\) & \(64\) & \longcell{\(0.21\)\\{\tiny(\(0.03\))}} & \longcell{\(0.11\)\\{\tiny(\(0.02\))}} \\[2.2ex]
LG {\scriptsize(S)} & \(0\) & \(256\) & \longcell{\(0.27\)\\{\tiny(\(0.05\))}} & \longcell{\(0.17\)\\{\tiny(\(0.03\))}} \\[2.2ex]
LG {\scriptsize(S)} & \(0\) & \(1024\) & \longcell{\(0.30\)\\{\tiny(\(0.05\))}} & \longcell{\(0.20\)\\{\tiny(\(0.04\))}} \\[2.2ex]
LG {\scriptsize(RL)} & \(0\) & \(16\) & \longcell{\(0.20\)\\{\tiny(\(0.04\))}} & \longcell{\(0.10\)\\{\tiny(\(0.02\))}} \\[2.2ex]
LG {\scriptsize(RL)} & \(0\) & \(64\) & \longcell{\(0.38\)\\{\tiny(\(0.10\))}} & \longcell{\(0.24\)\\{\tiny(\(0.06\))}} \\
\bottomrule
\end{tabular}
\end{table*}

\begin{table*}[t]
\linespread{0.6}\selectfont\centering
\centering
\caption{Test accuracy with SD for different game variants, using CelebA dataset and over 10 seeds. During test \(\lambda_{\text{test}}\) is set to \(0.25\).}
\label{table:lg_celeba_025_test_reward}
\begin{tabular}{lrrrr}
\toprule
Game & \(\lambda\) & \(|\sC|\) & \multicolumn{2}{c}{\(|\sC|\) (test)} \\[1ex]
 &  &  & \multicolumn{1}{c}{\(1024\)} & \multicolumn{1}{c}{\(4096\)} \\
\midrule
LG {\scriptsize(S)} & \(0\) & \(16\) & \longcell{\(0.14\)\\{\tiny(\(0.02\))}} & \longcell{\(0.06\)\\{\tiny(\(0.01\))}} \\[2.2ex]
LG {\scriptsize(S)} & \(0\) & \(64\) & \longcell{\(0.21\)\\{\tiny(\(0.03\))}} & \longcell{\(0.11\)\\{\tiny(\(0.01\))}} \\[2.2ex]
LG {\scriptsize(S)} & \(0\) & \(256\) & \longcell{\(0.21\)\\{\tiny(\(0.04\))}} & \longcell{\(0.12\)\\{\tiny(\(0.02\))}} \\[2.2ex]
LG {\scriptsize(S)} & \(0\) & \(1024\) & \longcell{\(0.19\)\\{\tiny(\(0.03\))}} & \longcell{\(0.10\)\\{\tiny(\(0.02\))}} \\[2.2ex]
LG {\scriptsize(RL)} & \(0\) & \(16\) & \longcell{\(0.18\)\\{\tiny(\(0.06\))}} & \longcell{\(0.07\)\\{\tiny(\(0.03\))}} \\[2.2ex]
LG {\scriptsize(RL)} & \(0\) & \(64\) & \longcell{\(0.29\)\\{\tiny(\(0.05\))}} & \longcell{\(0.16\)\\{\tiny(\(0.03\))}} \\
\bottomrule
\end{tabular}
\end{table*}

\begin{table*}[t]
\linespread{0.6}\selectfont\centering
\centering
\caption{Test accuracy with SD for different game variants, using ImageNet dataset and over 10 seeds. During test \(\lambda_{\text{test}}\) is set to \(0.5\).}
\label{table:lg_imagenet_05_test_reward}
\begin{tabular}{lrrrr}
\toprule
Game & \(\lambda\) & \(|\sC|\) & \multicolumn{2}{c}{\(|\sC|\) (test)} \\[1ex]
 &  &  & \multicolumn{1}{c}{\(1024\)} & \multicolumn{1}{c}{\(4096\)} \\
\midrule
LG {\scriptsize(S)} & \(0\) & \(16\) & \longcell{\(0.03\)\\{\tiny(\(0.01\))}} & \longcell{\(0.01\)\\{\tiny(\(0.00\))}} \\[2.2ex]
LG {\scriptsize(S)} & \(0\) & \(64\) & \longcell{\(0.04\)\\{\tiny(\(0.01\))}} & \longcell{\(0.02\)\\{\tiny(\(0.01\))}} \\[2.2ex]
LG {\scriptsize(S)} & \(0\) & \(256\) & \longcell{\(0.06\)\\{\tiny(\(0.02\))}} & \longcell{\(0.03\)\\{\tiny(\(0.01\))}} \\[2.2ex]
LG {\scriptsize(S)} & \(0\) & \(1024\) & \longcell{\(0.05\)\\{\tiny(\(0.01\))}} & \longcell{\(0.03\)\\{\tiny(\(0.01\))}} \\[2.2ex]
LG {\scriptsize(RL)} & \(0\) & \(16\) & \longcell{\(0.03\)\\{\tiny(\(0.01\))}} & \longcell{\(0.01\)\\{\tiny(\(0.01\))}} \\[2.2ex]
LG {\scriptsize(RL)} & \(0\) & \(64\) & \longcell{\(0.09\)\\{\tiny(\(0.07\))}} & \longcell{\(0.05\)\\{\tiny(\(0.03\))}} \\
\bottomrule
\end{tabular}
\end{table*}

\begin{table*}[t]
\linespread{0.6}\selectfont\centering
\centering
\caption{Test accuracy with SD for different game variants, using CelebA dataset and over 10 seeds. During test \(\lambda_{\text{test}}\) is set to \(0.5\).}
\label{table:lg_celeba_05_test_reward}
\begin{tabular}{lrrrr}
\toprule
Game & \(\lambda\) & \(|\sC|\) & \multicolumn{2}{c}{\(|\sC|\) (test)} \\[1ex]
 &  &  & \multicolumn{1}{c}{\(1024\)} & \multicolumn{1}{c}{\(4096\)} \\
\midrule
LG {\scriptsize(S)} & \(0\) & \(16\) & \longcell{\(0.03\)\\{\tiny(\(0.01\))}} & \longcell{\(0.01\)\\{\tiny(\(0.00\))}} \\[2.2ex]
LG {\scriptsize(S)} & \(0\) & \(64\) & \longcell{\(0.04\)\\{\tiny(\(0.01\))}} & \longcell{\(0.02\)\\{\tiny(\(0.01\))}} \\[2.2ex]
LG {\scriptsize(S)} & \(0\) & \(256\) & \longcell{\(0.04\)\\{\tiny(\(0.01\))}} & \longcell{\(0.02\)\\{\tiny(\(0.01\))}} \\[2.2ex]
LG {\scriptsize(S)} & \(0\) & \(1024\) & \longcell{\(0.03\)\\{\tiny(\(0.01\))}} & \longcell{\(0.01\)\\{\tiny(\(0.01\))}} \\[2.2ex]
LG {\scriptsize(RL)} & \(0\) & \(16\) & \longcell{\(0.04\)\\{\tiny(\(0.02\))}} & \longcell{\(0.01\)\\{\tiny(\(0.01\))}} \\[2.2ex]
LG {\scriptsize(RL)} & \(0\) & \(64\) & \longcell{\(0.05\)\\{\tiny(\(0.02\))}} & \longcell{\(0.02\)\\{\tiny(\(0.01\))}} \\
\bottomrule
\end{tabular}
\end{table*}

\begin{table*}[t]
\linespread{0.6}\selectfont\centering
\centering
\caption{Test accuracy with SD for different game variants, using ImageNet dataset and over 10 seeds. During test \(\lambda_{\text{test}}\) is set to \(0.75\).}
\label{table:lg_imagenet_075_test_reward}
\begin{tabular}{lrrrr}
\toprule
Game & \(\lambda\) & \(|\sC|\) & \multicolumn{2}{c}{\(|\sC|\) (test)} \\[1ex]
 &  &  & \multicolumn{1}{c}{\(1024\)} & \multicolumn{1}{c}{\(4096\)} \\
\midrule
LG {\scriptsize(S)} & \(0\) & \(16\) & \longcell{\(0.00\)\\{\tiny(\(0.00\))}} & \longcell{\(0.00\)\\{\tiny(\(0.00\))}} \\[2.2ex]
LG {\scriptsize(S)} & \(0\) & \(64\) & \longcell{\(0.01\)\\{\tiny(\(0.00\))}} & \longcell{\(0.00\)\\{\tiny(\(0.00\))}} \\[2.2ex]
LG {\scriptsize(S)} & \(0\) & \(256\) & \longcell{\(0.01\)\\{\tiny(\(0.00\))}} & \longcell{\(0.00\)\\{\tiny(\(0.00\))}} \\[2.2ex]
LG {\scriptsize(S)} & \(0\) & \(1024\) & \longcell{\(0.01\)\\{\tiny(\(0.00\))}} & \longcell{\(0.00\)\\{\tiny(\(0.00\))}} \\[2.2ex]
LG {\scriptsize(RL)} & \(0\) & \(16\) & \longcell{\(0.00\)\\{\tiny(\(0.00\))}} & \longcell{\(0.00\)\\{\tiny(\(0.00\))}} \\[2.2ex]
LG {\scriptsize(RL)} & \(0\) & \(64\) & \longcell{\(0.01\)\\{\tiny(\(0.02\))}} & \longcell{\(0.01\)\\{\tiny(\(0.01\))}} \\
\bottomrule
\end{tabular}
\end{table*}

\begin{table*}[t]
\linespread{0.6}\selectfont\centering
\centering
\caption{Test accuracy with SD for different game variants, using CelebA dataset and over 10 seeds. During test \(\lambda_{\text{test}}\) is set to \(0.75\).}
\label{table:lg_celeba_075_test_reward}
\begin{tabular}{lrrrr}
\toprule
Game & \(\lambda\) & \(|\sC|\) & \multicolumn{2}{c}{\(|\sC|\) (test)} \\[1ex]
 &  &  & \multicolumn{1}{c}{\(1024\)} & \multicolumn{1}{c}{\(4096\)} \\
\midrule
LG {\scriptsize(S)} & \(0\) & \(16\) & \longcell{\(0.00\)\\{\tiny(\(0.00\))}} & \longcell{\(0.00\)\\{\tiny(\(0.00\))}} \\[2.2ex]
LG {\scriptsize(S)} & \(0\) & \(64\) & \longcell{\(0.00\)\\{\tiny(\(0.00\))}} & \longcell{\(0.00\)\\{\tiny(\(0.00\))}} \\[2.2ex]
LG {\scriptsize(S)} & \(0\) & \(256\) & \longcell{\(0.00\)\\{\tiny(\(0.00\))}} & \longcell{\(0.00\)\\{\tiny(\(0.00\))}} \\[2.2ex]
LG {\scriptsize(S)} & \(0\) & \(1024\) & \longcell{\(0.01\)\\{\tiny(\(0.00\))}} & \longcell{\(0.00\)\\{\tiny(\(0.00\))}} \\[2.2ex]
LG {\scriptsize(RL)} & \(0\) & \(16\) & \longcell{\(0.01\)\\{\tiny(\(0.00\))}} & \longcell{\(0.00\)\\{\tiny(\(0.00\))}} \\[2.2ex]
LG {\scriptsize(RL)} & \(0\) & \(64\) & \longcell{\(0.01\)\\{\tiny(\(0.00\))}} & \longcell{\(0.00\)\\{\tiny(\(0.00\))}} \\
\bottomrule
\end{tabular}
\end{table*}



\end{document}